\documentclass[10pt,twocolumn,letterpaper]{article}

\usepackage{cvpr}              
\definecolor{cvprblue}{rgb}{0.21,0.49,0.74}
\usepackage[pagebackref,breaklinks,colorlinks,allcolors=cvprblue]{hyperref}
\usepackage{multirow}
\usepackage[table]{xcolor}
\usepackage{amsthm}

\def\paperID{} 
\def\confName{CVPR}
\def\confYear{2026}

\title{MacTok: Robust Continuous Tokenization for Image Generation}

\author{
    Hengyu Zeng$^{1}$\thanks{These authors contributed equally to this work.}\qquad 
    Xin Gao$^{1}$\footnotemark[1]\qquad 
    Guanghao Li$^1$\qquad 
    Yuxiang Yan$^1$ \\
    Jiaoyang Ruan$^1$\qquad 
    Junpeng Ma$^1$\qquad 
    Haoyu Albert Wang$^1$\qquad  
    Jian Pu$^1$\thanks{Corresponding author.} \\
    $^1$ Fudan University \\ 
    {\tt\small \{hyzeng24, gaoxin23\}@m.fudan.edu.cn, \{jianpu\}@fudan.edu.cn}
}

\begin{document}
\maketitle
\begin{abstract}
Continuous image tokenizers enable efficient visual generation, and those based on variational frameworks can learn smooth, structured latent representations through KL regularization. Yet this often leads to posterior collapse when using fewer tokens, where the encoder fails to encode informative features into the compressed latent space. To address this, we introduce \textbf{MacTok}, a \textbf{M}asked \textbf{A}ugmenting 1D \textbf{C}ontinuous \textbf{Tok}enizer that leverages image masking and representation alignment to prevent collapse while learning compact and robust representations. MacTok applies both random masking to regularize latent learning and DINO-guided semantic masking to emphasize informative regions in images, forcing the model to encode robust semantics from incomplete visual evidence. Combined with global and local representation alignment, MacTok preserves rich discriminative information in a highly compressed 1D latent space, requiring only 64 or 128 tokens. On ImageNet, MacTok achieves a competitive gFID of 1.44 at 256$\times$256 and a state-of-the-art 1.52 at 512$\times$512 with SiT-XL, while reducing token usage by up to 64$\times$. These results confirm that masking and semantic guidance together prevent posterior collapse and achieve efficient, high-fidelity tokenization.
\end{abstract}    
\section{Introduction}
\label{sec:intro}

\begin{figure}[t]
    \centering
    \includegraphics[width=\linewidth]{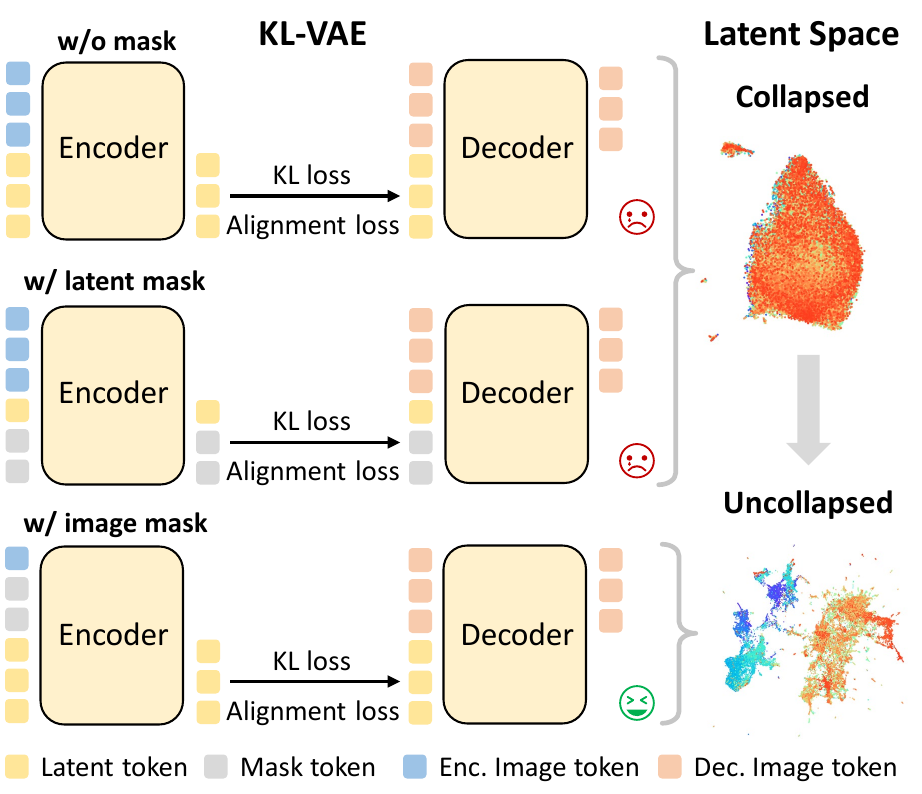}
    \caption{\textbf{Effect of random masking in continuous tokenizers.} 
    Left: plain KL-VAE, latent token masking, and image token masking, with only the latter preventing posterior collapse. 
    Right: collapsed latent space shows poor structure, while the uncollapsed one yields well-structured and diverse representations.}
    \label{fig:comparison of KL-VAE}
    \vskip -0.05in
\end{figure}
\begin{figure*}
    \centering
    \includegraphics[width=\linewidth]{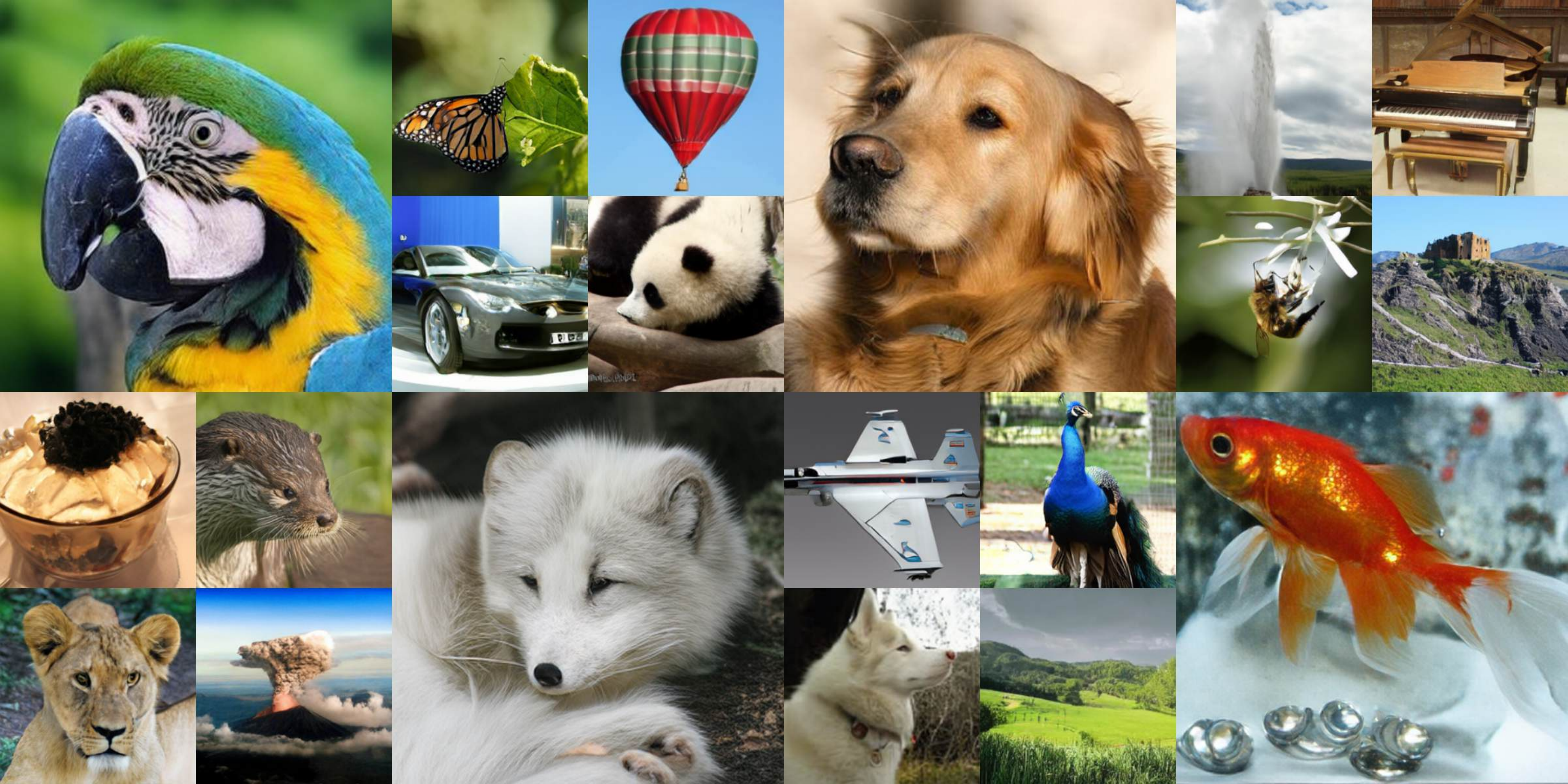}
    \caption{Generation results produced by generative models with MacTok using 64 and 128 tokens on ImageNet at 256$\times$256 and 512$\times$512.}
    \label{fig:256_512_vis}

\end{figure*}

In recent years, visual generative models have rapidly advanced by modeling data in compressed latent spaces, substantially reducing the cost of training and inference. A crucial component in this paradigm is the \textit{image tokenizer}, which maps raw images into latent representations that are then used by diffusion~\cite{DiT}, flow \cite{SiT}, or autoregressive models \cite{maskgit}. 
Tokenizers generally fall into two categories: \textit{discrete} and \textit{continuous}.
Vector-Quantized Variational Auto-Encoders (VQ-VAE) \cite{vq-vae} and its variants (\eg, VQ-GAN \cite{VQ-GAN}, TiTok \cite{TiTok}) represent the discrete family, discretizing the latent space through a finite codebook with straight-through estimation to ensure stability but at the cost of quantization error.
In contrast, Kullback–Leibler Variational Auto-Encoders (KL-VAE) \cite{kl-vae} and related methods (\eg, SD-VAE \cite{SD-VAE}, MAR-VAE \cite{MAR}) define a continuous latent space regularized by a Gaussian prior via KL divergence, yielding smooth representations but prone to posterior collapse under strong compression~\cite{vae-lossy,softvq-vae}.

Although both discrete and continuous tokenizers have been extensively studied, achieving an optimal balance between compression efficiency and generation quality remains challenging.
Recent studies have investigated strategies to enhance the efficiency of VQ-based discrete tokenizers, including two-stage training schemes to obtain more compact token representations~\cite{TiTok}, as well as the integration of pre-trained visual features~\cite{imagefolder}.
In contrast, KL-based continuous tokenizer remain underexplored, primarily due to the persistent issue of posterior collapse. 
When trained with strong KL regularization, KL-VAEs tend to excessively constrain the latent distribution, pushing it toward an isotropic Gaussian prior and thereby discarding meaningful semantics~\cite{vae-lossy,cyclical_kl,beta-vae}. Our experiments with KL-VAEs reveal frequent posterior collapse under strong compression, where latents lose essential information, leading to poor reconstruction and generation. While prior work addresses this through KL weight tuning~\cite{beta-vae,cyclical_kl}, these methods require delicate hyperparameter tuning and do not fundamentally resolve latent degeneration.

\begin{figure}[t]
    \centering
    \includegraphics[width=\linewidth]{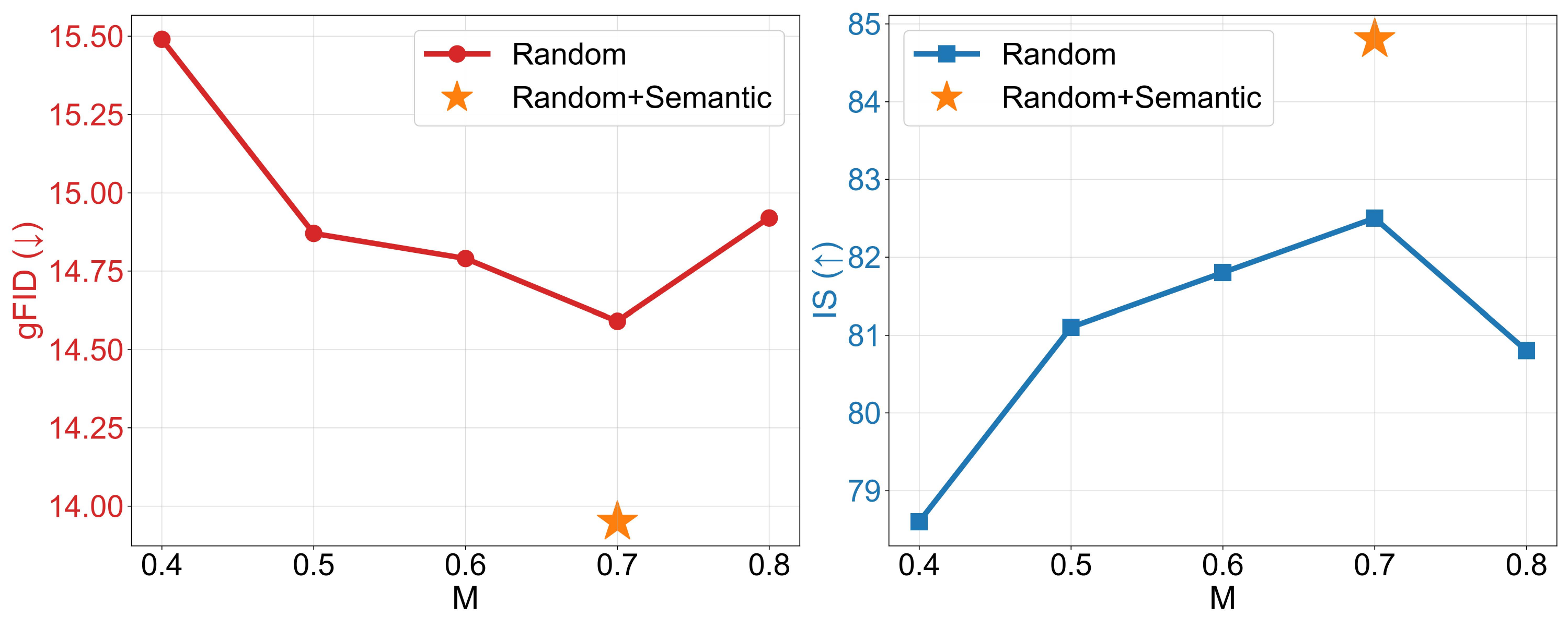}
\caption{Generation performance of MacTok with varying mask ratios sampled up to $M$ as detailed in \cref{sec:method3.2}. The orange star corresponds to random and semantic mask with equal probability.}
    \label{fig:mask_ratio_performance}
\vskip -0.1in
\end{figure}

The key to overcoming posterior collapse lies in learning a robust and discriminative latent representation that preserves semantics under strong compression.
A well-structured latent space mitigates collapse by maintaining sufficient mutual information between the input and its encoded representation, preventing overreliance on the prior~\cite{qian2019enhancing,infovae}.
To this end, we draw inspiration from masked representation learning~\cite{MAE} and require the tokenizer to reconstruct images from partial inputs. Specifically, we investigate two masking schemes, as shown in \cref{fig:comparison of KL-VAE}: one operates on latent tokens, and the other on image tokens. For latent tokens, previous work suggests that randomly dropping some of them can improve representation robustness~\cite{DeTok}, and our experiments likewise show that latent token masking temporarily delays posterior collapse compared with a standard KL-VAE.
However, this stabilization is short-lived and the model eventually collapses as training continues (details in Appendix~\ref{appendix:appendix_kl_loss}).
In contrast, masking image tokens yields consistently stable optimization and more robust representations, as it forces both encoder and decoder to infer from incomplete inputs, encouraging the latent space to capture global structural semantics.
Building on this observation, we further hypothesize that masking can make the latent space robust and semantically discriminative. Therefore, we propose a DINO-guided semantic masking strategy that selectively occludes the most informative regions identified by computing the similarity between the classification token and each patch token in DINOv2~\cite{dinov2}. Unlike random masking, this targeted strategy enforces the reconstruction of key semantic regions from incomplete observations, implicitly transferring semantic priors from the image space into the latent space. As shown in \cref{fig:mask_ratio_performance}, semantic masking significantly improves generation quality by producing more semantically rich latent representations.

Furthermore, state-of-the-art tokenizers, both discrete and continuous, often improve reconstruction and generation quality by aligning latent representations with external semantic features~\cite{repa}, such as those extracted by DINOv2. However, previous approaches have notable drawbacks: they rely on a fixed token count~\cite{va-vae}, perform only coarse global alignment~\cite{imagefolder}, or require multiple auxiliary objectives~\cite{MAETok}. To address these issues, we propose the global and local representation alignment that aligns each latent token with both holistic and local region features. This approach facilitates more consistent semantic guidance across varying token lengths, leading to a well-structured latent space for improved reconstruction and generation.

Extensive experiments on ImageNet demonstrate the effectiveness of MacTok in mitigating posterior collapse and enhancing generation quality under strong compression.
We observe that an appropriate mask ratio consistently improves generation performance. Combining masking and representation alignment, MacTok achieves a superior trade-off between reconstruction fidelity and generative quality, reaching competitive rFID and state-of-the-art gFID with only 64 and 128 tokens on ImageNet at 256$\times$256 and 512$\times$512. Our contributions are summarized as follows:
\begin{itemize}
\item We identify posterior collapse in continuous tokenizers as a form of over-regularization that suppresses semantic information, and show that maintaining information flow through reconstruction from incomplete inputs stabilizes latent learning under strong compression.
\item We introduce MacTok, which jointly regularizes and semantically structures the latent space through both image masking and representation alignment. This prevents posterior collapse while preserving discriminative features.
\item MacTok achieves competitive reconstruction and state-of-the-art generation results on ImageNet, with a gFID of \textbf{1.44} at 256$\times$256 and \textbf{1.52} at 512$\times$512 using only \textbf{64 and 128 tokens}, ensuring high fidelity, strong compression, and stable training.
\end{itemize}

\section{Related Work}
\label{sec:related}
\subsection{Image Tokenization}

Image tokenization has emerged as a fundamental component of modern generative modeling, converting images from the pixel space into compact latent representations for efficient synthesis.
Most approaches build on autoencoders equipped with quantizers or variational posteriors, each exhibiting characteristic limitations.
\textit{Discrete tokenizers} such as VQ-VAE~\cite{vq-vae} and VQ-GAN~\cite{VQ-GAN} learn a codebook of discrete tokens, with the latter employing adversarial training for improved perceptual fidelity.
However, they suffer from limited gradient flow and suboptimal codebook utilization due to the straight-through estimator used during training.
Subsequent works improve efficiency through better codebook learning strategies. For example, IBQ~\cite{IBQ} updates the full codebook via index backpropagation, while MAGVIT-v2~\cite{MAGVIT-v2} adopts Look-up Free Quantization (LFQ) to bypass explicit codebook management.
\textit{Continuous tokenizers}, in contrast, map images to continuous latent spaces without discrete quantization.
MAETok~\cite{MAETok} employs a plain autoencoder with auxiliary losses for semantic guidance, while SoftVQ-VAE~\cite{softvq-vae} enhances latent space capacity under high compression by bridging between discrete and continuous encoding via soft codeword aggregation.
However, KL-based continuous models remain vulnerable to posterior collapse~\cite{vae-lossy,softvq-vae} under strong compression, where the encoder fails to preserve informative features, leading to poor reconstruction and limited generative fidelity.
Our work introduces a KL-VAE framework that combines random and semantic masking with global and local representation alignment, effectively mitigating collapse and enhancing both reconstruction fidelity and generative quality.

\subsection{Image Generation}
The integration of tokenization into generative models has driven significant advances in image synthesis, particularly in diffusion models \cite{DiT,SiT,va-vae} and autoregressive \cite{MAR,var,llamagen} architectures. Diffusion-based approaches excel at generating high-resolution images by iteratively denoising representations in continuous latent spaces \cite{gaogood}. In contrast, autoregressive~\cite{var,llamagen} and masked prediction models \cite{maskgit,MAGVIT-v2} often operate in discrete token spaces. Convolutional backbones \cite{u-net,pixelcnn} have gradually been replaced by transformer-based architectures \cite{DiT,SiT}, which offer improved scalability and representational capacity for both 2D synthesis and broader spatial modeling \cite{li2025papl, li2025constrained, li2026ec, liartdeco, chen2024multi}.

\subsection{Representation Alignment for Generation}

 More recent works explore incorporating semantic information into tokenization and generation to enhance image synthesis. In diffusion-based models, methods such as~\cite{repa,repa-e,REG} align transformer representations with pretrained visual embeddings, improving both training efficiency and generation quality. For discrete tokenizers, VQGAN-LC~\cite{VQGAN-LC} leverages CLIP~\cite{CLIP} features for codebook initialization to boost utilization and perceptual fidelity, while VQ-KD~\cite{VQ-KD} trains tokenizers to reconstruct features from pretrained visual encoders. ImageFolder~\cite{imagefolder} adopts product quantization to produce spatially aligned semantic and detail tokens, reducing sequence length without compromising quality. In continuous tokenizers, VA-VAE~\cite{va-vae} aligns its latent space with vision foundation models to stabilize optimization. TexTok~\cite{TexTok} introduces textual information into tokenization to enhance both reconstruction and generation. SoftVQ-VAE~\cite{softvq-vae} employs soft categorical posteriors for feature alignment, and MAETok~\cite{MAETok} incorporates auxiliary semantic targets from HOG~\cite{HOG}, DINOv2~\cite{dinov2}, and SigCLIP~\cite{SigCLIP}. Our work combine global and local representation alignment as latent space regularization, leading to improved reconstruction fidelity and generative performance.

\begin{figure*}[t]
    \centering
    \includegraphics[width=\linewidth]{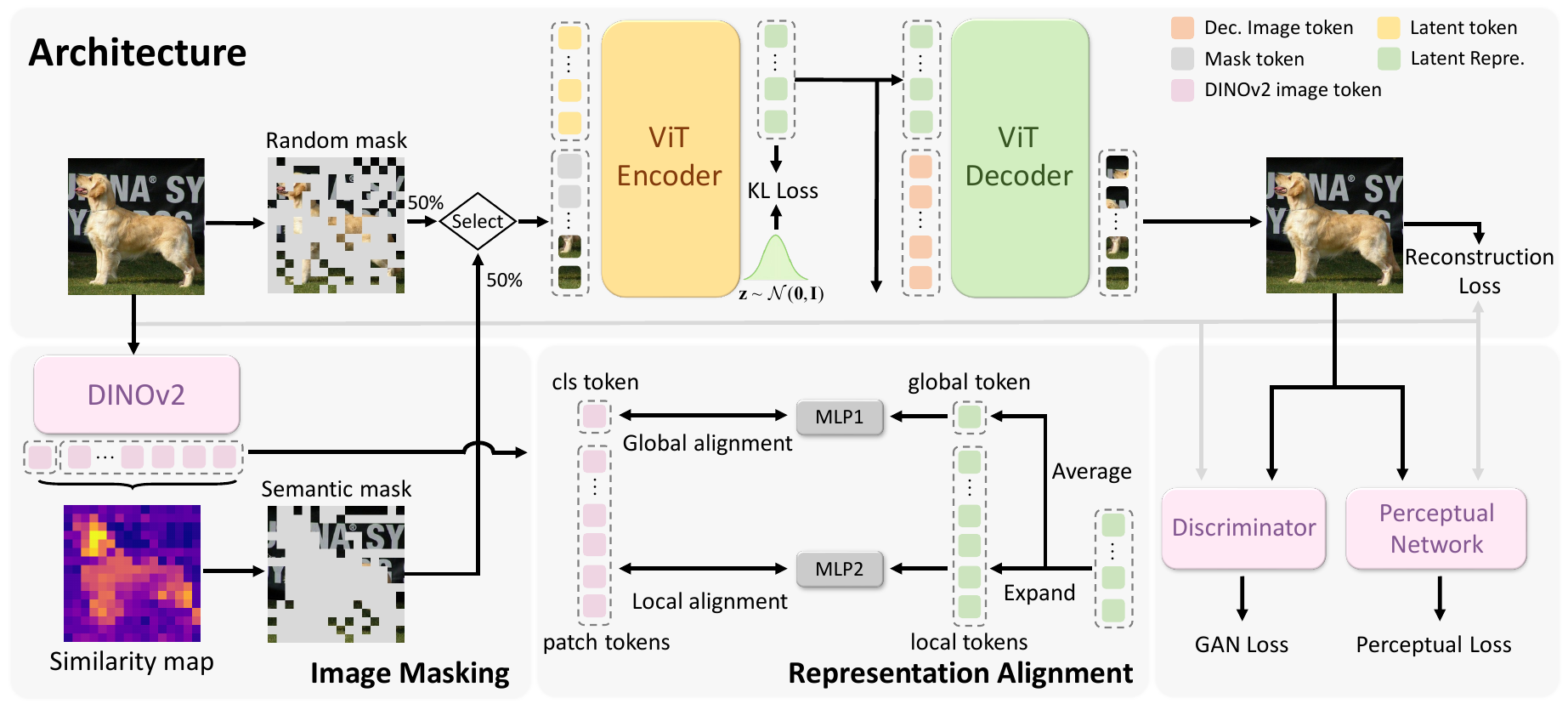}
\caption{
\textbf{Overview of the MacTok framework.}
\textit{Top:} Transformer-based encoder and decoder operating on image, latent, and mask tokens. 
\textit{Bottom left:} DINO-guided image masking introduces semantic priors. 
\textit{Bottom center:} Global and local representation alignment between latent and pretrained visual representations. 
\textit{Bottom right:} Discriminator and perceptual networks provide auxiliary supervision. 
}
    \label{fig:framework}
\end{figure*}
\section{Method}
\label{sec:method}

We present MacTok, a 1D continuous tokenizer that prevents posterior collapse by enforcing information preservation through image masking and feature alignment.
As illustrated in \cref{fig:framework}, MacTok reconstructs complete images from incomplete visual evidence using two complementary masking strategies, while aligning the latent space with pretrained features to maintain robust and discriminative representations even under strong compression.

\subsection{Continuous Tokenizer Architecture}
As illustrated in~\cref{fig:framework}, MacTok adopts a Vision Transformer (ViT) as both the encoder $E$ and decoder $D$~\cite{ViT,ViT-VQGAN,ordermind,pointssc}. 
Building on recent one-dimensional tokenizer designs \cite{TiTok,softvq-vae,MAETok} that allow flexible token lengths for image representation, we extend ViT to jointly process \emph{image tokens} and \emph{latent tokens}, where the latent tokens serve as compact representations for reconstruction and generation.

The encoder $E$ first partitions an input image $\mathbf{I} \in \mathbb{R}^{H \times W \times 3}$ into non-overlapping patches of size $P$, producing image tokens $\mathbf{x} \in \mathbb{R}^{N \times D}$, 
where $N=\frac{HW}{P^2}$ is the number of patches and $D$ denotes the embedding dimension. 
A set of learnable tokens $\mathbf{z} \in \mathbb{R}^{L \times D}$ is concatenated with the image tokens, forming the sequence $[\mathbf{x};\mathbf{z}] \in \mathbb{R}^{(N+L)\times D}$, which is then passed through the encoder. 
The encoder outputs corresponding to the latent tokens are taken as the latent representations $\hat{\mathbf{z}} \in \mathbb{R}^{L \times Z}$: $\mathbf{\hat{z}} = E([\mathbf{x};\mathbf{z}])$, where $Z$ denotes the latent dimensionality.
To model a continuous latent space, the latent vector $\hat{\mathbf{z}}$ is treated as a Gaussian random variable parameterized by its mean and variance.
A KL divergence regularizes the posterior toward an isotropic Gaussian prior, promoting smoothness in the latent space. However, under strong compression, this constraint may excessively regularize latent representations, leading to posterior collapse~\cite{vae-lossy,softvq-vae, gaodeep}. This issue is precisely what MacTok is designed to address.

During decoding, the sampled latent representations $\hat{\mathbf{z}}$ are concatenated with learnable reconstruction tokens $\mathbf{h} \in \mathbb{R}^{N \times Z}$ and passed to the decoder: $\hat{\mathbf{x}} = D([\mathbf{h};\hat{\mathbf{z}}])$.
Decoder outputs corresponding to $\mathbf{h}$ are projected through a linear layer to reconstruct the image $\hat{\mathbf{I}}$.
We use 2D absolute positional embeddings for image tokens to preserve spatial structure and 1D embeddings for latent and reconstruction tokens without explicit spatial coordinates.

\subsection{Image Masking for Latent Preservation}\label{sec:method3.2}

Posterior collapse occurs when the latent variables fail to capture informative content, causing the decoder to reconstruct images primarily from priors. 
To address this, MacTok enforces information flow via \emph{masked reconstruction}, requiring the model to infer missing content from partial inputs and thus maintain robust representations.
Two complementary masking strategies, \textbf{Random} and \textbf{Semantic}, are applied with equal probability during training.

\noindent \textbf{Random Masking.}
At each iteration, a random subset of image patches is replaced with mask tokens before encoding, forcing the latent tokens to infer complete information and reconstruct the missing regions from visible context.
The mask ratio $m$ is uniformly sampled from $[-0.1, M]$ and clipped to $[0, M]$ (typically $M{=}0.7$), which allows the model to occasionally observe unmasked images ($m{=}0$) and prevents excessive reconstruction degradation from persistent masking.
This stochastic corruption of images compels the latent variables to encode essential information for reconstruction, thereby preventing posterior collapse (see analysis in Appendix \ref{appendix:theory}).

\noindent \textbf{Semantic Masking.}
While random masking enhances robustness, it is agnostic to semantic structure.
To inject semantic priors into the latent space, we further employ a guided masking strategy based on DINOv2 features.
Given the classification token $\mathbf{c} \in \mathbb{R}^{D}$ and patch tokens $\mathbf{P} = \{\mathbf{p}_i \in \mathbb{R}^{D}\}_{i=1}^{N}$, 
we compute the cosine similarity between the classification token and each patch token:
\begin{equation}
s_i = \tfrac{\mathbf{c}^\top \mathbf{p}_i}{\|\mathbf{c}\| \|\mathbf{p}_i\|}, \quad i = 1, \dots, N,
\end{equation}
where $s_i$ denotes the semantic relevance of patch $i$. 
We then select the top $\lfloor m\times N \rfloor$ patches with the highest similarity scores:
\begin{equation}
M_p = \mathrm{TopK}(\{s_i\}_{i=1}^{N}, \lfloor m \times N \rfloor),
\end{equation}
where $M_p$ is the index set of masked patches.
When semantically important regions are masked, the reconstruction task becomes substantially harder, encouraging the latent tokens to capture object-level structures and global context. This semantic masking implicitly transfers knowledge from the image space to the latent space and yields more discriminative representations.

\subsection{Local and Global Representation Alignment}

We further align MacTok’s latent representations with pretrained DINOv2 features~\cite{dinov2} to enforce semantic consistency, as representation alignment has been shown to be an effective auxiliary objective for enhancing visual generation~\cite{repa,softvq-vae,va-vae}.
Unlike previous approaches, we introduce a lightweight \emph{global and local alignment} that links each latent token to both regional and holistic semantics, improving structural coherence and detail preservation.

\noindent \textbf{Local and Gobal Feature Construction.}
Let $\hat{\mathbf{z}} = \{\hat{\mathbf{z}}_i\}_{i=1}^{L}$ denote the latent tokens produced by the encoder.  
To match the spatial granularity of DINOv2 patch features $\{\mathbf{p}_i\}_{i=1}^{N}$, we expand $\hat{\mathbf{z}}$ into a sequence $\tilde{\mathbf{z}}_{\text{loc}} \in \mathbb{R}^{N \times Z}$ by repeating each latent token $r$ times, where $r = N / L$:
\begin{equation}
\tilde{\mathbf{z}}_{\text{loc}} = \text{Expand}(\hat{\mathbf{z}}, r).
\end{equation}
In parallel, we obtain a global latent representation by average pooling:
\begin{equation}
\tilde{\mathbf{z}}_{\text{glob}} = \frac{1}{L}\sum_{i=1}^{L}\hat{\mathbf{z}}_i.
\end{equation}
Both local and global features are linearly projected into the DINOv2 feature space using two lightweight MLPs:
\begin{equation}
\mathbf{o}_{\text{loc}} = \mathrm{MLP}_1(\tilde{\mathbf{z}}_{\text{loc}}), 
\quad
\mathbf{o}_{\text{glob}} = \mathrm{MLP}_2(\tilde{\mathbf{z}}_{\text{glob}}).
\end{equation}

\noindent \textbf{Representation Alignment Loss.}
We encourage the projected latent features to align with the corresponding DINOv2 features using cosine similarity.  
The local alignment compares $\mathbf{o}_{\text{loc}}$ with patch tokens $\{\mathbf{p}_i\}$, while the global alignment matches $\mathbf{o}_{\text{glob}}$ with the classification token $\mathbf{c}$:
\begin{equation}
L_{\text{RA}} = 
-\frac{1}{(N+1)}\Big(\sum_{i=1}^{N}\text{sim}(\mathbf{o}_{\text{loc},i}, \mathbf{p}_i)
+\text{sim}(\mathbf{o}_{\text{glob}}, \mathbf{c})\Big).
\end{equation}
This loss encourages latent tokens to encode semantically coherent information at both patch and image levels, resulting in a well-structured latent space that enhances reconstruction and generation quality under strong compression.

\subsection{Training Objectives}

MacTok is optimized with a composite objective that includes reconstruction, perceptual~\cite{perceptual_loss,perceptual_loss_2,perceptual_loss_3,perceptual_loss_4}, and adversarial~\cite{GAN_loss} terms, following~\cite{VQ-GAN,softvq-vae}, as well as regularization~\cite{kl-vae} and representation alignment terms:
\begin{equation}
L = L_{\text{recon}} 
  + \lambda_1 L_{\text{percep}} 
  + \lambda_2 L_{\text{adv}} 
  + \lambda_3 L_{\text{KL}} 
  + \lambda_4 L_{\text{RA}},
\end{equation}
where $\lambda_1$-$\lambda_4$ are weighting coefficients. 
$L_{\text{recon}}$ is a pixel-wise reconstruction loss, 
$L_{\text{percep}}$ enforces perceptual similarity in a pretrained feature space, 
$L_{\text{adv}}$ encourages realistic image synthesis through adversarial learning, 
$L_{\text{KL}}$ regularizes the latent distribution toward a Gaussian prior, 
and $L_{\text{RA}}$ is the proposed representation alignment loss.

\section{Experiments}
\label{sec:experiments}

\begin{table*}[t]
\caption{
\textbf{System-level comparison on ImageNet 256$\times$256 conditional generation.}
``\# Params (G)'' denotes generator parameters; ``Tok. Model'' refers to the tokenizer model type; ``Token Type'' indicates 1D or 2D tokenization; ``\# Params (T)'' denotes tokenizer parameters; and ``\# Tokens'' represents the number of latent tokens. $^\ddagger$ denotes methods that rely on pretrained vision models.
}
\label{tab:256}
\centering
\resizebox{0.9\linewidth}{!}
{
\begin{tabular}{@{}lcccccccccc}
\toprule
\multicolumn{1}{l|}{}                                                              &                                                &                                              &                                              &                                                &                             & \multicolumn{1}{c|}{}                             & \multicolumn{2}{c|}{w/o CFG}                               & \multicolumn{2}{c}{w/ CFG} \\
\multicolumn{1}{@{}l|}{\multirow{-2}{*}{Method}}                                      & \multirow{-2}{*}{\# Params (G)}                 & \multirow{-2}{*}{Tok. Model}                 & \multirow{-2}{*}{Token Type}                  & \multirow{-2}{*}{\# Params (T)}                 & \multirow{-2}{*}{\# Tokens↓} & \multicolumn{1}{c|}{\multirow{-2}{*}{Tok. rFID↓}} & gFID↓ & \multicolumn{1}{c|}{IS↑}                           & gFID↓        & IS↑         \\ \midrule
\textit{Auto-regressive}                                                           &                                                &                                              &                                              &                                                &                             &                                                   &       &                                                    &              &             \\
\multicolumn{1}{l|}{ViT-VQGAN \cite{ViT-VQGAN}}                                                     & 1.7B                                           & VQ                                           & 2D                                           & 64M                                            & 1024                        & \multicolumn{1}{c|}{1.28}                         & 4.17  & \multicolumn{1}{c|}{175.1}                         & -            & -           \\
\multicolumn{1}{l|}{RQ-Trans. \cite{rq-vae}}                                                     & 3.8B                                           & RQ                                           & 2D                                           & 66M                                            & 256                         & \multicolumn{1}{c|}{3.20}                          & -     & \multicolumn{1}{c|}{-}                             & 3.80         & 323.7       \\
\multicolumn{1}{l|}{MaskGIT \cite{maskgit}}                                                       & 227M                                           & VQ                                           & 2D                                           & 66M                                            & 256                         & \multicolumn{1}{c|}{2.28}                         & 6.18  & \multicolumn{1}{c|}{182.1}                         & -            & -           \\
\multicolumn{1}{l|}{LlamaGen-3B \cite{llamagen}}                                                   & 3.1B                                           & VQ                                           & 2D                                           & 72M                                            & 576                         & \multicolumn{1}{c|}{2.19}                         & -     & \multicolumn{1}{c|}{-}                             & 2.18         & 263.3       \\
\multicolumn{1}{l|}{WeTok \cite{wetok}}                                                         & 1.5B                                           & VQ                                           & 2D                                           & 400M                                           & 256                         & \multicolumn{1}{c|}{0.60}                          & -     & \multicolumn{1}{c|}{-}                             & 2.31         & 276.6       \\
\multicolumn{1}{l|}{VAR \cite{var}}                                                           & 2B                                             & \text{MSRQ}                                      & 2D                                           & 109M                                           & 680                         & \multicolumn{1}{c|}{0.90}                          & -     & \multicolumn{1}{c|}{-}                             & 1.92         & 323.1       \\
\multicolumn{1}{l|}{MaskBit \cite{maskbit}}                                                       & 305M                                           & LFQ                                          & 2D                                           & 54M                                            & 256                         & \multicolumn{1}{c|}{1.61}                         & -     & \multicolumn{1}{c|}{-}                             & 1.52         & 328.6       \\
\multicolumn{1}{l|}{MAR-H \cite{MAR}}                                                         & 943M                                           & KL                                           & 2D                                           & 66M                                            & 256                         & \multicolumn{1}{c|}{1.22}                         & 2.35  & \multicolumn{1}{c|}{227.8}                         & 1.55         & 303.7       \\
\multicolumn{1}{l|}{\textit{l}-DeTok \cite{DeTok}}                                                       & 479M                                           & KL                                           & 2D                                           & 172M                                           & 256                         & \multicolumn{1}{c|}{0.62}                         & 1.86  & \multicolumn{1}{c|}{238.6}                         & 1.35         & 304.1       \\
\multicolumn{1}{l|}{TiTok-S-128 \cite{TiTok}}                                                   & 287M                                           & VQ                                           & 1D                                           & 72M                                            & 128                         & \multicolumn{1}{c|}{1.61}                         & -     & \multicolumn{1}{c|}{-}                             & 1.97         & 281.8       \\
\multicolumn{1}{l|}{GigaTok$^\ddagger$ \cite{gigatok}}                                                       & 111M                                           & VQ                                           & 1D                                           & 622M                                           & 256                         & \multicolumn{1}{c|}{0.51}                         & -     & \multicolumn{1}{c|}{-}                             & 3.15         & 224.3       \\
\multicolumn{1}{l|}{ImageFolder$^\ddagger$ \cite{imagefolder}}                                                   & 362M                                           & MSRQ                                         & 1D                                           & 176M                                           & 286                         & \multicolumn{1}{c|}{0.80}                          & -     & \multicolumn{1}{c|}{-}                             & 2.60         & 295.0       \\ \midrule
\textit{Diffusion-based}                                                           &                                                &                                              &                                              &                                                &                             &                                                   &       &                                                    &              &             \\
\multicolumn{1}{l|}{LDM-4 \cite{SD-VAE}}                                                         & 400M                                           &                                              & 2D                                           &                                                &                             & \multicolumn{1}{c|}{}                             & 10.56 & \multicolumn{1}{c|}{103.5}                         & 3.60         & 247.7       \\
\multicolumn{1}{l|}{U-ViT-H/2 \cite{U-ViT}}                                                     & 501M                                           &                                              & 2D                                           &                                                &                             & \multicolumn{1}{c|}{}                             & -     & \multicolumn{1}{c|}{-}                             & 2.29         & 263.9       \\
\multicolumn{1}{l|}{MDTv2-XL/2 \cite{MDTv2}}                                                    & 676M                                           & \multirow{-3}{*}{\text{KL}}                        & 2D                                           & \multirow{-3}{*}{55M}                          & \multirow{-3}{*}{4096}      & \multicolumn{1}{c|}{\multirow{-3}{*}{0.27}}       & 5.06  & \multicolumn{1}{c|}{155.6}                         & 1.58         & 314.7       \\
\multicolumn{1}{l|}{DiT-XL/2 \cite{DiT}}                                                      & 675M                                           &                                              & 2D                                           &                                                &                             & \multicolumn{1}{c|}{}                             & 9.62  & \multicolumn{1}{c|}{121.5}                         & 2.27         & 278.2       \\
\multicolumn{1}{l|}{SiT-XL/2 \cite{SiT}}                                                      &                                                &                                              & 2D                                           &                                                &                             & \multicolumn{1}{c|}{}                             & 8.30  & \multicolumn{1}{c|}{131.7}                         & 2.06         & 270.3       \\
\multicolumn{1}{l|}{+REPA$^\ddagger$ \cite{repa}}                                                         & \multirow{-2}{*}{675M}                         & \multirow{-3}{*}{\text{KL}}                        & 2D                                           & \multirow{-3}{*}{84M}                          & \multirow{-3}{*}{1024}      & \multicolumn{1}{c|}{\multirow{-3}{*}{0.62}}       & 5.90  & \multicolumn{1}{c|}{157.8}                         & 1.42         & 305.7       \\
\multicolumn{1}{l|}{LightningDiT$^\ddagger$ \cite{va-vae}}                                                  & 675M                                           & KL                                           & 2D                                           & 70M                                            & 256                         & \multicolumn{1}{c|}{0.28}                         & 2.17  & \multicolumn{1}{c|}{205.6}                         & 1.35         & 295.3       \\
\multicolumn{1}{l|}{TexTok-256 \cite{TexTok}}                                                     & 675M                                           & KL                                           & 1D                                           & 176M                                           & 256                          & \multicolumn{1}{c|}{0.73}                         & -     & \multicolumn{1}{c|}{-}                             & 1.46         & 303.1       \\
\multicolumn{1}{l|}{MAETok$^\ddagger$ \cite{MAETok}}                                                        & 675M                                           & AE                                           & 1D                                           & 176M                                           & 128                         & \multicolumn{1}{c|}{0.48}                         & 2.31  & \multicolumn{1}{c|}{216.5}                         & 1.67         & 311.2       \\
\multicolumn{1}{l|}{SoftVQ-VAE$^\ddagger$ \cite{softvq-vae}}                                                    & 675M                                           & SoftVQ                                       & 1D                                           & 176M                                           & 64                          & \multicolumn{1}{c|}{0.88}                         & 5.98  & \multicolumn{1}{c|}{138.0}                         & 1.78         & 279.0       \\ \midrule
\textit{Ours}                                                                      &                                                &                                              &                                              &                                                &                             &                                                   &       &                                                    &              &             \\
\rowcolor[HTML]{EFEFEF} 
\multicolumn{1}{l|}{\cellcolor[HTML]{EFEFEF}}                                      & \cellcolor[HTML]{EFEFEF}                       & \cellcolor[HTML]{EFEFEF}                     & \cellcolor[HTML]{EFEFEF}                     & \cellcolor[HTML]{EFEFEF}                       & 64                          & \multicolumn{1}{c|}{\cellcolor[HTML]{EFEFEF}0.75} & 4.15  & \multicolumn{1}{c|}{\cellcolor[HTML]{EFEFEF}167.8} & 1.68         & 307.3       \\
\rowcolor[HTML]{EFEFEF} 
\multicolumn{1}{l|}{\multirow{-2}{*}{\cellcolor[HTML]{EFEFEF}MacTok+LightningDiT$^\ddagger$}} & \multirow{-2}{*}{\cellcolor[HTML]{EFEFEF}675M} & \cellcolor[HTML]{EFEFEF}                     & \cellcolor[HTML]{EFEFEF}                     & \cellcolor[HTML]{EFEFEF}                       & 128                         & \multicolumn{1}{c|}{\cellcolor[HTML]{EFEFEF}0.43} & 3.12  & \multicolumn{1}{c|}{\cellcolor[HTML]{EFEFEF}186.2} & 1.50         & 299.8       \\
\rowcolor[HTML]{EFEFEF} 
\multicolumn{1}{l|}{\cellcolor[HTML]{EFEFEF}}                                      & \cellcolor[HTML]{EFEFEF}                       & \cellcolor[HTML]{EFEFEF}                     & \cellcolor[HTML]{EFEFEF}                     & \cellcolor[HTML]{EFEFEF}                       & 64                          & \multicolumn{1}{c|}{\cellcolor[HTML]{EFEFEF}0.75} & 3.77  & \multicolumn{1}{c|}{\cellcolor[HTML]{EFEFEF}181.6} & 1.58         & 310.4       \\
\rowcolor[HTML]{EFEFEF} 
\multicolumn{1}{l|}{\multirow{-2}{*}{\cellcolor[HTML]{EFEFEF}MacTok+SiT-XL$^\ddagger$}}       & \multirow{-2}{*}{\cellcolor[HTML]{EFEFEF}675M} & \multirow{-4}{*}{\cellcolor[HTML]{EFEFEF}KL} & \multirow{-4}{*}{\cellcolor[HTML]{EFEFEF}1D} & \multirow{-4}{*}{\cellcolor[HTML]{EFEFEF}176M} & 128                         & \multicolumn{1}{c|}{\cellcolor[HTML]{EFEFEF}0.43} & 2.82  & \multicolumn{1}{c|}{\cellcolor[HTML]{EFEFEF}189.2} & 1.44         & 302.5       \\ \bottomrule
\end{tabular}
}
\end{table*}
\subsection{Experiments Setup}
\textbf{Implementation Details of Our Method.}
By default, MacTok adopts a ViT-Base backbone for both the encoder and decoder, totaling 176M parameters. We use DINOv2~\cite{dinov2} pretrained features and initialize the encoder with DINOv2 weights to inject richer semantic priors into the latent space, following~\cite{softvq-vae}. DINOv2 features are also used to guide the semantic masking process, promoting more robust latent space as shown in~\cref{sec:method3.2}. MacTok is trained on ImageNet~\cite{imagenet} at 256$\times$256 for 250K iterations and 512$\times$512 for 500K iterations. A frozen DINO-S~\cite{caron2021emerging,dinov2} discriminator is used, with DiffAug~\cite{diffaug}, consistency regularization~\cite{consistency_regularization}, and LeCAM~\cite{LeCAM} as in~\cite{var,softvq-vae}. During training, we apply random and semantic masking with equal probability, using $M$ of 70\%. For decoder fine-tuning, the encoder is frozen and the decoder is trained for 10 epochs without mask. Unless otherwise specified, the image token channel dimension in MacTok is set to 32. The loss weights are set to $\lambda_1{=}1.0$, $\lambda_2{=}0.2$, $\lambda_3{=}10^{-6}$, and $\lambda_4{=}0.1$, following common practice. 
More training details are provided in Appendix~\ref{appendix:appendix_tokenizer_training}.


\noindent \textbf{Implementation Details of Generative Modeling.}
For downstream generation, we employ SiT~\cite{SiT} and LightningDiT~\cite{va-vae} due to their strength and flexibility in modeling 1D token sequences. SiT uses a patch size of 1 with absolute positional embeddings, while LightningDiT adopts rotary positional embeddings. In the main experiments, LightningDiT-XL is trained for 400K steps and SiT-XL for 4M steps, compared to 4M steps in REPA~\cite{repa} and 7M steps in the original SiT~\cite{SiT}. For additional experiments, SiT-B is trained for 500K steps. Additional implementation details can are shown in Appendix~\ref{appendix:appendix_generative_training}.

\noindent \textbf{Evaluation.} 
We evaluate reconstruction quality using the reconstruction Fréchet Inception Distance (rFID)~\cite{rFID-gFID}, Peak Signal-to-Noise Ratio (PSNR), and Structural Similarity Index Measure (SSIM) on 50K validation images from ImageNet.
For generation performance, we report the generation FID (gFID)~\cite{rFID-gFID}, Inception Score (IS)~\cite{IS}, and Precision and Recall~\cite{prec-recall} (see Appendix~\ref{appendix:appendix_main_results} for details), both with and without classifier-free guidance (CFG)~\cite{cfg}, following the ADM~\cite{ADM} evaluation protocol and toolkit.

\subsection{Main Results}
\begin{table*}[t]
\centering
\caption{
\textbf{System-level comparison on ImageNet 512$\times$512 conditional generation.}
SiT-XL trained with MacTok achieves state-of-the-art generation performance using only 64 and 128 tokens ($\dagger$: Large decoder for fair comparison; $\ddagger$: relies on pretrained vision models).
}
\resizebox{0.92\linewidth}{!}{
\label{tab:512}
\begin{tabular}{@{}lcccccccccc}
\toprule
\multicolumn{1}{l|}{}                                                        &                                                &                                              &                                              &                                &                             & \multicolumn{1}{c|}{}                             & \multicolumn{2}{c|}{w/o CFG}                               & \multicolumn{2}{c}{w/ CFG}     \\
\multicolumn{1}{@{}l|}{\multirow{-2}{*}{Method}}                                & \multirow{-2}{*}{\# Params (G)}                 & \multirow{-2}{*}{Tok. Model}                 & \multirow{-2}{*}{Token Type}                  & \multirow{-2}{*}{\# Params (T)\phantom{$^\dagger$}} & \multirow{-2}{*}{\# Tokens↓} & \multicolumn{1}{c|}{\multirow{-2}{*}{Tok. rFID↓}} & gFID↓ & \multicolumn{1}{c|}{IS↑}                           & gFID↓         & IS↑            \\ \midrule
\textit{GAN}                                                                 &                                                &                                              &                                              &                                &                             &                                                   &       &                                                    &               &                \\
\multicolumn{1}{l|}{BigGAN~\cite{maskgit}}                                                  & -                                              & -                                            & -                                            & -\phantom{$^\dagger$}                              & -                           & \multicolumn{1}{c|}{-}                            & -     & \multicolumn{1}{c|}{-}                             & 8.43          & 177.9          \\
\multicolumn{1}{l|}{StyleGAN-XL~\cite{stylegan}}                                             & 168M                                           & -                                            & -                                            & -\phantom{$^\dagger$}                              & -                           & \multicolumn{1}{c|}{-}                            & -     & \multicolumn{1}{c|}{-}                             & 2.41          & 267.7          \\ \midrule
\textit{Auto-regressive}                                                     &                                                &                                              &                                              &                                &                             &                                                   &       &                                                    &               &                \\
\multicolumn{1}{l|}{MaskGIT~\cite{maskgit}}                                                 & 227M                                           & VQ                                           & 2D                                           & 66M\phantom{$^\dagger$}                            & 1024                        & \multicolumn{1}{c|}{1.97}                         & 7.32  & \multicolumn{1}{c|}{156.0}                           & -             & -              \\
\multicolumn{1}{l|}{MAGVIT-v2~\cite{MAGVIT-v2}}                                               & 307M                                           & LFQ                                          & 2D                                           & 116M\phantom{$^\dagger$}                           & 1024                        & \multicolumn{1}{c|}{-}                            & -     & \multicolumn{1}{c|}{-}                             & 1.91          & 324.3          \\
\multicolumn{1}{l|}{MAR-H~\cite{MAR}}                                                   & 943M                                           & KL                                           & 2D                                           & 66M\phantom{$^\dagger$}                            & 1024                        & \multicolumn{1}{c|}{-}                            & 2.74  & \multicolumn{1}{c|}{205.2}                         & 1.73          & 279.9          \\
\multicolumn{1}{l|}{TiTok-B-128~\cite{TiTok}}                                             & 177M                                           & VQ                                           & 1D                                           & 202M\phantom{$^\dagger$}                           & 128                         & \multicolumn{1}{c|}{1.52}                         & -     & \multicolumn{1}{c|}{-}                             & 2.13          & 261.2          \\ 
\multicolumn{1}{l|}{TiTok-L-64~\cite{TiTok}}                                             & 177M                                           & VQ                                           & 1D                                           & 644M\phantom{$^\dagger$}                           & 64                         & \multicolumn{1}{c|}{1.77}                         & -     & \multicolumn{1}{c|}{-}                             & 2.74          & 221.1          \\ \midrule
\textit{Diffusion-based}                                                     & \multicolumn{1}{l}{\textit{}}                  &                                              &                                              &                                &                             &                                                   &       &                                                    &               &                \\
\multicolumn{1}{l|}{ADM~\cite{ADM}}                                                     & -                                              & -                                            & -                                            & -                              & -                           & \multicolumn{1}{c|}{-}                            & 23.24 & \multicolumn{1}{c|}{58.1}                         & 3.85          & 221.7          \\
\multicolumn{1}{l|}{U-ViT-H/4~\cite{U-ViT}}                                               & 501M                                           &                                              & 2D                                           &                                &                             & \multicolumn{1}{c|}{}                             & -     & \multicolumn{1}{c|}{-}                             & 4.05          & 263.8          \\
\multicolumn{1}{l|}{DiT-XL/2~\cite{DiT}}                                                & 675M                                           &                                              & 2D                                           &                                &                             & \multicolumn{1}{c|}{}                             & 9.62  & \multicolumn{1}{c|}{121.5}                         & 3.04          & 240.8          \\
\multicolumn{1}{l|}{SiT-XL/2~\cite{SiT}}                                                & 675M                                           &                                              & 2D                                           &                                &                             & \multicolumn{1}{c|}{}                             & -     & \multicolumn{1}{c|}{-}                             & 2.62          & 252.2          \\
\multicolumn{1}{l|}{DiT-XL~\cite{DiT}}                                                  & 675M                                           &                                              & 2D                                           &                                &                             & \multicolumn{1}{c|}{}                             & 9.56  & \multicolumn{1}{c|}{-}                             & 2.84          & -              \\
\multicolumn{1}{l|}{UViT-H~\cite{U-ViT}}                                                  & 501M                                           & \multirow{-5}{*}{\text{KL}}                        & 2D                                           & \multirow{-5}{*}{84M\phantom{$^\dagger$}}          & \multirow{-5}{*}{4096}      & \multicolumn{1}{c|}{\multirow{-5}{*}{0.62}}       & 9.83  & \multicolumn{1}{c|}{-}                             & 2.53          & -              \\
\multicolumn{1}{l|}{UViT-H~\cite{U-ViT}}                                                  & 501M                                           &                                              & 2D                                           &                                &                             & \multicolumn{1}{c|}{}                             & 12.26 & \multicolumn{1}{c|}{-}                             & 2.66          & -              \\
\multicolumn{1}{l|}{UViT-2B~\cite{U-ViT}}                                                 & 2B                                             & \multirow{-2}{*}{\text{AE}}                        & 2D                                           & \multirow{-2}{*}{323M\phantom{$^\dagger$}}         & \multirow{-2}{*}{256}       & \multicolumn{1}{c|}{\multirow{-2}{*}{0.22}}       & 6.50   & \multicolumn{1}{c|}{-}                             & 2.25          & -              \\
\multicolumn{1}{l|}{TexTok-128~\cite{TexTok}}                                              & 675M                                           & KL                                           & 1D                                           & 176M\phantom{$^\dagger$}                           & 128                         & \multicolumn{1}{c|}{0.97}                         & -     & \multicolumn{1}{c|}{-}                             & 1.80          & 305.4          \\
\multicolumn{1}{l|}{MAETok$^\ddagger$~\cite{MAETok}}                                                  & 675M                                           & AE                                           & 1D                                           & 176M\phantom{$^\dagger$}                           & 128                         & \multicolumn{1}{c|}{0.62}                         & 2.79  & \multicolumn{1}{c|}{204.3}                         & 1.69          & 304.2          \\
\multicolumn{1}{l|}{SoftVQ-VAE$^\ddagger$~\cite{softvq-vae}}                                              & 675M                                           & SoftVQ                                       & 1D                                           & 391M\phantom{$^\dagger$}                           & 64                          & \multicolumn{1}{c|}{0.71}                         & 7.96  & \multicolumn{1}{c|}{133.9}                         & 2.21          & 290.5          \\ \midrule
\textit{Ours}                                                                &                                                &                                              &                                              &                                &                             &                                                   &       &                                                    &               &                \\
\rowcolor[HTML]{EFEFEF} 
\multicolumn{1}{l|}{\cellcolor[HTML]{EFEFEF}}                                & \cellcolor[HTML]{EFEFEF}                       & \cellcolor[HTML]{EFEFEF}                     & \cellcolor[HTML]{EFEFEF}                     & 391M$^\dagger$                           & 64                          & \multicolumn{1}{c|}{\cellcolor[HTML]{EFEFEF}0.89} & 4.63  & \multicolumn{1}{c|}{\cellcolor[HTML]{EFEFEF}163.7} & \textbf{1.52} & 306.0          \\
\rowcolor[HTML]{EFEFEF} 
\multicolumn{1}{l|}{\multirow{-2}{*}{\cellcolor[HTML]{EFEFEF}MacTok+SiT-XL$^\ddagger$}} & \multirow{-2}{*}{\cellcolor[HTML]{EFEFEF}675M} & \multirow{-2}{*}{\cellcolor[HTML]{EFEFEF}KL} & \multirow{-2}{*}{\cellcolor[HTML]{EFEFEF}1D} & 176M\phantom{$^\dagger$}                           & 128                         & \multicolumn{1}{c|}{\cellcolor[HTML]{EFEFEF}0.79} & 5.12  & \multicolumn{1}{c|}{\cellcolor[HTML]{EFEFEF}156.3} & \textbf{1.52} & \textbf{316.0} \\ \bottomrule
\end{tabular}
}
\end{table*}
\textbf{Generation.}
We evaluate SiT-XL and LightningDiT trained with MacTok using \textbf{64 and 128 tokens} on ImageNet at 256$\times$256 and 512$\times$512 resolutions, respectively. Their performance is compared against state-of-the-art (SOTA) generative models. Both LightningDiT-XL and SiT-XL trained with MacTok variants show substantial improvements in generation quality, surpassing SiT-XL/2 with 1024 tokens without CFG, and outperforming other tokenizers with CFG under the same token length. 
At 256$\times$256 resolution, MacTok surpasses SoftVQ-VAE~\cite{softvq-vae} by \textbf{2.21} gFID using 64 tokens without CFG and achieves a gFID of \textbf{1.44} using 128 tokens with CFG, comparable to the state of the art. While LightningDiT-XL produces slightly lower quality than SiT-XL, it still outperforms other baselines. With CFG applied, SiT-XL with MacTok using 128 tokens achieves a new SOTA of \textbf{1.52} gFID and \textbf{316.0} IS on the 512 benchmark. Interestingly, MacTok with 64 tokens performs even better than 128 tokens without CFG at 512 resolution, mainly due to the larger decoder used for fair comparison with SoftVQ-VAE. It outperforms SoftVQ-VAE by \textbf{0.69} gFID using 64 tokens and surpasses MAETok~\cite{MAETok} with CFG using 128 tokens. These results demonstrate that MacTok effectively mitigates posterior collapse in KL-based tokenizers, while maintaining strong generation fidelity. 
We present representative samples across different resolutions in \cref{fig:256_512_vis}, with additional visual results provided in Appendix~\ref{appendix:appendix_generation_vis}.

\noindent \textbf{Reconstruction.} 
MacTok also exhibits strong reconstruction performance while using substantially fewer tokens. It achieves rFID scores of 0.75 and 0.43 with 64 and 128 tokens on the 256 benchmark, and 0.89 and 0.79 on the 512 benchmark. These results outperform VQ-based tokenizers that typically require at least 256 tokens~\cite{ViT-VQGAN, rq-vae, maskbit}. 
Moreover, MacTok achieves competitive results compared to KL-based tokenizers used in diffusion-based models while requiring up to \textbf{64×} fewer tokens. The superior performance with such compact representations highlights MacTok’s ability to learn latents rich in semantic information, maintaining fidelity for downstream generative modeling despite the significantly reduced token count.
Comprehensive reconstruction samples across varying token numbers, as well as visualization of posterior collapse scenarios, are included in Appendix~\ref{appendix:appendix_recon_vis}.


\subsection{Comparison of Tokenizers}\label{subsec: comparison of tokenizers}

We compare MacTok with several leading continuous tokenizers, including VA-VAE~\cite{va-vae}, MAETok~\cite{MAETok}, SoftVQ-VAE~\cite{softvq-vae}, SD-VAE~\cite{SD-VAE}, MAR-VAE~\cite{MAR}, and \textit{l}-DeTok~\cite{DeTok}. For these experiments, SiT-B is trained for 500K steps, and gFID and IS are evaluated on the 256$\times$256 benchmark under optimal CFG settings. MacTok achieves the better balance between reconstruction quality and token efficiency: with 128 tokens, it reaches rFID~0.43, PSNR~25.03 and SSIM~0.806, surpassing MAETok; with only 64 tokens, it still achieves competitive results, rFID~0.75, PSNR~23.10 and SSIM~0.738, outperforming SoftVQ-VAE. For generation, SiT-B trained with MacTok using 128 tokens achieves a gFID of 3.15, exceeding all other continuous tokenizers.
\begin{table}[t]
    \caption{
    \textbf{Comparison of continuous tokenizers.} MacTok attains a better balance between compression and reconstruction quality, while delivering the best generation performance.
    All generation results are reported with optimal CFG scales.
    }
    \label{tab:tokenizer comparison}
    \centering
    \resizebox{\linewidth}{!}{
\begin{tabular}{cc|ccc|cc}
\toprule
                                                 &                             & \multicolumn{3}{c|}{Tok.} & \multicolumn{2}{c}{SiT-B}      \\
\multirow{-2}{*}{Tokenizer}                      & \multirow{-2}{*}{\#Tokens↓} & rFID↓   & PSNR↑  & SSIM↑  & gFID↓         & IS↑            \\ \midrule
VA-VAE                                           & 256                         & 0.28    & 26.30  & 0.846  & 4.33          & 222.1          \\
MAETok                                           & 128                         & 0.48    & 23.61  & 0.763  & 4.77          & 243.2          \\
SoftVQ-VAE                                       & 64                          & 0.88    & 22.13  & 0.706  & 4.09          & 256.9          \\
SD-VAE                                           & 1024                        & 0.61    & 26.04  & 0.834  & 7.66          & 187.5          \\
MAR-VAE                                          & 256                         & 0.53    & -      & -      & 6.26          & 177.5          \\
\textit{l}-DeTok                                            & 256                         & 0.68    & -      & -      & 5.13          & 207.3          \\
\rowcolor[HTML]{EFEFEF} 
\cellcolor[HTML]{EFEFEF}                         & 64                          & 0.75    & 23.10  & 0.738  & 3.22          & \textbf{262.8} \\
\rowcolor[HTML]{EFEFEF} 
\multirow{-2}{*}{\cellcolor[HTML]{EFEFEF}MacTok} & 128                         & 0.43    & 25.03  & 0.806  & \textbf{3.15} & 258.3          \\ \bottomrule
\end{tabular}
}
\end{table}


\subsection{Latent Space Analysis}

We analyze how MacTok avoids posterior collapse and learns a semantically structured latent space.

\begin{figure}[t]
    \centering
    \begin{subfigure}{0.32\linewidth}
        \centering
        \includegraphics[width=\textwidth]{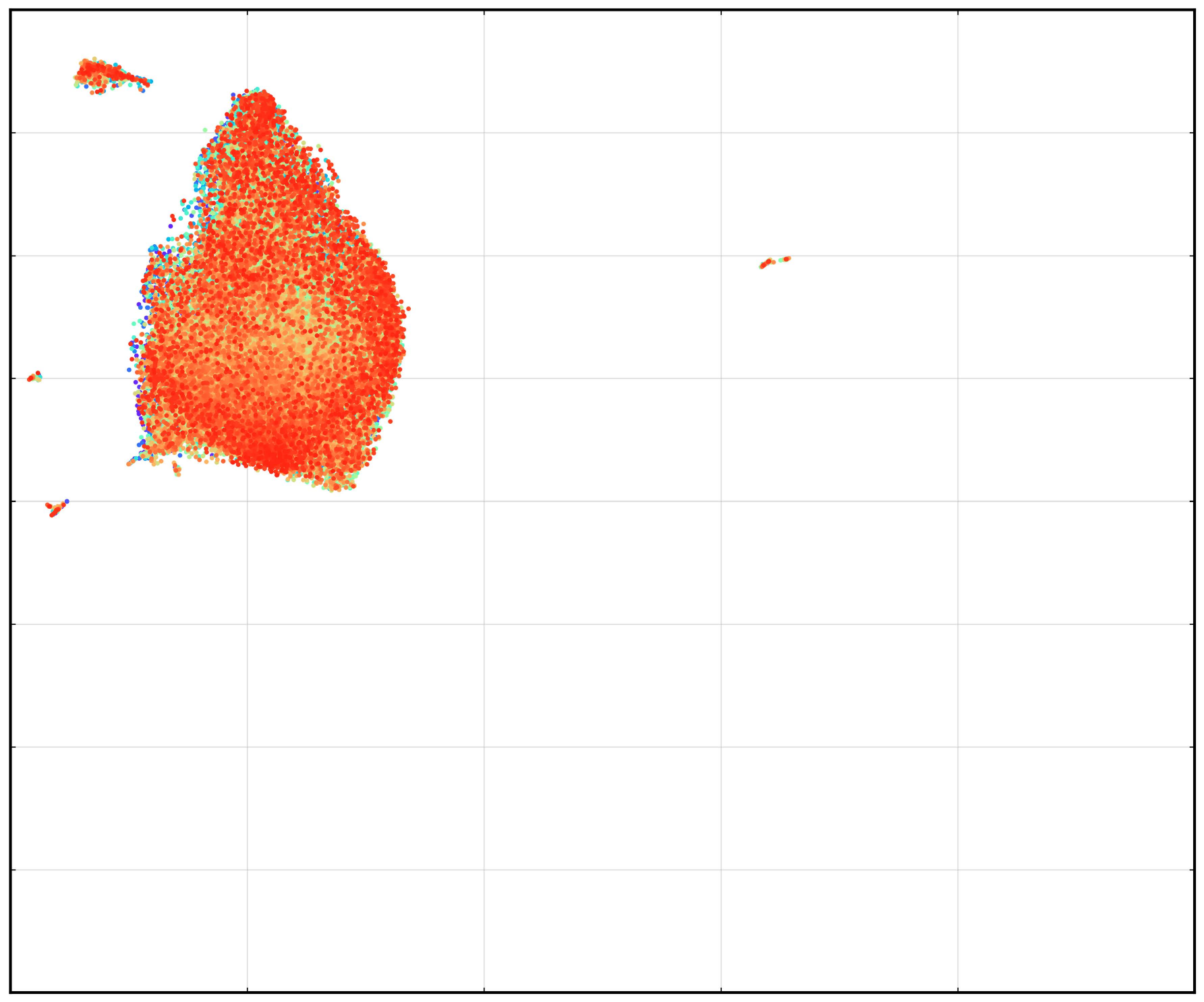} 
        \caption{\centering \scriptsize Collapsed} \label{fig:sub1}
    \end{subfigure}
    \hfill
    \begin{subfigure}{0.32\linewidth}
        \centering
        \includegraphics[width=\textwidth]{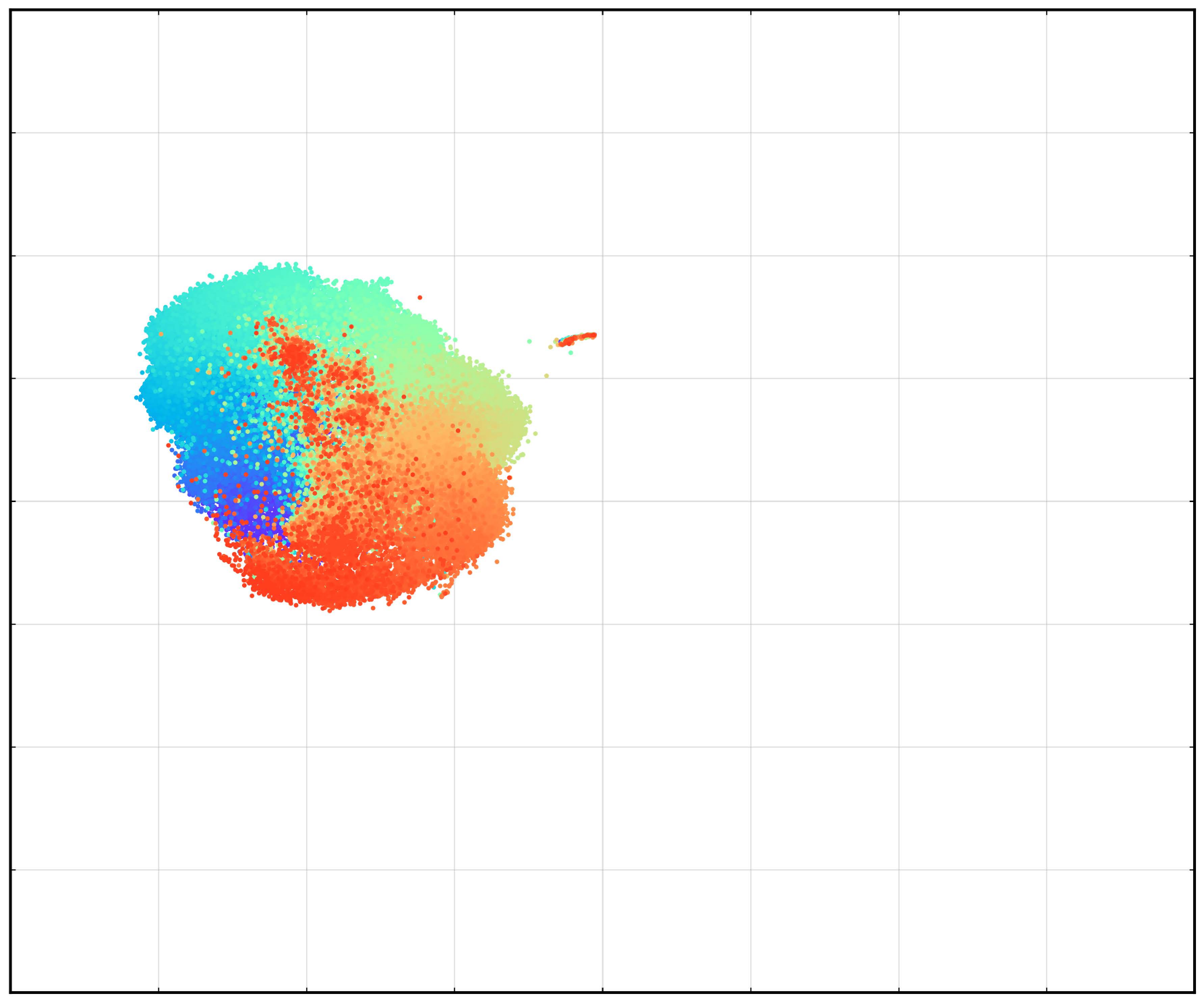} 
        \caption{\centering \scriptsize MacTok-128 w/o RA.} \label{fig:sub2}
    \end{subfigure}
    \hfill
    \begin{subfigure}{0.32\linewidth}
        \centering
        \includegraphics[width=\textwidth]{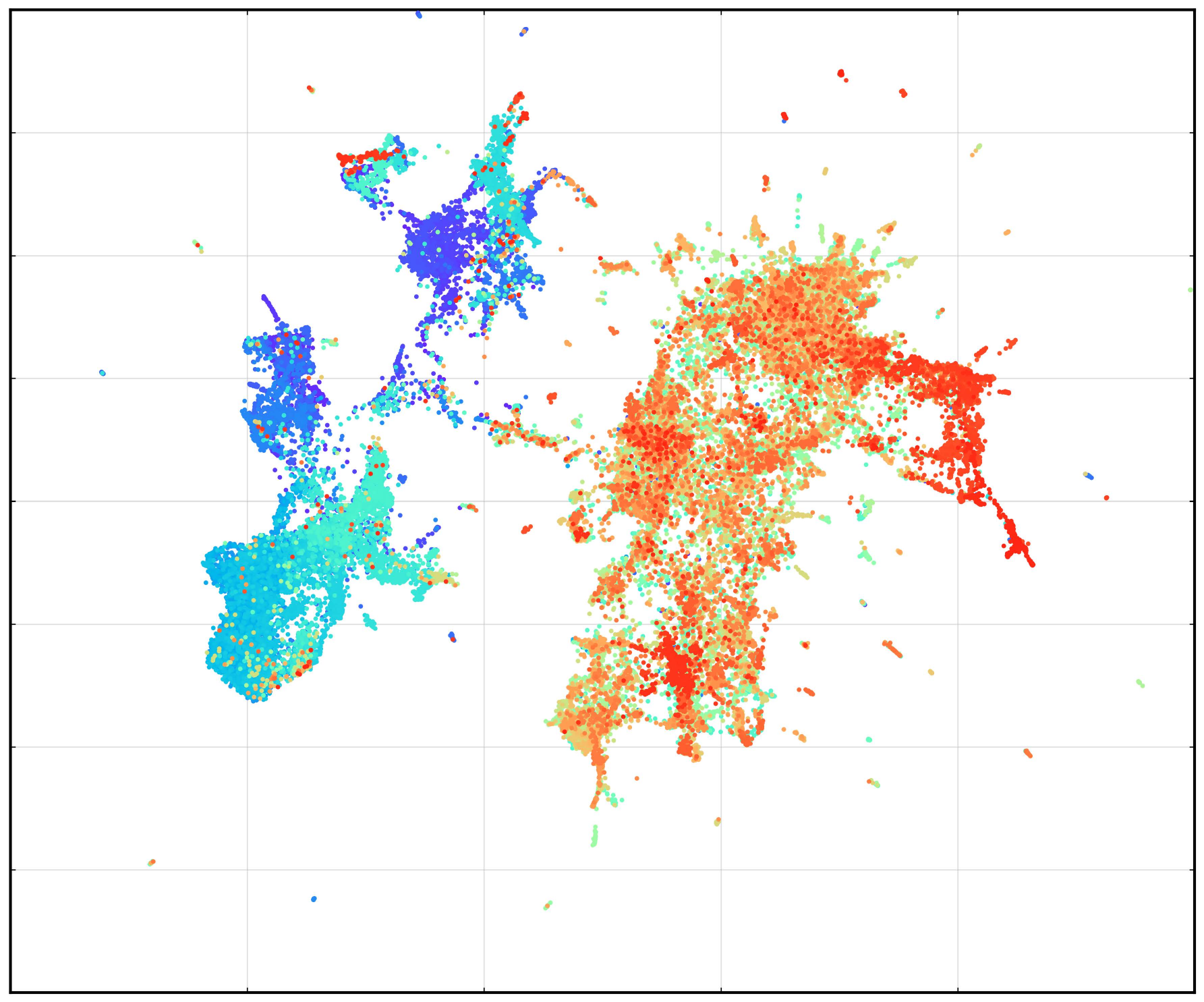} 
        \caption{\centering \scriptsize MacTok-128} \label{fig:sub3}
    \end{subfigure}

    \caption{Visualization of latent space from (a) Collapsed; (b) MacTok-128 trained without representation alignment; (c) MacTok-128}
    \label{fig:latent_vis}
\end{figure}

\begin{figure}[t]
    \begin{subfigure}{0.48\linewidth}
        \centering
        \includegraphics[width=\textwidth]{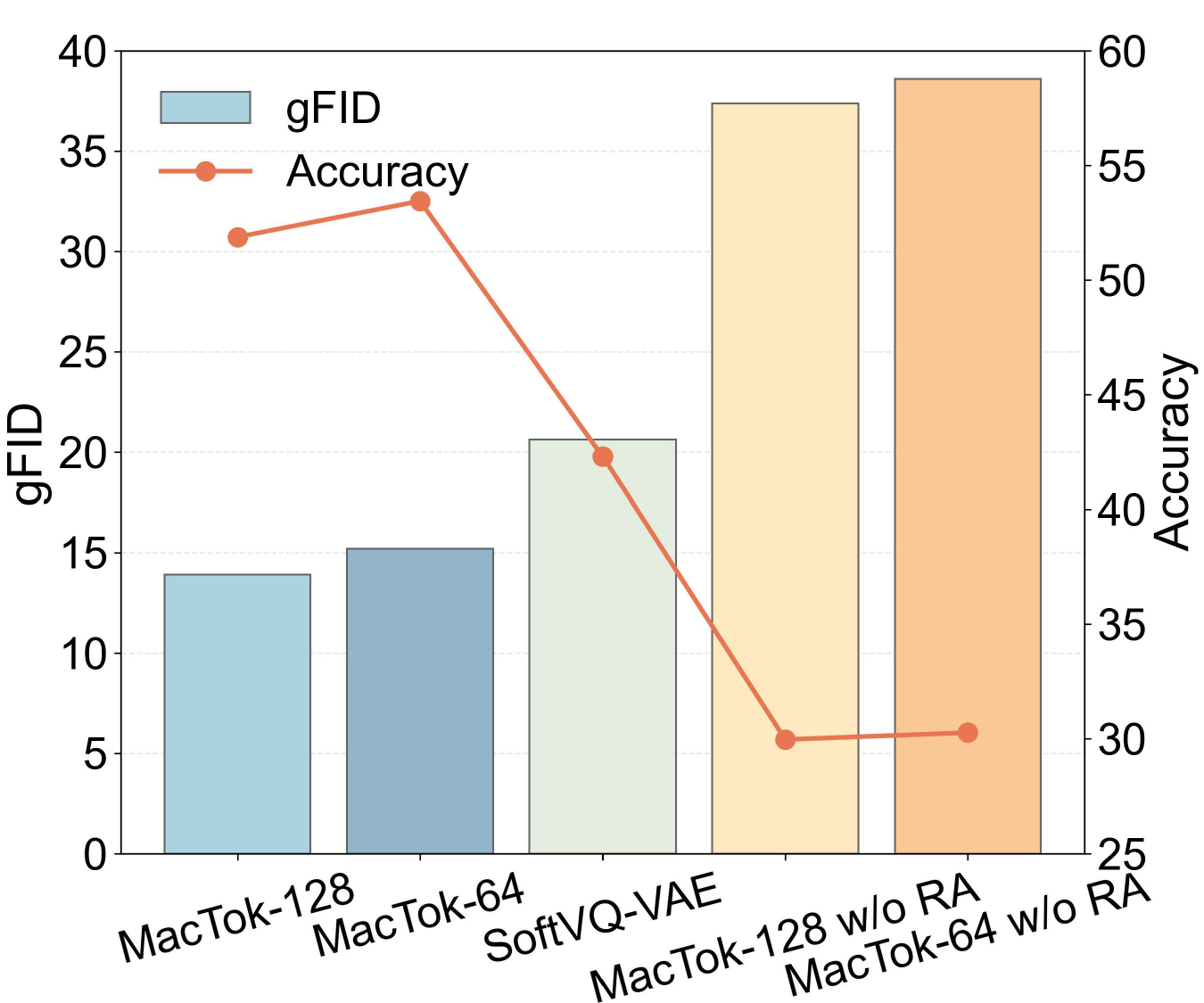} 
        \caption{\centering gFID vs. Accuracy.} \label{fig:accuracy_vs_gfid}
    \end{subfigure}
    \hfill
    \begin{subfigure}{0.48\linewidth}
        \centering
        \includegraphics[width=\textwidth]{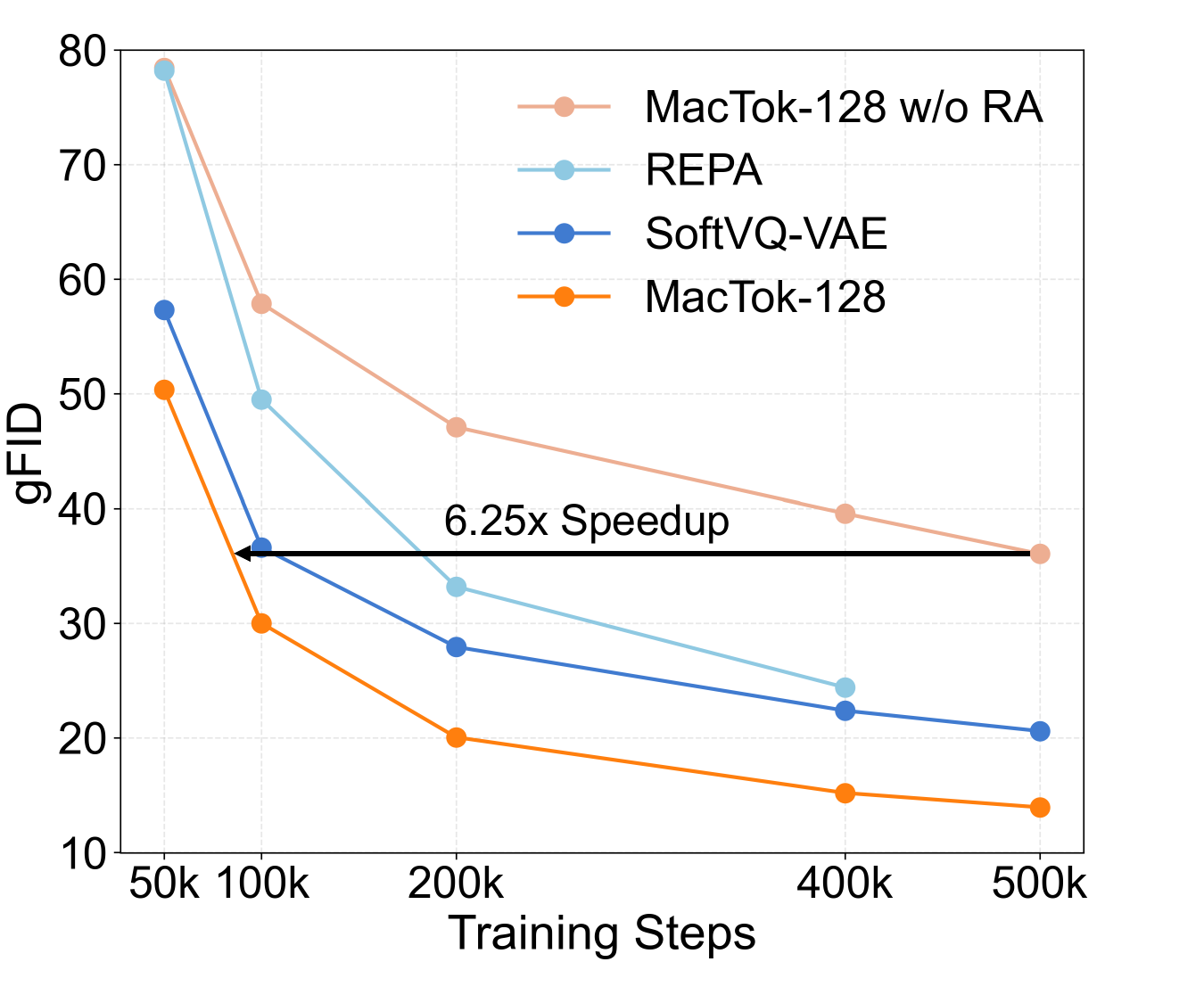} 
        \caption{\centering gFID vs. Training steps.} \label{fig:gfid_curve}
    \end{subfigure}
    \caption{Linear probing accuracy (a) of ImageNet-1k val. and generation performance (b) of MacTok with training steps.}
    \label{fig:linear_probe}
\end{figure}
\noindent \textbf{Latent Space Visualization.} 
\cref{fig:latent_vis} compares three latent spaces:
(a) a collapsed KL-VAE baseline,
(b) MacTok with masking but without representation alignment, and
(c) the full MacTok model.
In (a), the KL-VAE exhibits severe posterior collapse, forming an isotropic and uninformative latent distribution that collapses to the prior, which consequently fails to reconstruct meaningful and recognizable images.
Compared with (c), the latent space in (b) appears more compact and less dispersed across the feature space, as image-level masking imposes an implicit semantic prior that encourages the model to preserve finer visual details and structural information, thereby providing an explanation for MacTok's superior reconstruction performance.
Finally, (c) incorporates global and local representation alignment, resulting in a more well-structured and discriminative latent space where similar semantic concepts cluster together. More visualizations are provided in Appendix~\ref{appendix:appendix_latent_vis}.


\noindent \textbf{Linear Probing and Generation Performance.}
We evaluate latent space quality by correlating linear probing accuracy, which measures how well latent features linearly separate semantic categories, with generative performance.
As shown in \cref{fig:accuracy_vs_gfid}, higher probing accuracy indicates stronger semantic retention and better generation quality.
\cref{fig:gfid_curve} further shows that MacTok not only surpasses other strong baselines in generation fidelity~\cite{repa,softvq-vae}, but also exhibits significantly faster convergence during training.


\subsection{Ablation Studies}
\label{sec:ablation}
\begin{table}[t]

\caption{Ablation on maximum mask ratio $M$ (w/o Decoder Fine-tuning). MacTok is evaluated over mask ratios from 0.4 to 0.8 and different DINO-guided semantic masking settings: ``dino 100\%'' denotes full use of DINO-guided semantic masking, while ``dino 50\%'' applies random and semantic masking with equal probability. Generation performance is reported without CFG.}
\label{tab:Ablation of mask ratio}
\centering
\resizebox{\linewidth}{!}{
\begin{tabular}{c|ccc|cc}
\toprule
                               & \multicolumn{3}{c|}{Tok.} & \multicolumn{2}{c}{SiT-B}      \\
\multirow{-2}{*}{$M$} & rFID↓   & PSNR↑  & SSIM↑  & gFID↓          & IS↑           \\ \midrule
0.4                            & 0.49    & 24.95  & 0.809  & 15.49          & 78.6          \\
0.5                            & 0.54    & 25.02  & 0.812  & 14.87          & 81.1          \\
0.6                            & 0.57    & 24.95  & 0.808  & 14.79          & 81.8          \\
0.7                            & 0.56    & 24.93  & 0.808  & 14.59          & 82.5          \\
0.8                            & 0.59    & 24.89  & 0.805  & 14.92          & 80.8          \\
0.7+dino 100\%                 & 0.66    & 24.92  & 0.809  & 14.84          & 81.1          \\
\rowcolor[HTML]{EFEFEF} 
0.7+dino 50\%                  & 0.57    & 24.91  & 0.808  & 13.95 & 84.8 \\
\rowcolor[HTML]{EFEFEF} 
+Decoder Fine-tuning                  & \textbf{0.43}    & \textbf{25.03}  & 0.806  & \textbf{13.90} & \textbf{85.1} \\ \bottomrule
\end{tabular}
}
\end{table}
We conduct ablation studies to analyze the effect of key design choices in MacTok. Unless otherwise noted, experiments use MacTok-128 with SiT-B trained for 500K steps.

\noindent \textbf{Mask Ratio.}
As shown in \cref{tab:Ablation of mask ratio}, the gFID initially decreases and then increases as the $M$ grows.
A moderate $M$ of 70\% achieves the best generation performance, indicating that stronger masking enhances the robustness and information richness of latent representations.
Applying random and semantic masking with equal probability further improves generation quality.
Although stronger masking slightly reduces reconstruction fidelity, this degradation can be mitigated through decoder fine-tuning (see Appendix~\ref{appendix:appendix_ablation} for more details), which restores image quality while preserving the learned semantic structure.
\noindent \textbf{Key Modules.} 
\cref{tab:albation_module} reports the impact of each module sequentially added to MacTok under decoder fine-tuning and optimal CFG.
Random masking mitigates posterior collapse in KL-based tokenizers.
Local alignment improves both reconstruction and generation by imposing structured organization in the latent space.
DINO-guided semantic masking strengthens semantic robustness and improves gFID and IS.
Global alignment further enforces high-level semantic consistency through effective regularization.
Combining all modules yields the best overall performance.
\begin{table}[t]
\caption{
Ablation of different modules (w/ Decoder Fine-tuning). We report the impact of each module on MacTok's reconstruction and generation performance with optimal CFG scales.}
\label{tab:albation_module}
\centering
\resizebox{\linewidth}{!}{%

\begin{tabular}{l|ccc|cc}
\toprule
Setup              & \multicolumn{3}{c|}{Tok}                        & \multicolumn{2}{c}{SiT-B}      \\ \midrule
MacTok              & rFID↓         & PSNR↑          & SSIM↑          & gFID↓         & IS↑            \\ \midrule
+ random mask      & 0.58          & 24.30           & 0.779          & 6.01          & 234.8          \\
+ local alignment  & 0.44          & 25.02          & 0.806          & 3.53          & 241.9          \\
+ semantic mask       & 0.43          & 24.97          & 0.805          & 3.32           & 249.2            \\
\rowcolor[HTML]{EFEFEF}
+ global alignment & \textbf{0.43} & \textbf{25.03} & \textbf{0.806} & \textbf{3.15} & \textbf{258.3} \\ \bottomrule
\end{tabular}
}
\end{table}

\section{Conclusion}
\label{sec:conclusion}
We introduced MacTok, a continuous tokenizer driven by masking, which effectively mitigates posterior collapse and achieves efficient and high-fidelity image tokenization.
By combining random and DINO-guided semantic masking, MacTok learns robust and semantically structured latent representations, enabling strong generation and reconstruction with only 64 or 128 tokens.
Our findings demonstrate that posterior collapse in continuous tokenizers can be mitigated through masking, and learning a more discriminative latent space is key to advancing generative modeling.

{
    \small
    \bibliographystyle{ieeenat_fullname}
    \bibliography{main}
}



\clearpage
\appendix 

\section{Additional Theoretical and Empirical Analysis}

\subsection{KL-VAE Formulation}
\label{appendix:appendix_kl_vae}
In this section, we provide a detailed description of KL-VAE~\cite{kl-vae,beta-vae}. KL-VAE models both the prior and posterior distributions as Gaussians. Specifically, the prior $p(z)$ is defined as an isotropic unit Gaussian $\mathcal{N}(0, \mathbf{I})$. The posterior distribution $q_\phi(z|x)$ is parameterized by an encoder that predicts the mean $\mu_\phi(x)$ and variance $\sigma^2_\phi(x)$. Using the reparameterization trick, the latent variable $z$ is obtained as
\begin{equation}
\begin{aligned}
    q_\phi(z|x) &= \mathcal{N}(z; \mu_\phi(x), \sigma^2_\phi(x)), \\
    z &= \mu_\phi(x) + \sigma_\phi(x) \odot \epsilon, \quad \epsilon \sim \mathcal{N}(0, \mathbf{I}).
\end{aligned}
\end{equation}
The KL divergence between the posterior and the prior is given by
\begin{equation}
\begin{aligned}
    & \mathcal{L}_{\text{KL}}(q_\phi(z) \| p(z)) \\
    &= \int q_\phi(z|x) \big( \log q_\phi(z|x) - \log p(z) \big) \, dz \\
    &= -\frac{1}{2} \sum_{i=1}^D \big( 1 + \log(\sigma^2_i) - \mu_i^2 - \sigma^2_i \big),
\end{aligned}
\end{equation}
where $D$ denotes the dimensionality of the latent space.The KL term plays a crucial role in the overall training objective, i.e., the Evidence Lower Bound (ELBO). Specifically, it acts as a regularizer that enforces the learned posterior $q_\phi(z|x)$ to stay close to the prior $p(z)$, thereby encouraging smooth and continuous representations.

\subsection{Mitigating Posterior Collapse via Masked Reconstruction}
\label{appendix:theory}

\subsubsection{Corrupted Evidence Lower Bound (ELBO)}
Standard VAE training optimizes the Evidence Lower Bound (ELBO):
\begin{equation}
\mathcal{L}_{\text{ELBO}} = \mathbb{E}_{q_\phi(Z|X)}[\log p_\theta(X|Z)] - \beta \cdot \mathrm{KL}(q_\phi(Z|X) \| p(Z)),
\end{equation}
which balances reconstruction (first term) against regularization of the posterior $q_\phi(Z|X)$ toward the prior $p(Z)$ (second term).
Under strong compression and large $\beta$, this KL penalty can push $q_\phi(Z|X)$ too close to $p(Z)$, causing \emph{posterior collapse}: $q_\phi(Z|X) \approx p(Z)$.
At this point, $Z$ carries no information about $X$, and the decoder effectively becomes an unconditional model $p_\theta(X)$, leading to poor reconstructions.

MacTok takes a different approach by training on \emph{masked} images.
Let $\tilde{X}$ be the masked image after applying a stochastic masking operation $C_m(\tilde{X}|X)$ with ratio $m$.
The encoder sees only $\tilde{X}$, but the decoder must still reconstruct the full image $X$.
This gives us the \emph{corrupted ELBO}:
\begin{equation}
\begin{aligned}
    \mathcal{L}_{\text{corrupted}} = \mathbb{E}_{X, \tilde{X} \sim C_m(\cdot|X)} [ & \mathbb{E}_{q_\phi(Z|\tilde{X})}[-\log p_\theta(X|Z)] \\
    &+ \beta \cdot \mathrm{KL}(q_\phi(Z|\tilde{X}) \| p(Z)) ].
\end{aligned}
\label{eq:corrupted_elbo}
\end{equation}
The key difference is this information asymmetry: the encoder only gets partial information $\tilde{X}$, while the decoder has to predict everything, including what was masked.
This forces $Z$ to actually encode useful information from $\tilde{X}$—otherwise the decoder has no way to reconstruct the missing parts.

\subsubsection{Why Collapsed Solutions Become Suboptimal}
Consider what happens when the posterior collapses: $q_\phi(Z|\tilde{X}) = p(Z)$.
Now $Z$ is independent of both $\tilde{X}$ and $X$, so:
\begin{equation}
\begin{aligned}
\mathbb{E}_{q_\phi(Z|\tilde{X})=p(Z)}[-\log p_\theta(X|Z)] &= \mathbb{E}_{Z \sim p(Z)}[-\log p_\theta(X|Z)] \\
&= \mathbb{E}_{Z \sim p(Z)}[-\log p_\theta(X)] \\
& = -\log p_\theta(X), 
\end{aligned}
\end{equation}
where $p_\theta(X)$ is just the unconditional image distribution.

We can break this down by what's visible versus what's masked:
\begin{equation}
-\log p_\theta(X) = -\log p_\theta(X_{\text{visible}}) - \log p_\theta(X_{\text{masked}}).
\end{equation}
The problem is the second term: $-\log p_\theta(X_{\text{masked}})$.
Without any context, the decoder has to guess what's in the masked regions based purely on dataset statistics—maybe ``skies are usually blue'' or ``grass is usually green.''
But this fails for any specific image.
As we mask more pixels (higher $m$), this blind guessing gets worse and $-\log p_\theta(X)$ shoots up.

Compare this to when $Z$ actually encodes information from $\tilde{X}$.
Now the decoder can use contextual clues—if it sees grass and trees in the visible parts, it knows this is probably an outdoor scene; if the visible colors are warm, maybe it's sunset. This capability of recovering latent details from partial or degraded visual cues shares underlying principles with robust image processing pipelines designed for severely suboptimal conditions.
This gives much better predictions:
\begin{equation}
-\log p_\theta(X|Z) = -\log p_\theta(X_{\text{visible}}|Z) - \log p_\theta(X_{\text{masked}}|Z),
\end{equation}
where $-\log p_\theta(X_{\text{masked}}|Z)$ is now significantly smaller because the decoder can make informed guesses based on what $Z$ encoded.

Let's define the benefit of having an informative $Z$ as:
\begin{equation}
\Delta \triangleq -\log p_\theta(X) - \mathbb{E}_{q_\phi(Z|\tilde{X})}[-\log p_\theta(X|Z)].
\label{eq:delta}
\end{equation}
Larger $\Delta$ means $Z$ is more useful.
Now compare total losses:
\begin{align}
\text{Loss}_{\text{collapse}} &= -\log p_\theta(X), \\
\text{Loss}_{\text{informative}} &= \mathbb{E}_{q_\phi(Z|\tilde{X})}[-\log p_\theta(X|Z)] + \beta \cdot \epsilon,
\end{align}
where $\epsilon = \mathrm{KL}(q_\phi(Z|\tilde{X}) \| p(Z)) > 0$ is the KL cost of keeping $Z$ informative.
The informative solution wins when:
\begin{equation}
\Delta > \beta \cdot \epsilon.
\label{eq:condition}
\end{equation}
So the collapsed solution is suboptimal whenever $\beta < \Delta / \epsilon$.

Here's where masking matters: it directly increases $\Delta$.
As we mask more:
\begin{itemize}[leftmargin=*, itemsep=2pt, topsep=2pt]
    \item Without context (collapsed case), predicting more masked pixels becomes exponentially harder, pushing $-\log p_\theta(X)$ way up.
    \item With context from $Z$ (informative case), we can still make reasonable predictions based on visible cues, so $\mathbb{E}[-\log p_\theta(X|Z)]$ stays relatively controlled.
\end{itemize}
Higher $m$ widens the gap $\Delta$, which means informative posteriors stay optimal for a broader range of $\beta$ (Eq.~\ref{eq:condition}).

Without masking, there's a loophole: the decoder can just copy local patterns from the input.
Even if $Z$ is mostly useless, reconstructions still look okay, so $\Delta$ stays small and collapse becomes competitive.
Masking closes this loophole—the decoder \emph{has to} use $Z$ to fill in the missing parts, which keeps information flowing through the latent space even under strong regularization.

In conclusion, masking prevents collapse through a simple mechanism.
First, it makes the reconstruction task harder, so $Z$ needs to be informative.
Second, if $Z$ collapses and becomes useless, the decoder is forced to blindly guess large portions of the image, incurring huge losses.
Third, by increasing $\Delta$, masking ensures that keeping $Z$ informative remains the better strategy across a wide range of $\beta$ values.
This is how MacTok maintains meaningful continuous tokens even with aggressive compression and regularization.
\subsection{Visualization of KL Divergence Dynamics}
\label{appendix:appendix_kl_loss}

\begin{figure}[tp]
    \centering
    \includegraphics[width=\linewidth]{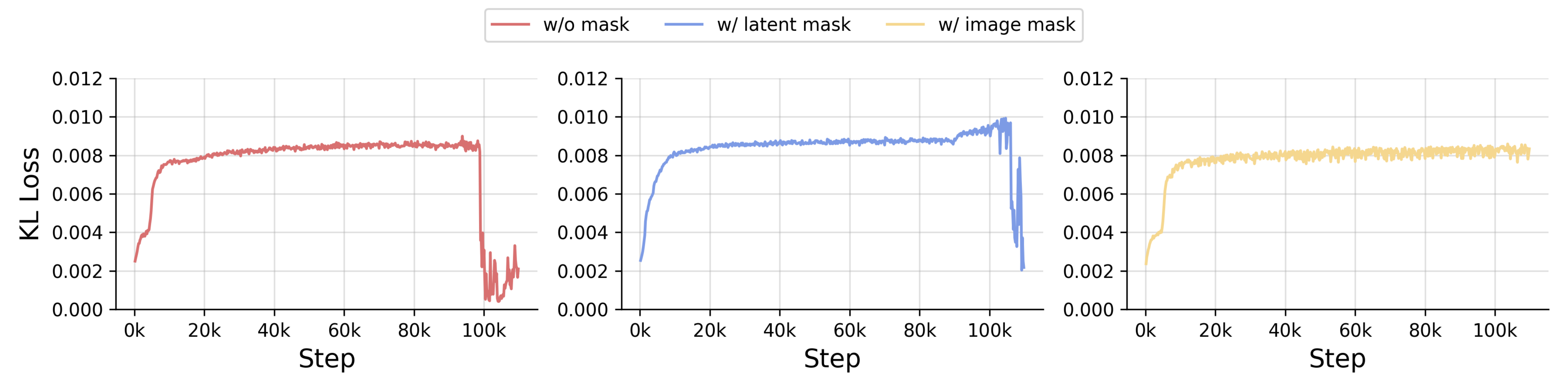}
    \caption{Comparison of different masking strategies of the KL loss curve.}
    \label{fig:comparison of KL loss}
\end{figure}

As illustrated in Fig.~\ref{fig:comparison of KL loss}, applying latent token masking postpones posterior collapse compared to the conventional KL-VAE baseline. Nevertheless, this improvement is transient, as the model ultimately converges to a degenerate solution over the course of training. In contrast, masking image tokens yields a markedly steadier optimization process and produces more resilient latent representations. We attribute this behavior to the fact that image masking encourages both the encoder and decoder to reason over incomplete visual inputs, thereby encouraging the latent space to encode more structural and semantic information.

\section{Additional Implementation Details}

In this section, we present additional implementation details for tokenizer training and downstream generative model training.

\subsection{Implementation Details of MacTok}
\label{appendix:appendix_tokenizer_training}
We train the MacTok tokenizers on ImageNet at resolution of 256$\times$256 for 250K iterations with a batch size of 256 and at 512$\times$512 for 500K iterations with a batch size of 128. Data augmentation includes horizontal flipping and center cropping. We use AdamW optimizer with $\beta_1=0.9,\beta_2=0.95$, a weight decay of $1\times10^{-4}$. The learning rate follows a cosine annealing schedule, peaking at $1\times10^{-4}$ and preceded by a linear warm-up of 5K and 10K steps for the 256 and 512 resolutions.
To improve the stability of adversarial learning, we employ a frozen DINO-S~\cite{caron2021emerging,dinov2} network as the discriminator as in~\cite{softvq-vae,var} and incorporate the adaptive weighting scheme. Moreover, we enhance discriminator training by introducing DiffAug~\cite{diffaug}, consistency regularization~\cite{consistency_regularization}, and LeCAM regularization~\cite{LeCAM}, as used in~\cite{softvq-vae}. The regularization weights for the consistency and LeCAM terms are set to 4.0 and 0.001, respectively.
The overall training objective follows common practice with loss weights $\lambda_1=1.0$, $\lambda_2=0.2$, $\lambda_3=1\times10^{-6}$, and $\lambda_4=0.1$.

\subsection{Implementation Details of Generative Models}
\label{appendix:appendix_generative_training}
\textbf{LightningDiT}~\cite{va-vae} The training configuration of our LightningDiT models closely follows the original setup. As our model operates on 1D latent tokens, we set the patch size to 1. LightningDiT-XL is trained with a constant learning rate of $2\times10^{-4}$  and a global batch size of 1024. We adopt a cosine noise scheduler and rotary positional embeddings, consistent with the original implementation. In the main paper, we report results of LightningDiT-XL trained for 400K iterations. For conditional generation with classifier-free guidance (CFG), we use a guidance scale of 2.5 for LightningDiT models trained on MacTok with 128 tokens and 2.7 for those trained with 64 tokens. These values are selected via grid search based on gFID and IS metrics computed over 10K generated samples.\\
\noindent\textbf{SiT}~\cite{SiT} We follow the original training configuration of SiT, using a constant learning rate of $1\times10^{-4}$ and a global batch size of 256. A linear learning rate scheduler is adopted, as it demonstrates better empirical performance in our setting. The main results are reported after 4M training iterations. For conditional generation with CFG, we set the guidance scale to 2.3 for SiT models trained on MacTok with 128 tokens and 2.4 for those trained with 64 tokens. Following REPA~\cite{repa}, the guidance interval is set to $[0, 0.7]$ for CFG-based results. The optimal values are determined through grid search by evaluating gFID and IS over 10K generated samples.


\section{Additional Results}
In this appendix, we provide supplementary evidence to support the effectiveness of our approach. Specifically, we include further visualizations of the latent token space, more ablation studies, extended quantitative evaluations of generative models trained on MacTok, and additional qualitative examples of reconstructed and generated images. These results complement the main paper by highlighting the structural organization of the latent space, the generative fidelity across different resolutions and token settings.
\subsection{Latent Space Visualization}
\label{appendix:appendix_latent_vis}
\begin{figure}[tp]
    \centering
    \begin{minipage}{0.48\linewidth}
        \centering
        \includegraphics[width=\textwidth]{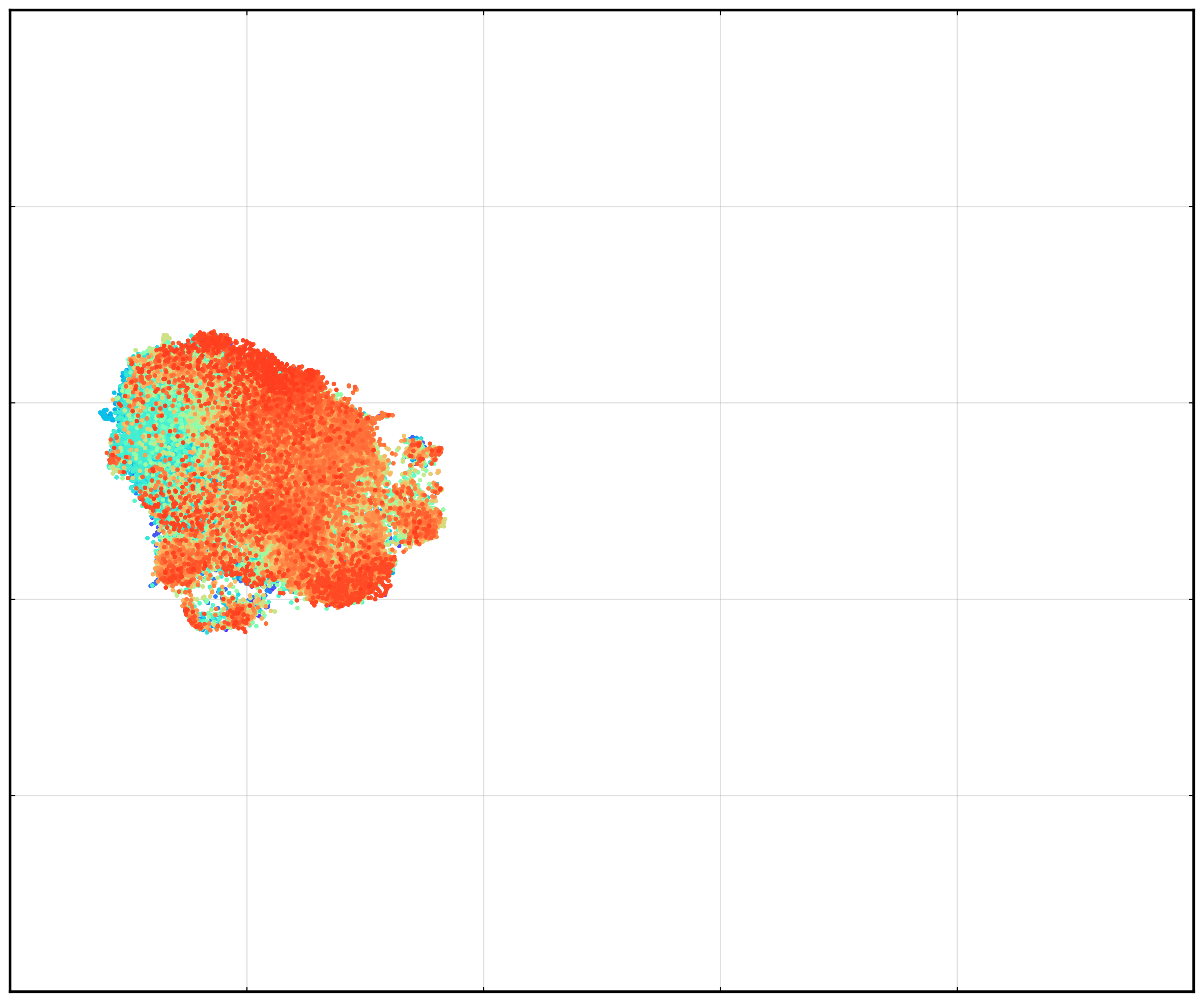} 
        \subcaption{\centering \scriptsize MacTok-64 w/o RA.}
    \end{minipage}
    \begin{minipage}{0.48\linewidth}
        \centering
        \includegraphics[width=\textwidth]{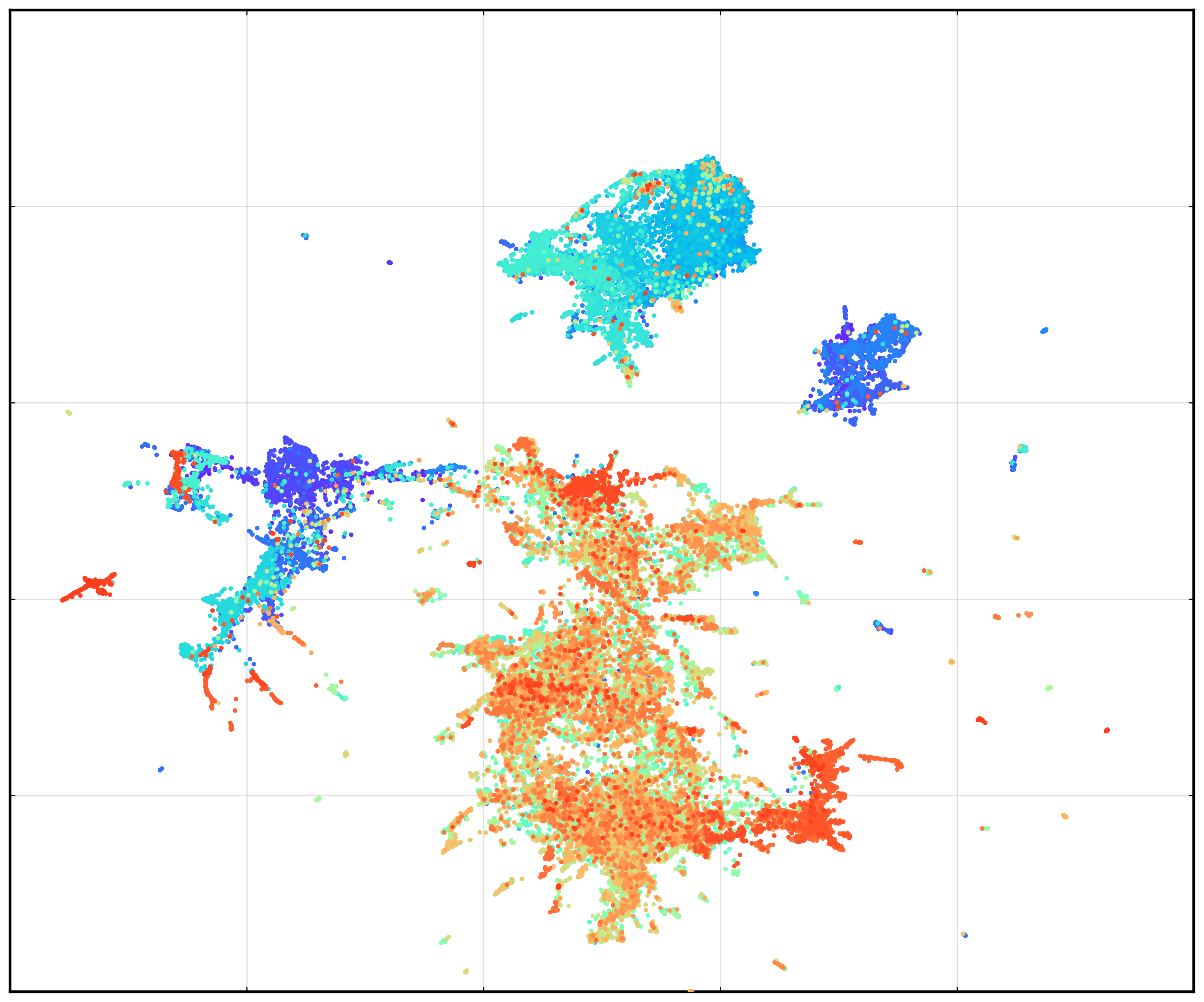} 
        \subcaption{\centering \scriptsize MacTok-64}
    \end{minipage}
    \caption{Visualization of laten space from (a) MacTok-64 trained without Representation alignment; (b) MacTok-64}
    \label{fig:appendix_latent_vis}
\end{figure}

Fig.~\ref{fig:appendix_latent_vis} illustrates the UMAP projection of the latent representations obtained with 64 tokens. We compare the latent space learned by MacTok-64 with and without Representation alignment (RA). As shown, MacTok-64 with Representation alignment generates more structured and separable embeddings compared to the model trained without alignment. This visualization confirms that MacTok effectively organizes the latent space with fewer tokens, supporting downstream tasks such as linear probing and generative modeling, and showing great promise for broader spatial perception applications that require dense structural consistency.

\subsection{Ablation Study}
\label{appendix:appendix_ablation}
\begin{table}[t]
    \centering
    \caption{Ablation studies of decoder fine-tuning and model size, showing their effects on MacTok's performance}
    \begin{subtable}{0.45\linewidth}
        \resizebox{\textwidth}{!}{
        \begin{tabular}{@{}lcc@{}}
        Tokenizer & rFID↓ & gFID↓ \\ \toprule
        MacTok-64  & 0.93  & 3.28  \\
        \rowcolor[HTML]{EFEFEF} 
        +FT       & 0.75  & 3.22  \\
        MacTok-128 & 0.57  & 3.19  \\
        \rowcolor[HTML]{EFEFEF} 
        +FT       & 0.43  & 3.15 
        \end{tabular}}
        \caption{Decoder fine-tuning.}
        \label{tab:Ablation of decoder finetuning}
    \end{subtable}
    \hfill
    \begin{subtable}{0.45\linewidth}
        \resizebox{\textwidth}{!}{
        \begin{tabular}{@{}lcc@{}}
        Model   Size & \#Params             & rFID                 \\ \toprule
        MacTok-S     & 45M                  & 0.78                     \\
        \rowcolor[HTML]{EFEFEF} 
        MacTok-B     & 176M                 & 0.57                 \\
        MacTok-BL    & 391M                 & 0.57                     \\
                     & \multicolumn{1}{l}{} & \multicolumn{1}{l}{}
        \end{tabular}}
        \caption{MacTok model size.}
        \label{tab:Ablation of MacTok model size}
    \end{subtable}

\end{table}
\noindent \textbf{Decoder Fine-tuning.} Tab.~\ref{tab:Ablation of decoder finetuning} reports MacTok’s performance when freezing the encoder and fine-tuning only the decoder without masking. Specifically, the encoder is frozen and the decoder is trained for 10 epochs without mask. This strategy notably improves rFID and slightly enhances gFID, indicating that decoder fine-tuning effectively restores reconstruction quality degraded by high mask ratios while preserving the latent space.\\
\noindent \textbf{Model Size.} Tab.~\ref{tab:Ablation of MacTok model size} evaluates MacTok model size on ImageNet at 256$\times$256. MacTok-B significantly outperforms MacTok-S, whereas further scaling does not yield additional gains. Consequently, MacTok-B is adopted as the default. For 512$\times$512 generation with 64 tokens, we use MacTok-BL to ensure fair comparison with SoftVQ-VAE and mitigate reconstruction degradation at higher resolution.

\subsection{Main Results}
\label{appendix:appendix_main_results}
We present the complete quantitative results, including both precision and recall, for the ImageNet 256$\times$256 and 512$\times$512 benchmarks in Tab.~\ref{tab:appendix_256} and Tab.~\ref{tab:appendix_512}, respectively. All evaluations are conducted on SiT-XL models trained for 4M steps and LightningDiT-XL models trained for 400K steps. Notably, our models achieve state-of-the-art generative performance at 512$\times$512 resolution and deliver results comparable to leading approaches at 256$\times$256 resolution. Moreover, our models exhibit superior conditional gFID scores even without applying classifier-free guidance (CFG), outperforming SoftVQ-VAE~\cite{softvq-vae} and other vanilla generative baselines~\cite{SD-VAE,U-ViT,DiT,SiT,repa} that utilize at least 256 or 1024 tokens. We further include results measured across different training durations, as summarized in Tab.~\ref{tab:appendix_along_training}.

\subsection{Reconstruction Visualization}
\label{appendix:appendix_recon_vis}
\begin{figure*}[tp]
    \centering
    \includegraphics[width=1\linewidth]{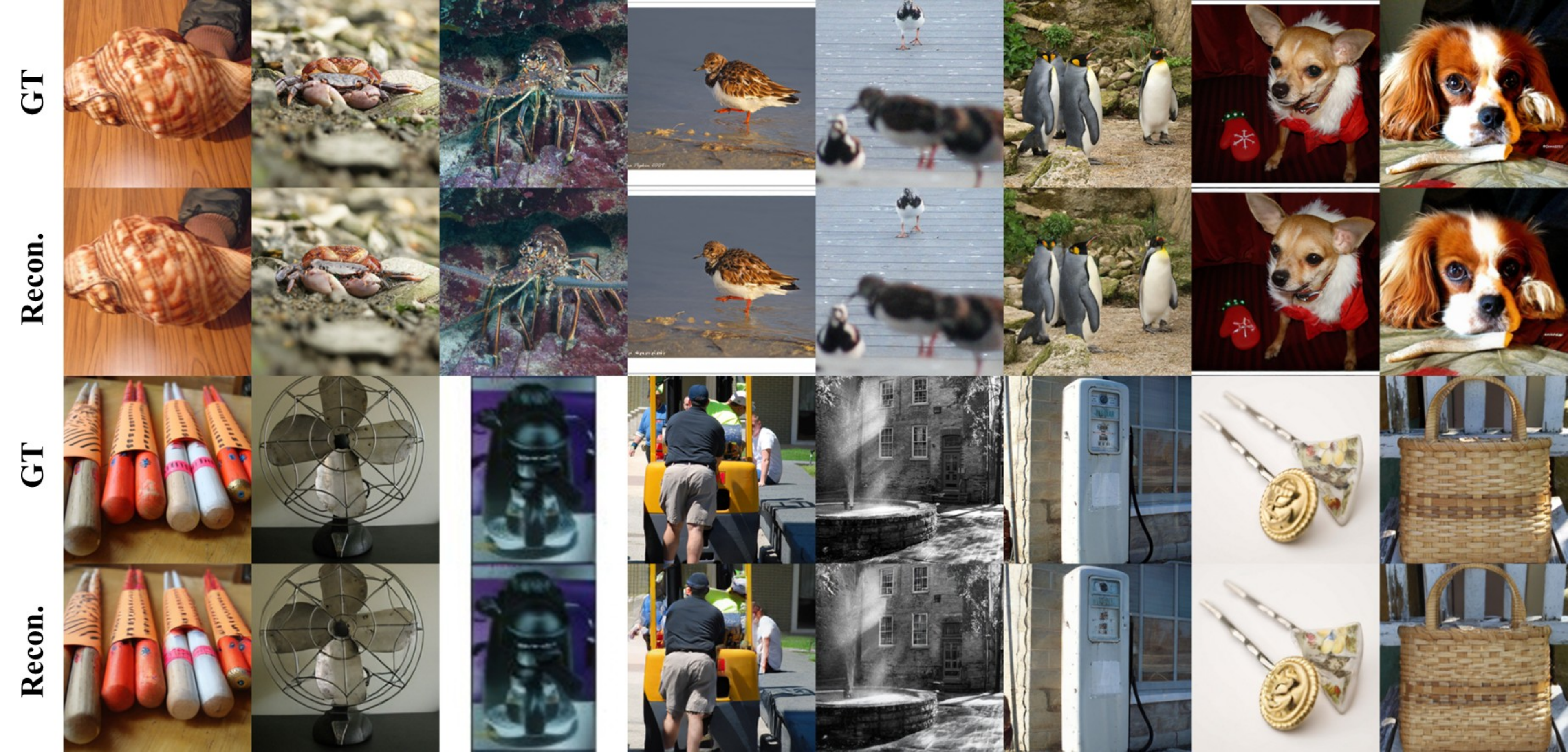}
    \caption{\centering Reconstruction results of MacTok with 64 tokens.}
    \label{fig:recon_64}
\end{figure*}

\begin{figure*}[tp]
    \centering
    \includegraphics[width=1\linewidth]{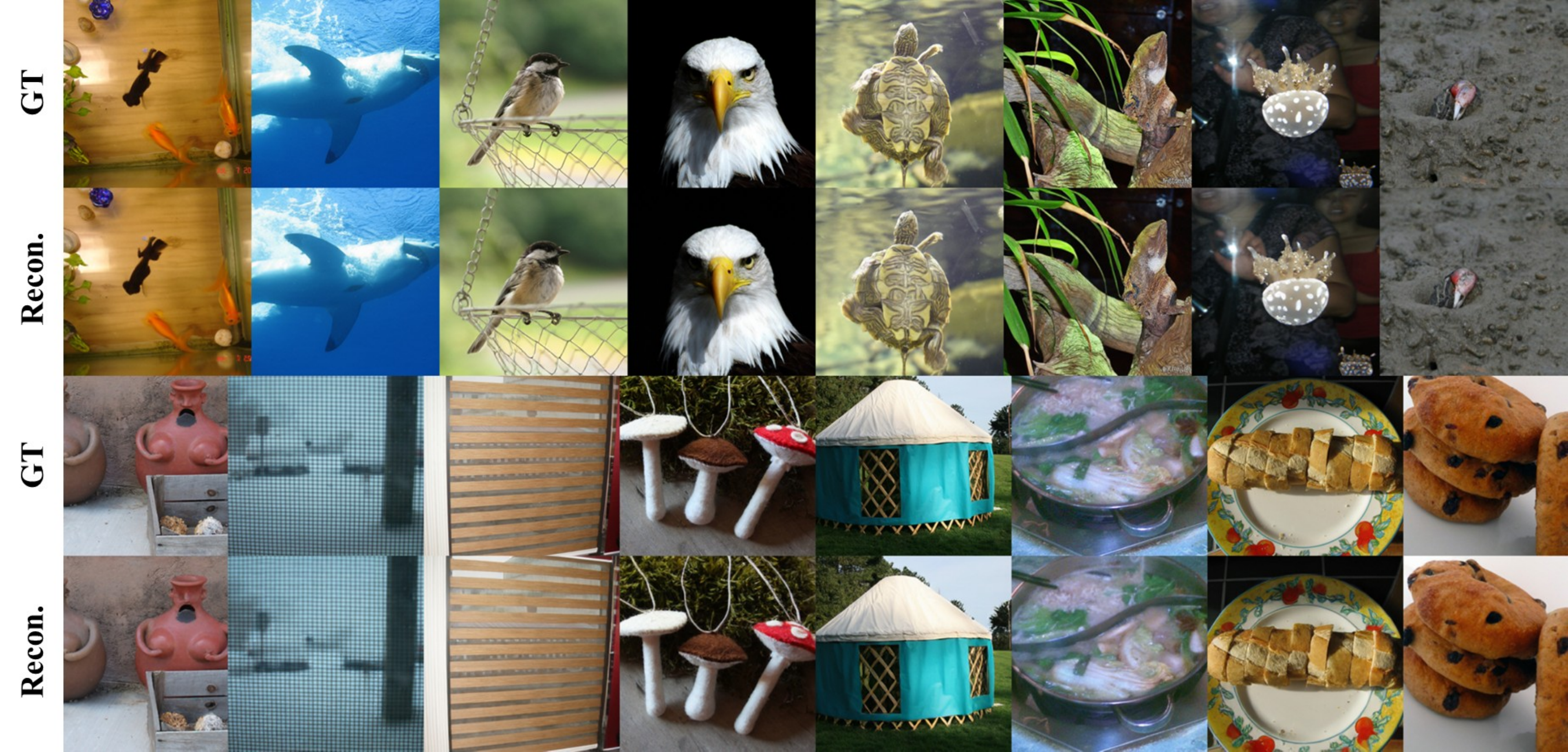}
    \caption{\centering Reconstruction results of MacTok with 128 tokens.}
    \label{fig:recon_128}
\end{figure*}

\begin{figure*}[tp]
    \centering
    \includegraphics[width=1\linewidth]{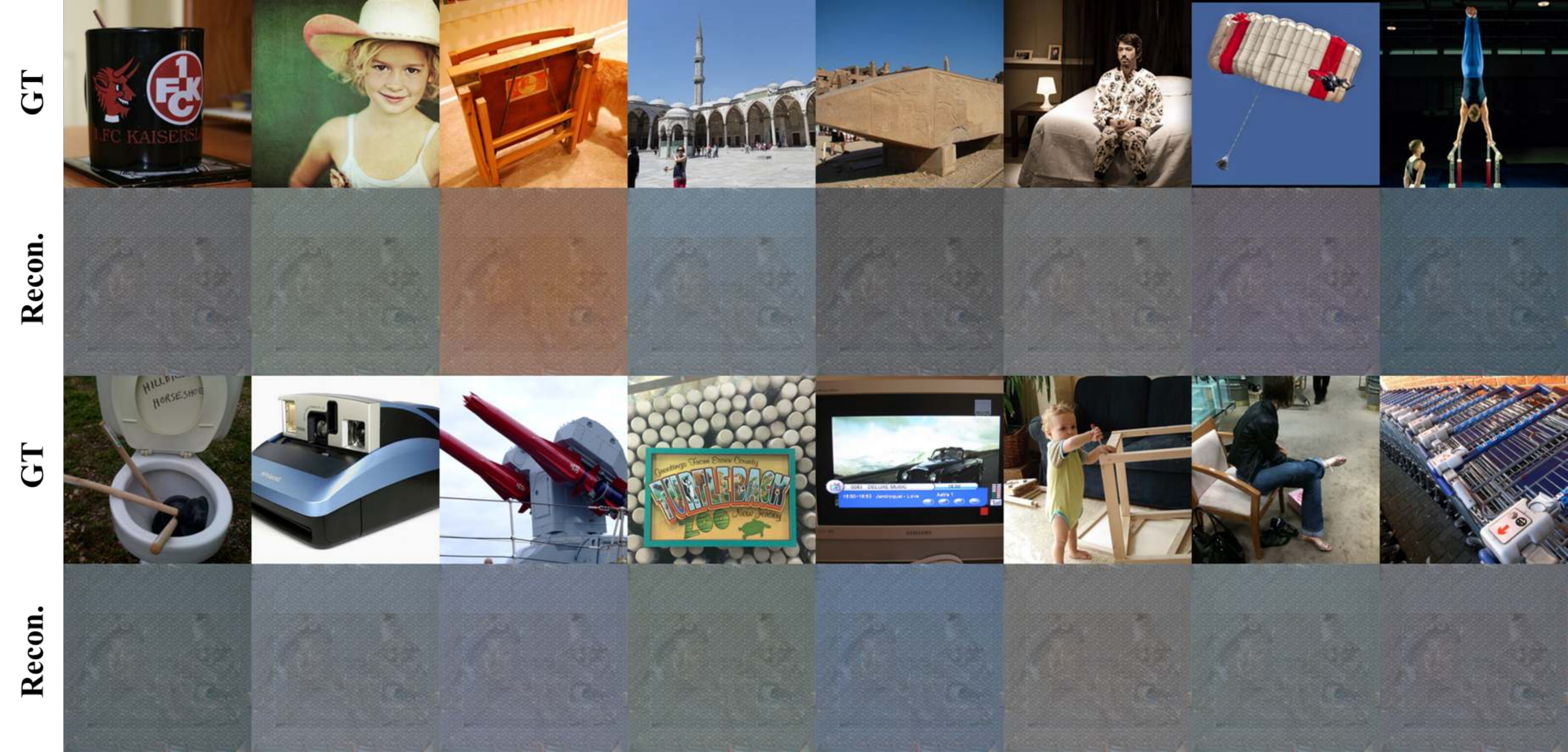}
    \caption{\centering Reconstruction results of collapsed KL-VAE.}
    \label{fig:recon_collapsed}
\end{figure*}

We present the reconstruction results of MacTok using 64 and 128 latent tokens in Fig.~\ref{fig:recon_64} and Fig.~\ref{fig:recon_128}, respectively.
As shown, increasing the number of tokens leads to finer spatial details and improved texture fidelity, demonstrating the scalability of MacTok’s latent representation.
In contrast, reconstructions from collapsed baselines (see Fig.~\ref{fig:recon_collapsed}) fail to recover meaningful visual content, indicating that posterior collapse severely limits the model’s representational capacity.
MacTok’s semantically structured latent space effectively preserves both global layout and local semantics, resulting in faithful and perceptually consistent reconstructions even under limited token budgets.
These visualizations complement the quantitative evaluation in the main paper and further verify the robustness of our latent modeling strategy.

\subsection{Generation Visualization}
\label{appendix:appendix_generation_vis}
More visualizations of LightningDiT-X and SiT-XL trained on MacTok with 64 and 128 tokens are provided here.

\begin{table*}[t]
\caption{
\textbf{System-level comparison} on ImageNet 256$\times$256 conditional generation.
We report both Precision and Recall under classifier-free guidance (CFG) and non-CFG settings. 
``\# Params (G)'' denotes generator parameters; ``Tok. Model'' refers to the tokenizer model type; ``Token Type'' indicates 1D or 2D tokenization; ``\# Params (T)'' denotes tokenizer parameters; and ``\# Tokens'' represents the number of latent tokens.
}
\label{tab:appendix_256}
\centering
\resizebox{1\linewidth}{!}
{
\begin{tabular}{@{}lcccccccccccccc@{}}
\toprule
\multicolumn{1}{l|}{}                                                              &                                                &                                              &                                              &                                                &                             & \multicolumn{1}{c|}{}                             & \multicolumn{4}{c|}{w/o CFG}                                              & \multicolumn{4}{c}{w/ CFG}      \\
\multicolumn{1}{l|}{\multirow{-2}{*}{Method}}                                      & \multirow{-2}{*}{\# Params(G)}                 & \multirow{-2}{*}{Tok. Model}                 & \multirow{-2}{*}{Token Type}                  & \multirow{-2}{*}{\# Params(T)}                 & \multirow{-2}{*}{\#Tokens↓} & \multicolumn{1}{c|}{\multirow{-2}{*}{Tok. rFID↓}} & gFID↓ & IS↑   & Prec↑ & \multicolumn{1}{c|}{Recall↑}                      & gFID↓ & IS↑   & Prec↑ & Recall↑ \\ \midrule
\textit{Auto-regressive}                                                           &                                                &                                              &                                              &                                                &                             &                                                   &       &       &       &                                                   &       &       &       &         \\
\multicolumn{1}{l|}{ViT-VQGAN~\cite{ViT-VQGAN}}                                                     & 1.7B                                           & VQ                                           & 2D                                           & 64M                                            & 1024                        & \multicolumn{1}{c|}{1.28}                         & 4.17  & 175.1 & -     & \multicolumn{1}{c|}{-}                            & -     & -     & -     & -       \\
\multicolumn{1}{l|}{RQ-Trans.~\cite{rq-vae}}                                                     & 3.8B                                           & RQ                                           & 2D                                           & 66M                                            & 256                         & \multicolumn{1}{c|}{3.20}                          & -     & -     & -     & \multicolumn{1}{c|}{-}                            & 3.80  & 323.7 & -     & -       \\
\multicolumn{1}{l|}{MaskGIT~\cite{maskgit}}                                                       & 227M                                           & VQ                                           & 2D                                           & 66M                                            & 256                         & \multicolumn{1}{c|}{2.28}                         & 6.18  & 182.1 & 0.80  & \multicolumn{1}{c|}{0.51}                         & -     & -     & -     & -       \\
\multicolumn{1}{l|}{LlamaGen-3B~\cite{llamagen}}                                                   & 3.1B                                           & VQ                                           & 2D                                           & 72M                                            & 576                         & \multicolumn{1}{c|}{2.19}                         & -     & -     & -     & \multicolumn{1}{c|}{-}                            & 2.18  & 263.3 & 0.80  & 0.58    \\
\multicolumn{1}{l|}{WeTok~\cite{wetok}}                                                         & 1.5B                                           & VQ                                           & 2D                                           & 400M                                           & 256                         & \multicolumn{1}{c|}{0.60}                          & -     & -     & -     & \multicolumn{1}{c|}{-}                            & 2.31  & 276.6 & 0.84  & 0.55    \\
\multicolumn{1}{l|}{VAR~\cite{var}}                                                           & 2B                                             & \text{MSRQ}                                       & 2D                                           & 109M                                           & 680                         & \multicolumn{1}{c|}{0.90}                          & -     & -     & -     & \multicolumn{1}{c|}{-}                            & 1.92  & 323.1 & 0.75  & 0.63    \\
\multicolumn{1}{l|}{MaskBit~\cite{maskbit}}                                                       & 305M                                           & LFQ                                          & 2D                                           & 54M                                            & 256                         & \multicolumn{1}{c|}{1.61}                         & -     & -     & -     & \multicolumn{1}{c|}{-}                            & 1.52  & 328.6 & -     & -       \\
\multicolumn{1}{l|}{MAR-H~\cite{MAR}}                                                         & 943M                                           & KL                                           & 2D                                           & 66M                                            & 256                         & \multicolumn{1}{c|}{1.22}                         & 2.35  & 227.8 & 0.79  & \multicolumn{1}{c|}{0.62}                         & 1.55  & 303.7 & 0.81  & 0.62    \\
\multicolumn{1}{l|}{\textit{l}-DeTok~\cite{DeTok}}                                                       & 479M                                           & KL                                           & 2D                                           & 172M                                           & 256                         & \multicolumn{1}{c|}{0.62}                         & 1.86  & 238.6 & 0.82  & \multicolumn{1}{c|}{0.61}                         & 1.35  & 304.1 & 0.81  & 0.62    \\
\multicolumn{1}{l|}{TiTok-S-128~\cite{TiTok}}                                                   & 287M                                           & VQ                                           & 1D                                           & 72M                                            & 128                         & \multicolumn{1}{c|}{1.61}                         & -     & -     & -     & \multicolumn{1}{c|}{-}                            & 1.97  & 281.8 & -     & -       \\
\multicolumn{1}{l|}{GigaTok~\cite{gigatok}}                                                       & 111M                                           & VQ                                           & 1D                                           & 622M                                           & 256                         & \multicolumn{1}{c|}{0.51}                         & -     & -     & -     & \multicolumn{1}{c|}{-}                            & 3.15  & 224.3 & 0.82  & 0.55    \\
\multicolumn{1}{l|}{ImageFolder~\cite{imagefolder}}                                                   & 362M                                           & MSRQ                                         & 1D                                           & 176M                                           & 286                         & \multicolumn{1}{c|}{0.80}                          & -     & -     & -     & \multicolumn{1}{c|}{-}                            & 2.60  & 295.0 & 0.75  & 0.63    \\ \midrule
\textit{Diffusion-based}                                                           &                                                &                                              &                                              &                                                &                             &                                                   &       &       &       &                                                   &       &       &       &         \\
\multicolumn{1}{l|}{LDM-4~\cite{SD-VAE}}                                                         & 400M                                           &                                              & 2D                                           &                                                &                             & \multicolumn{1}{c|}{}                             & 10.56 & 103.5 & 0.71  & \multicolumn{1}{c|}{0.62}                         & 3.60  & 247.7 & 0.87  & 0.48    \\
\multicolumn{1}{l|}{U-ViT-H/2~\cite{U-ViT}}                                                     & 501M                                           &                                              & 2D                                           &                                                &                             & \multicolumn{1}{c|}{}                             & -     & -     & -     & \multicolumn{1}{c|}{-}                            & 2.29  & 263.9 & 0.82  & 0.57    \\
\multicolumn{1}{l|}{MDTv2-XL/2~\cite{MDTv2}}                                                    & 676M                                           & \multirow{-3}{*}{\text{KL}}                        & 2D                                           & \multirow{-3}{*}{55M}                          & \multirow{-3}{*}{4096}      & \multicolumn{1}{c|}{\multirow{-3}{*}{0.27}}       & 5.06  & 155.6 & 0.72  & \multicolumn{1}{c|}{0.66}                         & 1.58  & 314.7 & 0.79  & 0.65    \\
\multicolumn{1}{l|}{DiT-XL/2~\cite{DiT}}                                                      & 675M                                           &                                              & 2D                                           &                                                &                             & \multicolumn{1}{c|}{}                             & 9.62  & 121.5 & 0.67  & \multicolumn{1}{c|}{0.67}                         & 2.27  & 278.2 & 0.79  & 0.65    \\
\multicolumn{1}{l|}{SiT-XL/2~\cite{SiT}}                                                      &                                                &                                              & 2D                                           &                                                &                             & \multicolumn{1}{c|}{}                             & 8.30  & 131.7 & 0.68  & \multicolumn{1}{c|}{0.67}                         & 2.06  & 270.3 & 0.83  & 0.53    \\
\multicolumn{1}{l|}{+REPA~\cite{repa}}                                                         & \multirow{-2}{*}{675M}                         & \multirow{-3}{*}{\text{KL}}                        & 2D                                           & \multirow{-3}{*}{84M}                          & \multirow{-3}{*}{1024}      & \multicolumn{1}{c|}{\multirow{-3}{*}{0.62}}       & 5.90  & 157.8 & 0.70  & \multicolumn{1}{c|}{0.69}                         & 1.42  & 305.7 & 0.82  & 0.59    \\
\multicolumn{1}{l|}{LightningDiT~\cite{va-vae}}                                                  & 675M                                           & KL                                           & 2D                                           & 70M                                            & 256                         & \multicolumn{1}{c|}{0.28}                         & 2.17  & 205.6 & -     & \multicolumn{1}{c|}{-}                            & 1.35  & 295.3 & -     & -       \\
\multicolumn{1}{l|}{TexTok-256~\cite{TexTok}}                                                    & 675M                                           & KL                                           & 1D                                           & 176M                                           & 256                         & \multicolumn{1}{c|}{0.73}                         & -     & -     & -     & \multicolumn{1}{c|}{-}                            & 1.46  & 303.1 & 0.79  & 0.64    \\
\multicolumn{1}{l|}{MAETok~\cite{MAETok}}                                                        & 675M                                           & AE                                           & 1D                                           & 176M                                           & 128                         & \multicolumn{1}{c|}{0.48}                         & 2.31  & 216.5 & 0.78  & \multicolumn{1}{c|}{0.62}                         & 1.67  & 311.2 & 0.81  & 0.63    \\
\multicolumn{1}{l|}{SoftVQ-VAE~\cite{softvq-vae}}                                                    & 675M                                           & SoftVQ                                       & 1D                                           & 176M                                           & 64                          & \multicolumn{1}{c|}{0.88}                         & 5.98  & 138.0 & 0.74  & \multicolumn{1}{c|}{0.64}                         & 1.78  & 279.0 & 0.80  & 0.63    \\ \midrule
\textit{Ours}                                                                      &                                                &                                              &                                              &                                                &                             &                                                   &       &       &       &                                                   &       &       &       &         \\
\rowcolor[HTML]{EFEFEF} 
\multicolumn{1}{l|}{\cellcolor[HTML]{EFEFEF}}                                      & \cellcolor[HTML]{EFEFEF}                       & \cellcolor[HTML]{EFEFEF}                     & \cellcolor[HTML]{EFEFEF}                     & \cellcolor[HTML]{EFEFEF}                       & 64                          & \multicolumn{1}{c|}{\cellcolor[HTML]{EFEFEF}0.75} & 4.15  & 167.8 & 0.75  & \multicolumn{1}{c|}{\cellcolor[HTML]{EFEFEF}0.65} & 1.68  & 307.3 & 0.77  & 0.66    \\
\rowcolor[HTML]{EFEFEF}  	 
\multicolumn{1}{l|}{\multirow{-2}{*}{\cellcolor[HTML]{EFEFEF}MacTok+LightningDiT}} & \multirow{-2}{*}{\cellcolor[HTML]{EFEFEF}675M} & \cellcolor[HTML]{EFEFEF}                     & \cellcolor[HTML]{EFEFEF}                     & \cellcolor[HTML]{EFEFEF}                       & 128                         & \multicolumn{1}{c|}{\cellcolor[HTML]{EFEFEF}0.43} & 3.12  & 186.2 & 0.75  & \multicolumn{1}{c|}{\cellcolor[HTML]{EFEFEF}0.66} & 1.50  & 299.8 & 0.78  & 0.67    \\
\rowcolor[HTML]{EFEFEF} 
\multicolumn{1}{l|}{\cellcolor[HTML]{EFEFEF}}                                      & \cellcolor[HTML]{EFEFEF}                       & \cellcolor[HTML]{EFEFEF}                     & \cellcolor[HTML]{EFEFEF}                     & \cellcolor[HTML]{EFEFEF}                       & 64                          & \multicolumn{1}{c|}{\cellcolor[HTML]{EFEFEF}0.75} & 3.77  & 181.6 & 0.77  & \multicolumn{1}{c|}{\cellcolor[HTML]{EFEFEF}0.63} & 1.58  & 310.4 & 0.78  & 0.66    \\
\rowcolor[HTML]{EFEFEF} 
\multicolumn{1}{l|}{\multirow{-2}{*}{\cellcolor[HTML]{EFEFEF}MacTok+SiT-XL}}       & \multirow{-2}{*}{\cellcolor[HTML]{EFEFEF}675M} & \multirow{-4}{*}{\cellcolor[HTML]{EFEFEF}KL} & \multirow{-4}{*}{\cellcolor[HTML]{EFEFEF}1D} & \multirow{-4}{*}{\cellcolor[HTML]{EFEFEF}176M} & 128                         & \multicolumn{1}{c|}{\cellcolor[HTML]{EFEFEF}0.43} & 2.82  & 189.2 & 0.77  & \multicolumn{1}{c|}{\cellcolor[HTML]{EFEFEF}0.64} & 1.44  & 302.5 & 0.79  & 0.66    \\ \bottomrule
\end{tabular}
}
\end{table*}
\begin{table*}[t]
\caption{
\textbf{System-level comparison} on ImageNet 512$\times$512 conditional generation.
We report both Precision and Recall under classifier-free guidance (CFG) and non-CFG settings.
}
\label{tab:appendix_512}
\centering
\resizebox{1\linewidth}{!}
{
\begin{tabular}{@{}lcccccccccccccc@{}}
\toprule
\multicolumn{1}{l|}{}                                                        &                                                &                                              &                                              &                                &                             & \multicolumn{1}{c|}{}                             & \multicolumn{4}{c|}{w/o CFG}                                              & \multicolumn{4}{c}{w/ CFG}                       \\
\multicolumn{1}{l|}{\multirow{-2}{*}{Method}}                                & \multirow{-2}{*}{\# Params(G)}                 & \multirow{-2}{*}{Tok. Model}                 & \multirow{-2}{*}{Token Type}                  & \multirow{-2}{*}{\# Params(T)} & \multirow{-2}{*}{\#Tokens↓} & \multicolumn{1}{c|}{\multirow{-2}{*}{Tok. rFID↓}} & gFID↓ & IS↑   & Prec↑ & \multicolumn{1}{c|}{Recall↑}                      & gFID↓         & IS↑            & Prec↑ & Recall↑ \\ \midrule
\textit{GAN}                                                                 &                                                &                                              &                                              &                                &                             &                                                   &       &       &       &                                                   &               &                &       &         \\
\multicolumn{1}{l|}{BigGAN~\cite{maskgit}}                                                  & -                                              & -                                            & -                                            & -                              & -                           & \multicolumn{1}{c|}{-}                            & -     & -     & -     & \multicolumn{1}{c|}{-}                            & 8.43          & 177.9          & -     & -       \\
\multicolumn{1}{l|}{StyleGAN-XL~\cite{stylegan}}                                             & 168M                                           & -                                            & -                                            & -                              & -                           & \multicolumn{1}{c|}{-}                            & -     & -     & -     & \multicolumn{1}{c|}{-}                            & 2.41          & 267.7          & -     & -       \\ \midrule
\textit{Auto-regressive}                                                     &                                                &                                              &                                              &                                &                             &                                                   &       &       &       &                                                   &               &                &       &         \\
\multicolumn{1}{l|}{MaskGIT~\cite{maskgit}}                                                 & 227M                                           & VQ                                           & 2D                                           & 66M                            & 1024                        & \multicolumn{1}{c|}{1.97}                         & 7.32  & 156.0   & -     & \multicolumn{1}{c|}{-}                            & -             & -              & -     & -       \\
\multicolumn{1}{l|}{MAGVIT-v2~\cite{MAGVIT-v2}}                                               & 307M                                           & LFQ                                          & 2D                                           & 116M                           & 1024                        & \multicolumn{1}{c|}{-}                            & -     & -     & -     & \multicolumn{1}{c|}{-}                            & 1.91          & 324.3          & -     & -       \\
\multicolumn{1}{l|}{MAR-H~\cite{MAR}}                                                   & 943M                                           & KL                                           & 2D                                           & 66M                            & 1024                        & \multicolumn{1}{c|}{-}                            & 2.74  & 205.2 & 0.69  & \multicolumn{1}{c|}{0.59}                         & 1.73          & 279.9          & 0.77  & 0.61    \\
\multicolumn{1}{l|}{TiTok-B-128~\cite{TiTok}}                                              & 177M                                           & VQ                                           & 1D                                           & 202M                           & 128                         & \multicolumn{1}{c|}{1.52}                         & -     & -     & -     & \multicolumn{1}{c|}{-}                            & 2.13          & 261.2          & -     & -       \\
\multicolumn{1}{l|}{TiTok-L-64~\cite{TiTok}}                                              & 177M                                           & VQ                                           & 1D                                           & 644M                           & 64                         & \multicolumn{1}{c|}{1.77}                         & -     & -     & -     & \multicolumn{1}{c|}{-}                            & 2.74          & 221.1          & -     & -       \\ \midrule
\textit{Diffusion-based}                                                     & \textit{}                                      &                                              &                                              &                                &                             &                                                   &       &       &       &                                                   &               &                &       &         \\
\multicolumn{1}{l|}{ADM~\cite{ADM}}                                                     & -                                              & -                                            & -                                            & -                              & -                           & \multicolumn{1}{c|}{-}                            & 23.24 & 58.1 & -     & \multicolumn{1}{c|}{-}                            & 3.85          & 221.7          & 0.84  & 0.53    \\
\multicolumn{1}{l|}{U-ViT-H/4~\cite{U-ViT}}                                               & 501M                                           &                                              & 2D                                           &                                &                             & \multicolumn{1}{c|}{}                             & -     & -     & -     & \multicolumn{1}{c|}{-}                            & 4.05          & 263.8          & 0.84  & 0.48    \\
\multicolumn{1}{l|}{DiT-XL/2~\cite{DiT}}                                                & 675M                                           &                                              & 2D                                           &                                &                             & \multicolumn{1}{c|}{}                             & 9.62  & 121.5 & -     & \multicolumn{1}{c|}{-}                            & 3.04          & 240.8          & 0.84  & 0.54    \\
\multicolumn{1}{l|}{SiT-XL/2~\cite{SiT}}                                                & 675M                                           &                                              & 2D                                           &                                &                             & \multicolumn{1}{c|}{}                             & -     & -     & -     & \multicolumn{1}{c|}{-}                            & 2.62          & 252.2          & 0.84  & 0.57    \\
\multicolumn{1}{l|}{DiT-XL~\cite{DiT}}                                                  & 675M                                           &                                              & 2D                                           &                                &                             & \multicolumn{1}{c|}{}                             & 9.56  & -     & -     & \multicolumn{1}{c|}{-}                            & 2.84          & -              & -     & -       \\
\multicolumn{1}{l|}{UViT-H~\cite{U-ViT}}                                                  & 501M                                           & \multirow{-5}{*}{\text{KL}}                        & 2D                                           & \multirow{-5}{*}{84M}          & \multirow{-5}{*}{4096}      & \multicolumn{1}{c|}{\multirow{-5}{*}{0.62}}       & 9.83  & -     & -     & \multicolumn{1}{c|}{-}                            & 2.53          & -              & -     & -       \\
\multicolumn{1}{l|}{UViT-H}                                                  & 501M                                           &                                              & 2D                                           &                                &                             & \multicolumn{1}{c|}{}                             & 12.26 & -     & -     & \multicolumn{1}{c|}{-}                            & 2.66          & -              & -     & -       \\
\multicolumn{1}{l|}{UViT-2B~\cite{U-ViT}}                                                 & 2B                                             & \multirow{-2}{*}{\text{AE}}                        & 2D                                           & \multirow{-2}{*}{323M}         & \multirow{-2}{*}{256}       & \multicolumn{1}{c|}{\multirow{-2}{*}{0.22}}       & 6.50   & -     & -     & \multicolumn{1}{c|}{-}                            & 2.25          & -              & -     & -       \\
\multicolumn{1}{l|}{TexTok-128~\cite{TexTok}}                                              & 675M                                           & KL                                           & 1D                                           & 176M                           & 128                         & \multicolumn{1}{c|}{0.97}                         & -     & -     & -     & \multicolumn{1}{c|}{-}                            & 1.80          & 305.4          & 0.81  & 0.63    \\
\multicolumn{1}{l|}{MAETok~\cite{MAETok}}                                                  & 675M                                           & AE                                           & 1D                                           & 176M                           & 128                         & \multicolumn{1}{c|}{0.62}                         & 2.79  & 204.3 & 0.81  & \multicolumn{1}{c|}{0.62}                         & 1.69          & 304.2          & 0.82  & 0.62    \\
\multicolumn{1}{l|}{SoftVQ-VAE~\cite{softvq-vae}}                                              & 675M                                           & SoftVQ                                       & 1D                                           & 391M                           & 64                          & \multicolumn{1}{c|}{0.71}                         & 7.96  & 133.9 & 0.73  & \multicolumn{1}{c|}{0.63}                         & 2.21          & 290.5          & 0.85  & 0.59    \\ \midrule
\textit{Ours}                                                                &                                                &                                              &                                              &                                &                             &                                                   &       &       &       &                                                   &               &                &       &         \\
\rowcolor[HTML]{EFEFEF} 
\multicolumn{1}{l|}{\cellcolor[HTML]{EFEFEF}}                                & \cellcolor[HTML]{EFEFEF}                       & \cellcolor[HTML]{EFEFEF}                     & \cellcolor[HTML]{EFEFEF}                     & 391M                           & 64                          & \multicolumn{1}{c|}{\cellcolor[HTML]{EFEFEF}0.89} & 4.63  & 163.7 & 0.80   & \multicolumn{1}{c|}{\cellcolor[HTML]{EFEFEF}0.61} & \textbf{1.52} & 306.0          & 0.80  & 0.63    \\
\rowcolor[HTML]{EFEFEF} 
\multicolumn{1}{l|}{\multirow{-2}{*}{\cellcolor[HTML]{EFEFEF}MacTok+SiT-XL}} & \multirow{-2}{*}{\cellcolor[HTML]{EFEFEF}675M} & \multirow{-2}{*}{\cellcolor[HTML]{EFEFEF}KL} & \multirow{-2}{*}{\cellcolor[HTML]{EFEFEF}1D} & 176M                           & 128                         & \multicolumn{1}{c|}{\cellcolor[HTML]{EFEFEF}0.79} & 5.12  & 156.3 & 0.79  & \multicolumn{1}{c|}{\cellcolor[HTML]{EFEFEF}0.61} & \textbf{1.52} & \textbf{316.0} & 0.80  & 0.63    \\ \bottomrule
\end{tabular}
}
\end{table*}
\begin{table*}[t]
\caption{\centering Generation performance over training of SiT-XL trained on MacTok with 64 and 128 tokens.
}
\label{tab:appendix_along_training}
\centering
\resizebox{0.7\linewidth}{!}
{
\begin{tabular}{@{}c|c|cccc|cccc@{}}
\toprule
\multirow{2}{*}{Method}                                                      & \multirow{2}{*}{Training Iter.} & \multicolumn{4}{c|}{w/o CFG}  & \multicolumn{4}{c}{w/ CFG}    \\
                                                                             &                                 & FID  & IS    & Prec. & Recall & FID  & IS    & Prec. & Recall \\ \midrule
\multirow{5}{*}{\begin{tabular}[c]{@{}c@{}}MacTok-64\end{tabular}}  & 400K                            & 7.60 & 121.4 & 0.72  & 0.63   & 2.15 & 268.2 & 0.77  & 0.63   \\
                                                                             & 1M                              & 5.34 & 147.7 & 0.74  & 0.63   & 1.73 & 290.4 & 0.77  & 0.65   \\
                                                                             & 2M                              & 4.58 & 159.9 & 0.75  & 0.63   & 1.60 & 303.0 & 0.78  & 0.65   \\
                                                                             & 3M                              & 3.98 & 174.7 & 0.76  & 0.63   & 1.60 & 308.2 & 0.78  & 0.66   \\
                                                                             & 4M                              & 3.77 & 181.6 & 0.77  & 0.63   & 1.58 & 310.4 & 0.78  & 0.66   \\ \midrule
\multirow{5}{*}{\begin{tabular}[c]{@{}c@{}}MacTok-128\end{tabular}} & 400K                            & 6.45 & 127.2 & 0.73  & 0.63   & 1.97 & 253.2 & 0.77  & 0.64   \\
                                                                             & 1M                              & 4.31 & 153.6 & 0.75  & 0.64   & 1.60 & 271.7 & 0.77  & 0.65   \\
                                                                             & 2M                              & 3.69 & 168.5 & 0.75  & 0.65   & 1.48 & 287.0 & 0.78  & 0.66   \\
                                                                             & 3M                              & 3.28 & 176.2 & 0.76  & 0.65   & 1.45 & 293.1 & 0.78  & 0.66   \\
                                                                             & 4M                              & 2.82 & 189.2 & 0.77  & 0.64   & 1.44 & 302.5 & 0.79  & 0.66   \\ \bottomrule
\end{tabular}
}
\end{table*}

\begin{figure*}[tp]
    \centering
    \includegraphics[width=1\linewidth]{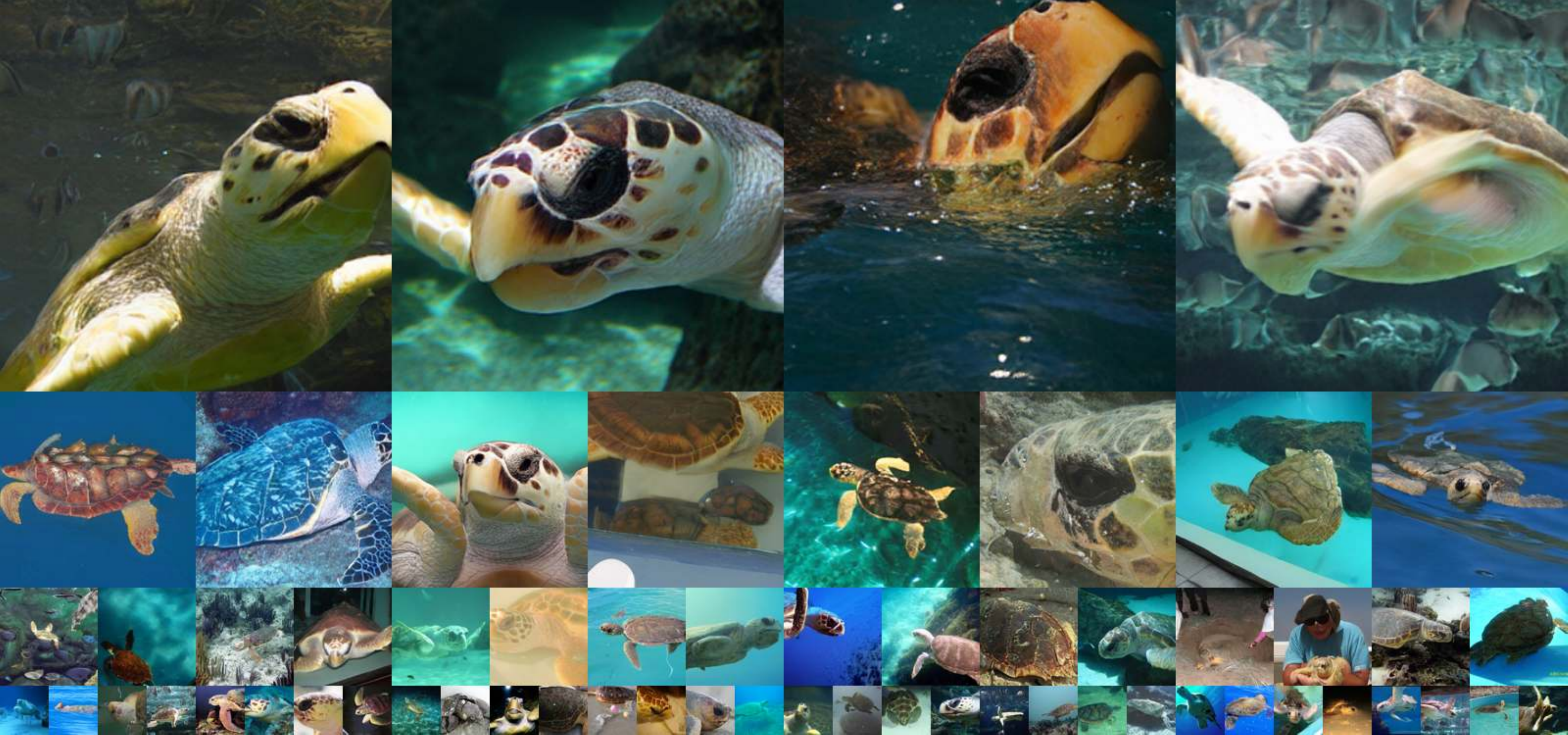}
    \caption{Uncurated 256$\times$256 generation results of SiT-XL with MacTok 128 tokens. We use CFG with 4.0. Class label =``loggerhead turtle'' (33).}
    \label{fig:033}
\end{figure*}

\begin{figure*}[tp]
    \centering
    \includegraphics[width=1\linewidth]{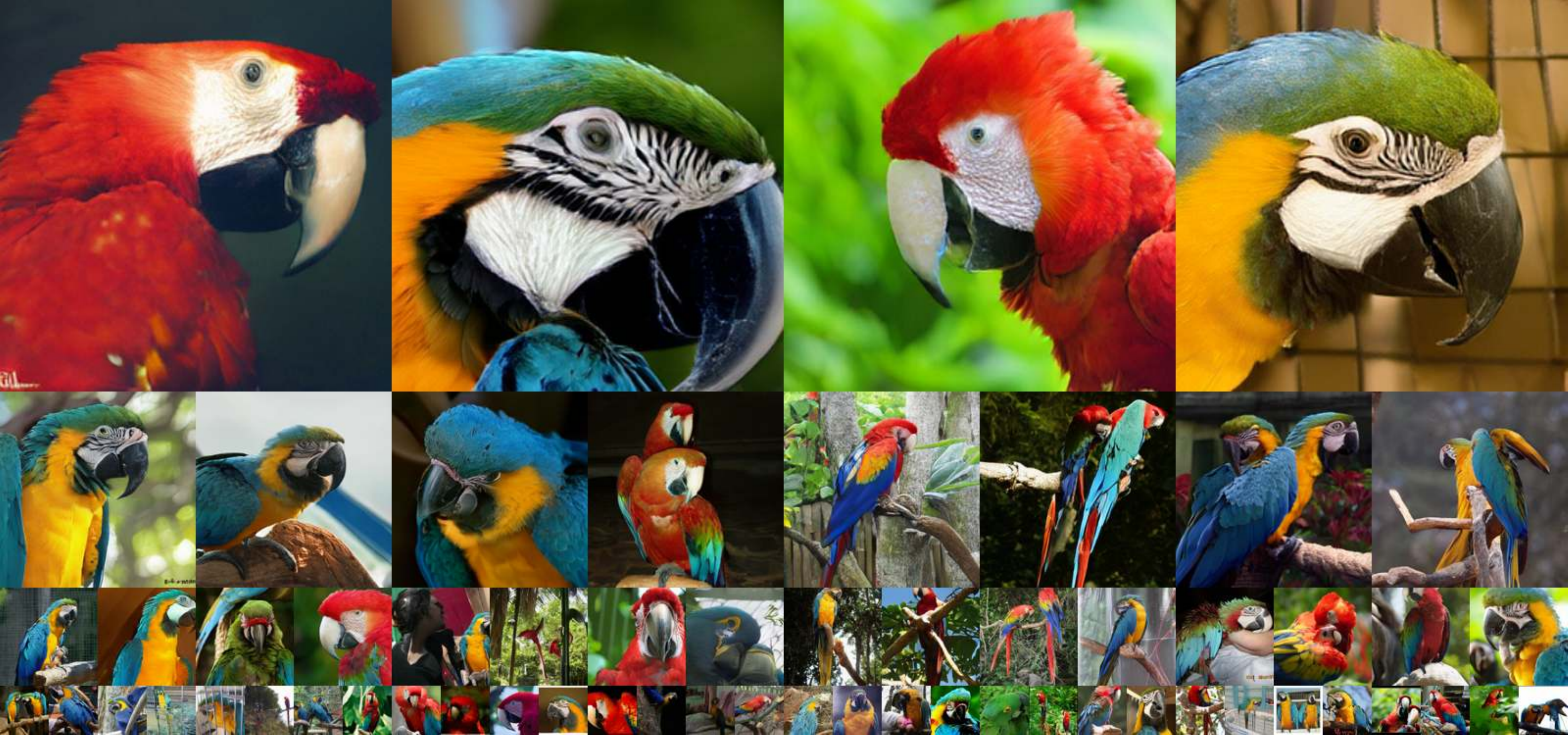}
    \caption{Uncurated 256$\times$256 generation results of SiT-XL with MacTok 128 tokens. We use CFG with 4.0. Class label =``macaw'' (88).}
    \label{fig:088}
\end{figure*}

\begin{figure*}[tp]
    \centering
    \includegraphics[width=1\linewidth]{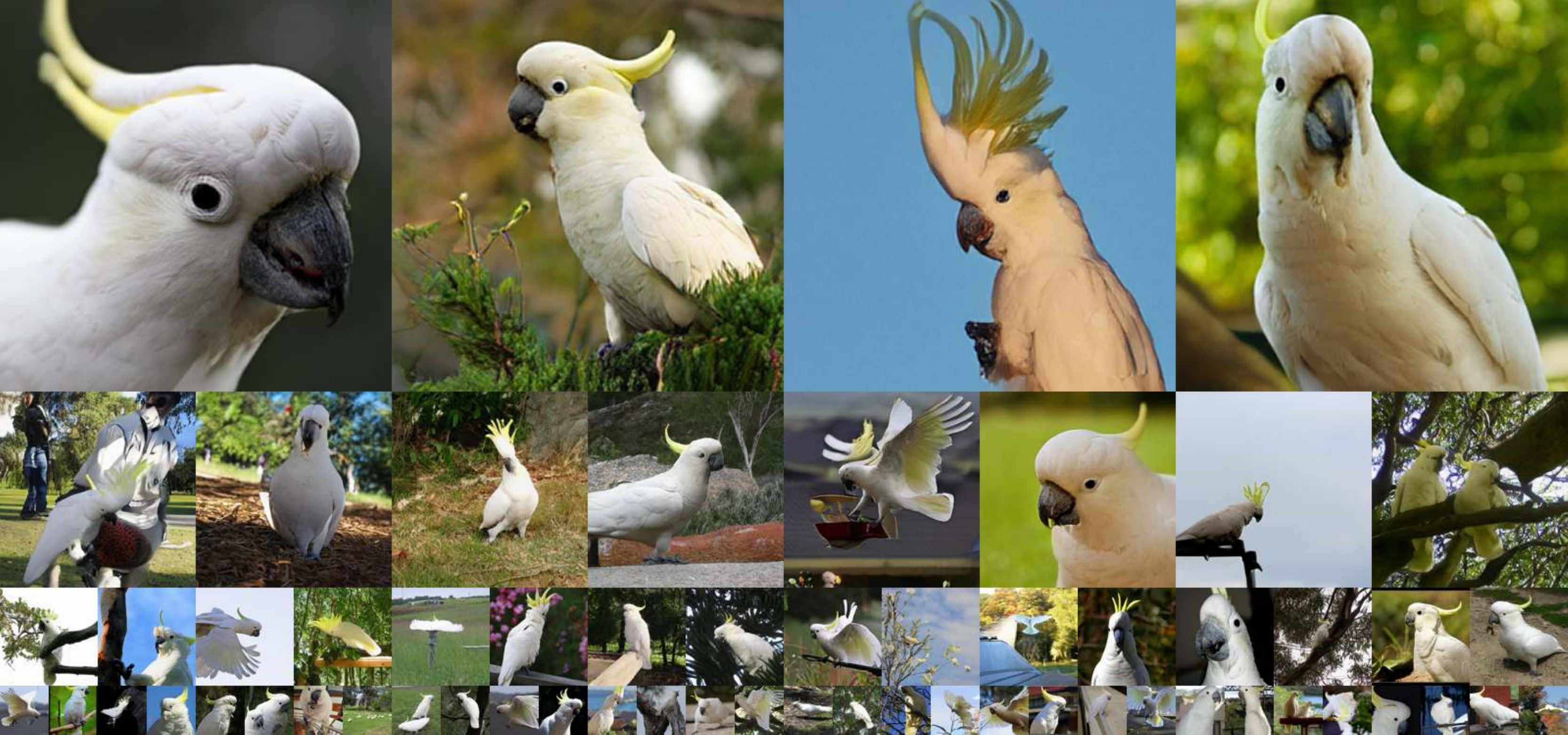}
    \caption{Uncurated 256$\times$256 generation results of SiT-XL with MacTok 128 tokens. We use CFG with 4.0. Class label =``Kakatoe galerita'' (89).}
    \label{fig:089}
\end{figure*}

\begin{figure*}[tp]
    \centering
    \includegraphics[width=1\linewidth]{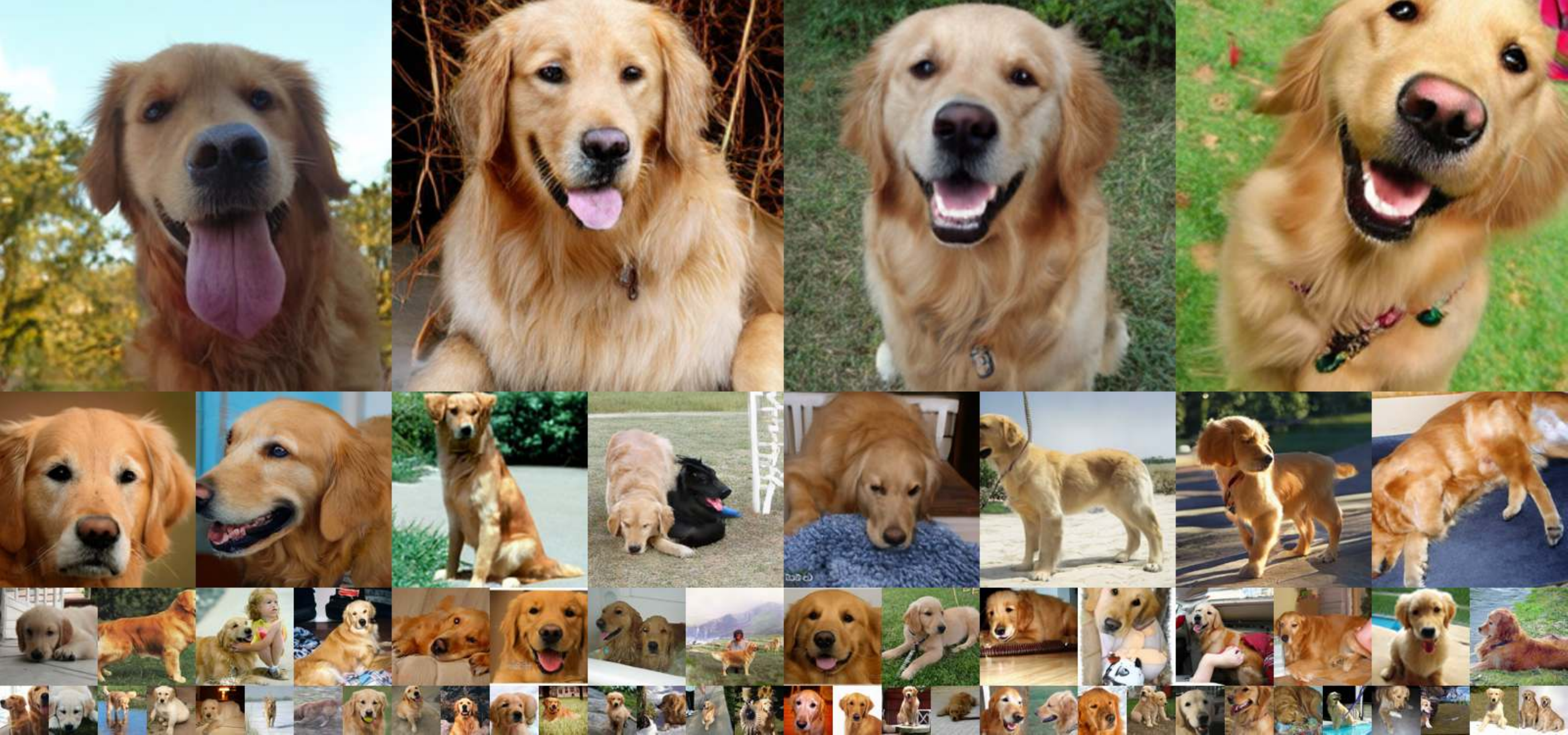}
    \caption{Uncurated 256$\times$256 generation results of SiT-XL with MacTok 128 tokens. We use CFG with 4.0. Class label =``golden retriever'' (207).}
    \label{fig:207}
\end{figure*}

\begin{figure*}[tp]
    \centering
    \includegraphics[width=1\linewidth]{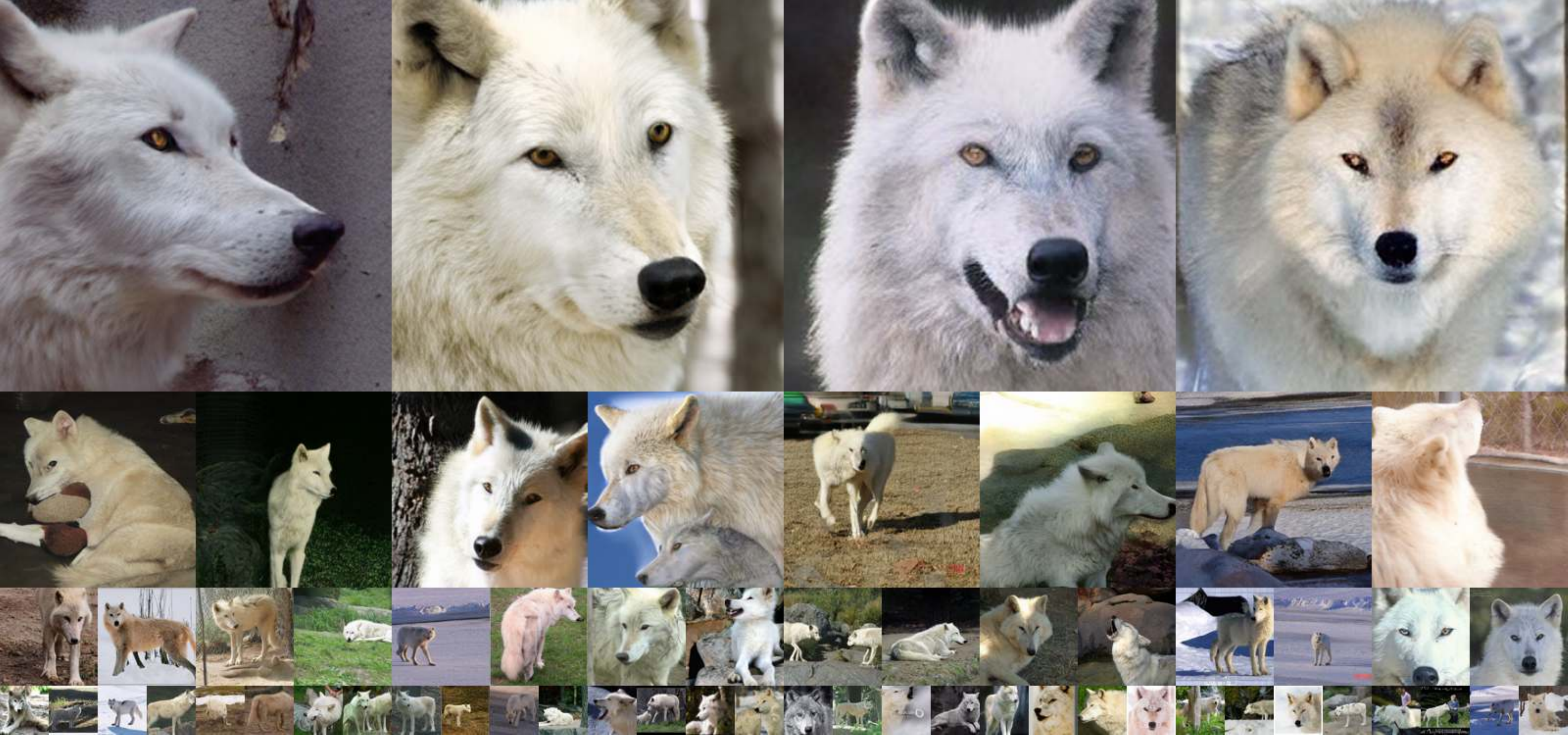}
    \caption{Uncurated 256$\times$256 generation results of SiT-XL with MacTok 128 tokens. We use CFG with 4.0. Class label =``Arctic wolf'' (270).}
    \label{fig:270}
\end{figure*}

\begin{figure*}[tp]
    \centering
    \includegraphics[width=1\linewidth]{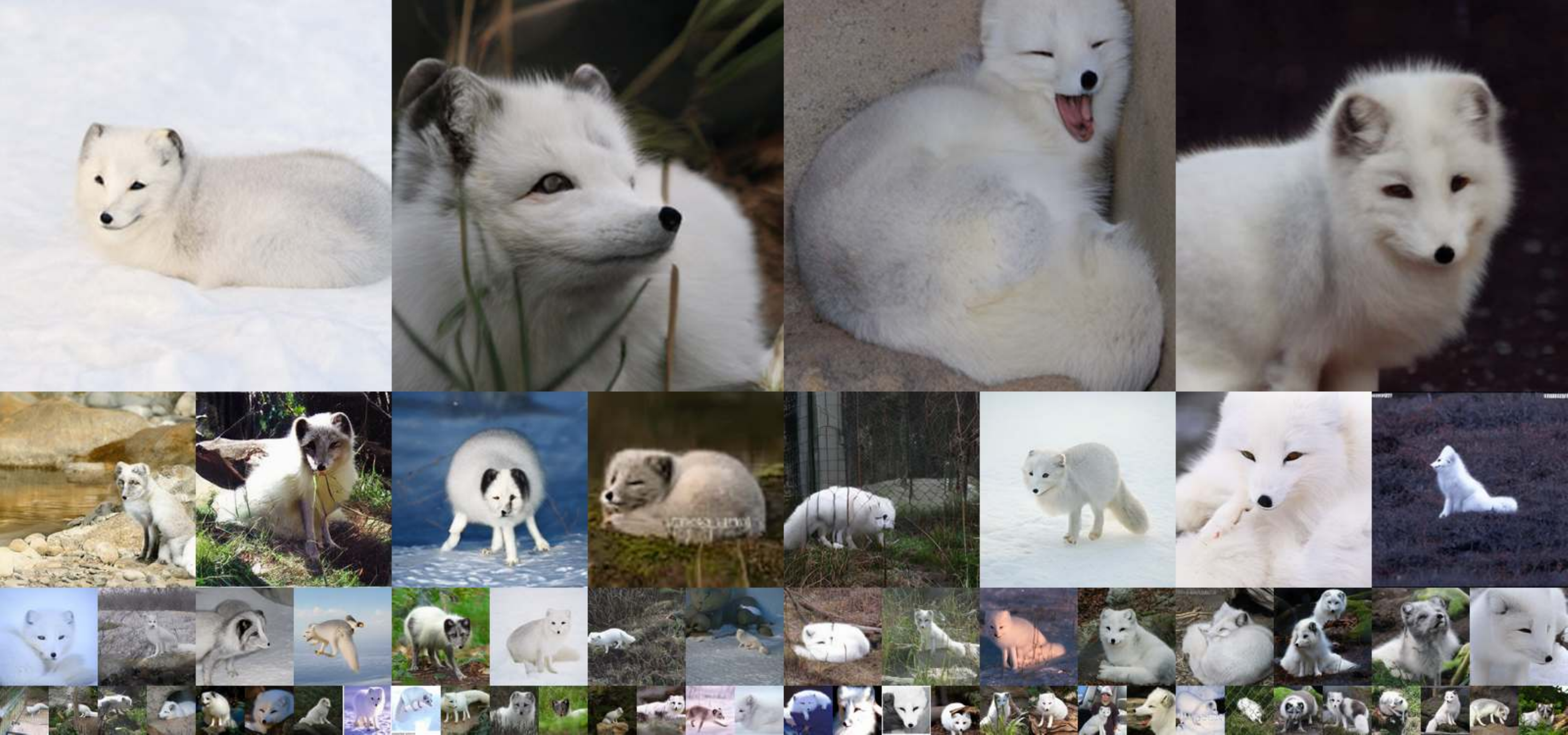}
    \caption{Uncurated 256$\times$256 generation results of SiT-XL with MacTok 128 tokens. We use CFG with 4.0. Class label =``Arctic fox'' (279).}
    \label{fig:279}
\end{figure*}

\begin{figure*}[tp]
    \centering
    \includegraphics[width=1\linewidth]{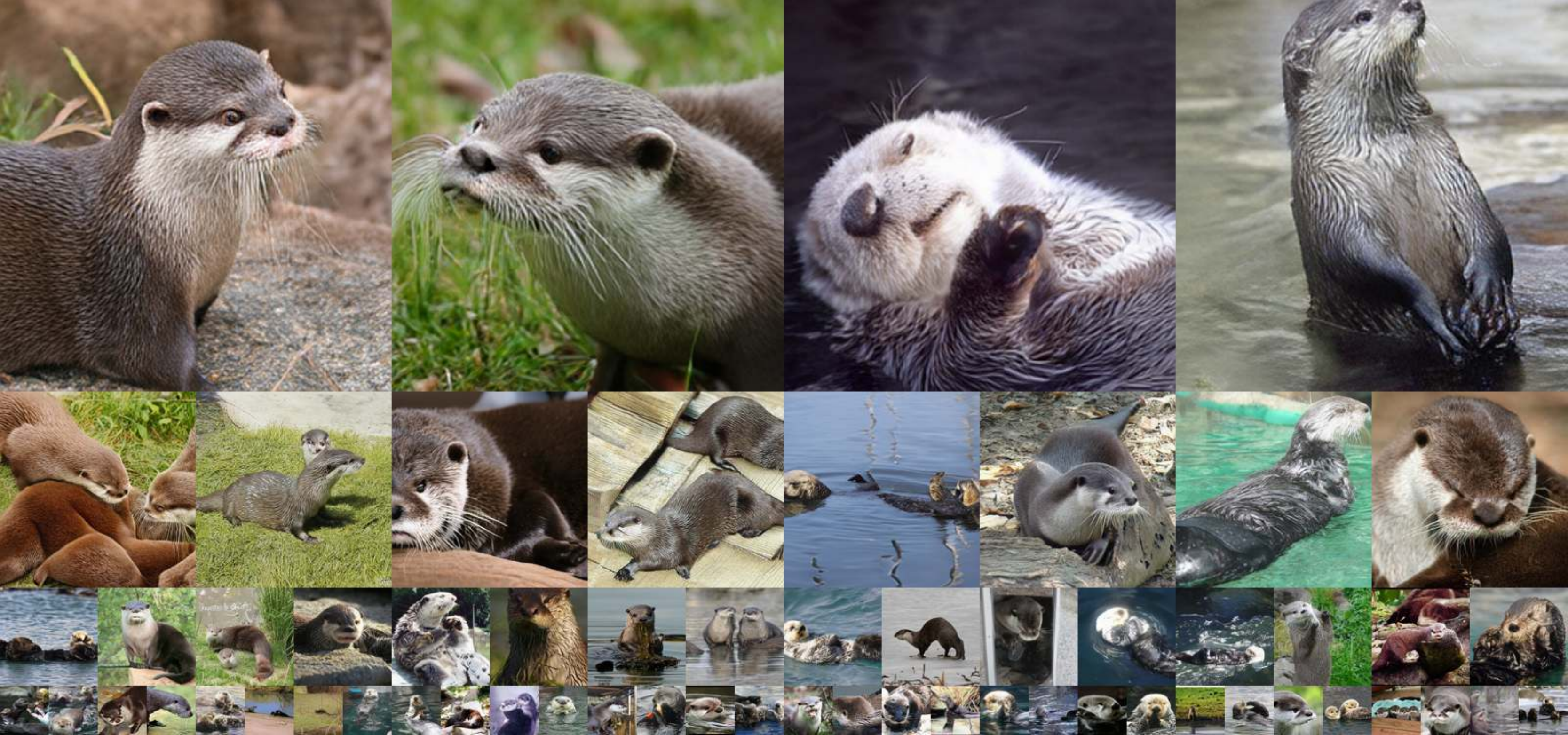}
    \caption{Uncurated 256$\times$256 generation results of SiT-XL with MacTok 128 tokens. We use CFG with 4.0. Class label =``otter'' (360).}
    \label{fig:360}
\end{figure*}

\begin{figure*}[tp]
    \centering
    \includegraphics[width=1\linewidth]{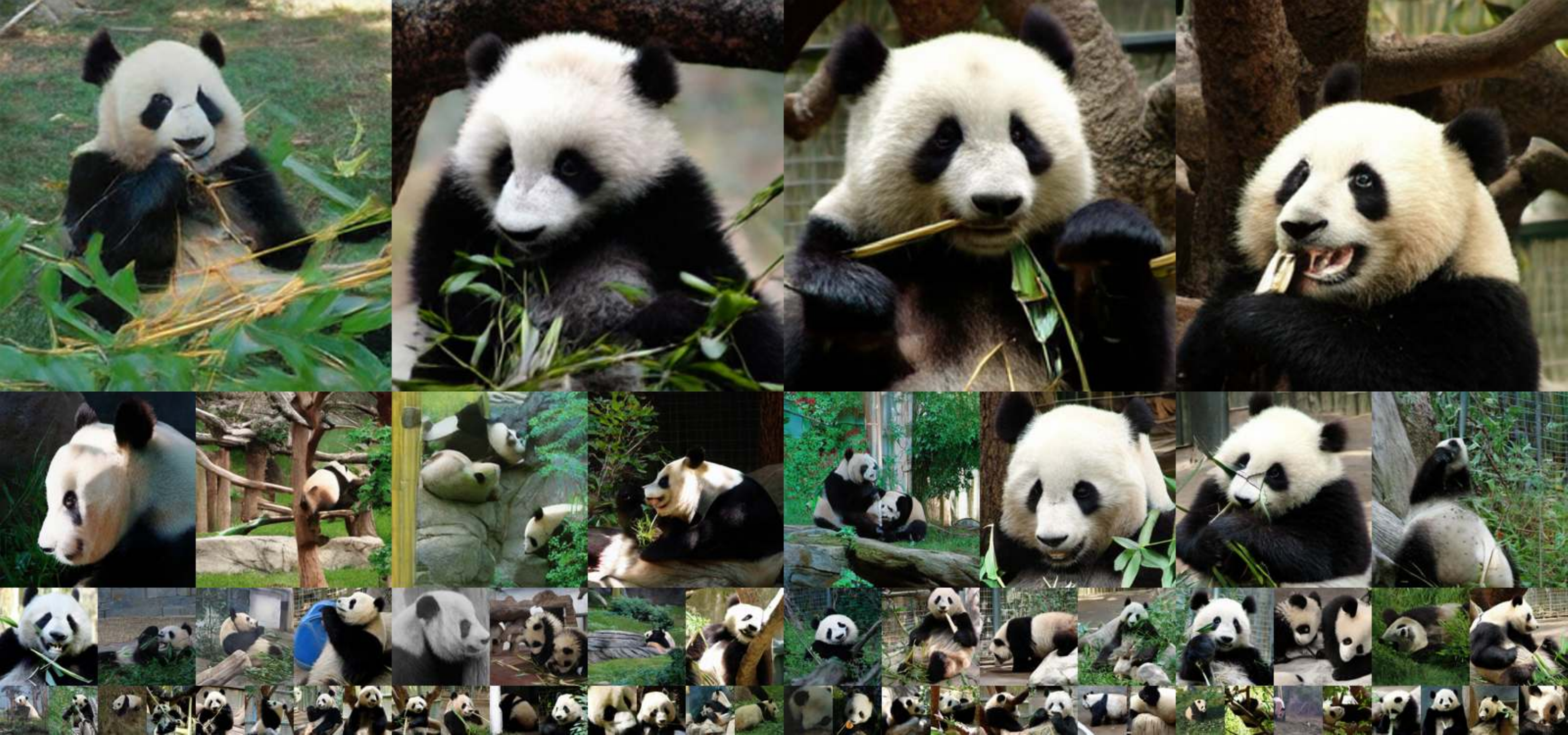}
    \caption{Uncurated 256$\times$256 generation results of SiT-XL with MacTok 128 tokens. We use CFG with 4.0. Class label =``panda'' (388).}
    \label{fig:388}
\end{figure*}

\begin{figure*}[tp]
    \centering
    \includegraphics[width=1\linewidth]{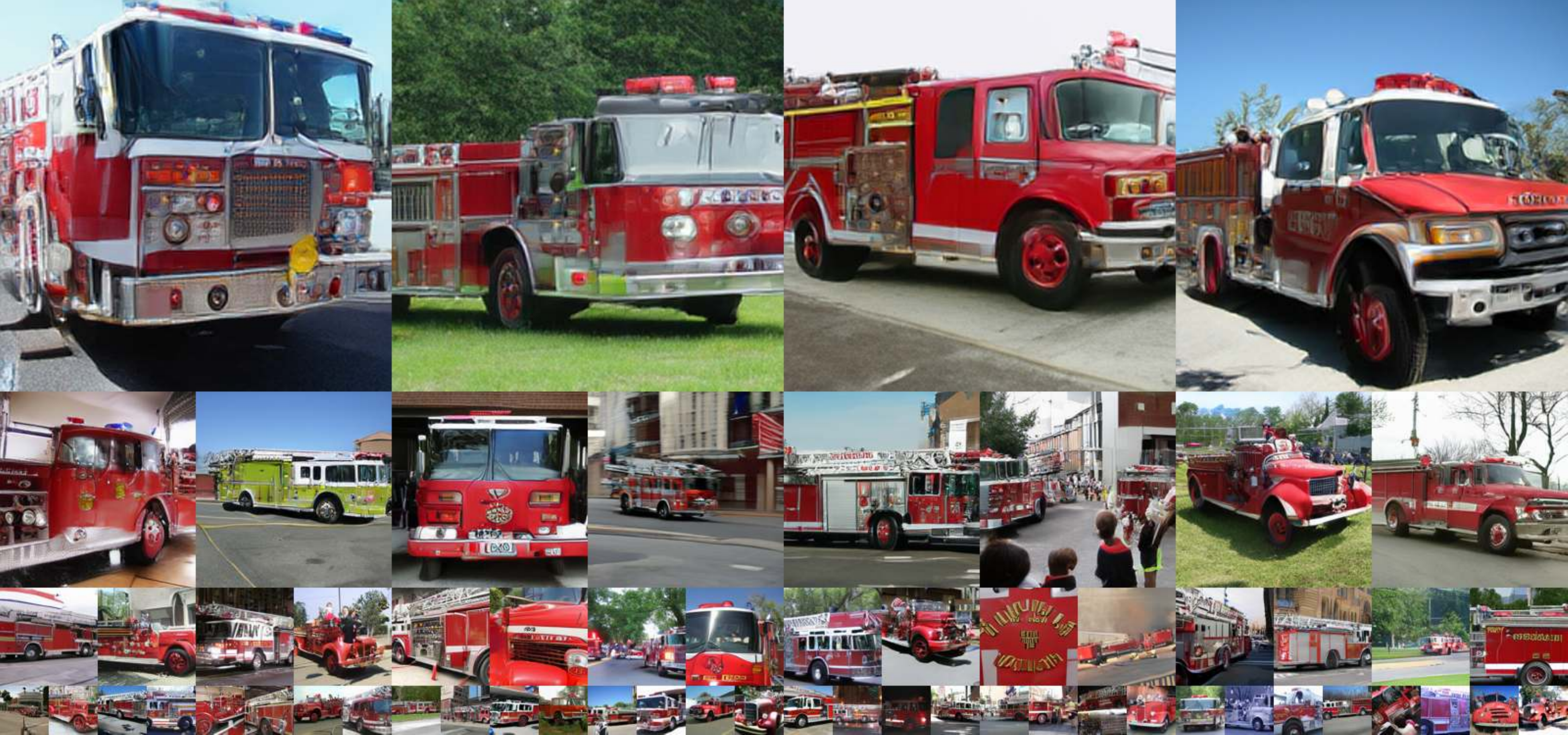}
    \caption{Uncurated 256$\times$256 generation results of SiT-XL with MacTok 64 tokens. We use CFG with 4.0. Class label =``fire engine'' (555).}
    \label{fig:555}
\end{figure*}

\begin{figure*}[tp]
    \centering
    \includegraphics[width=1\linewidth]{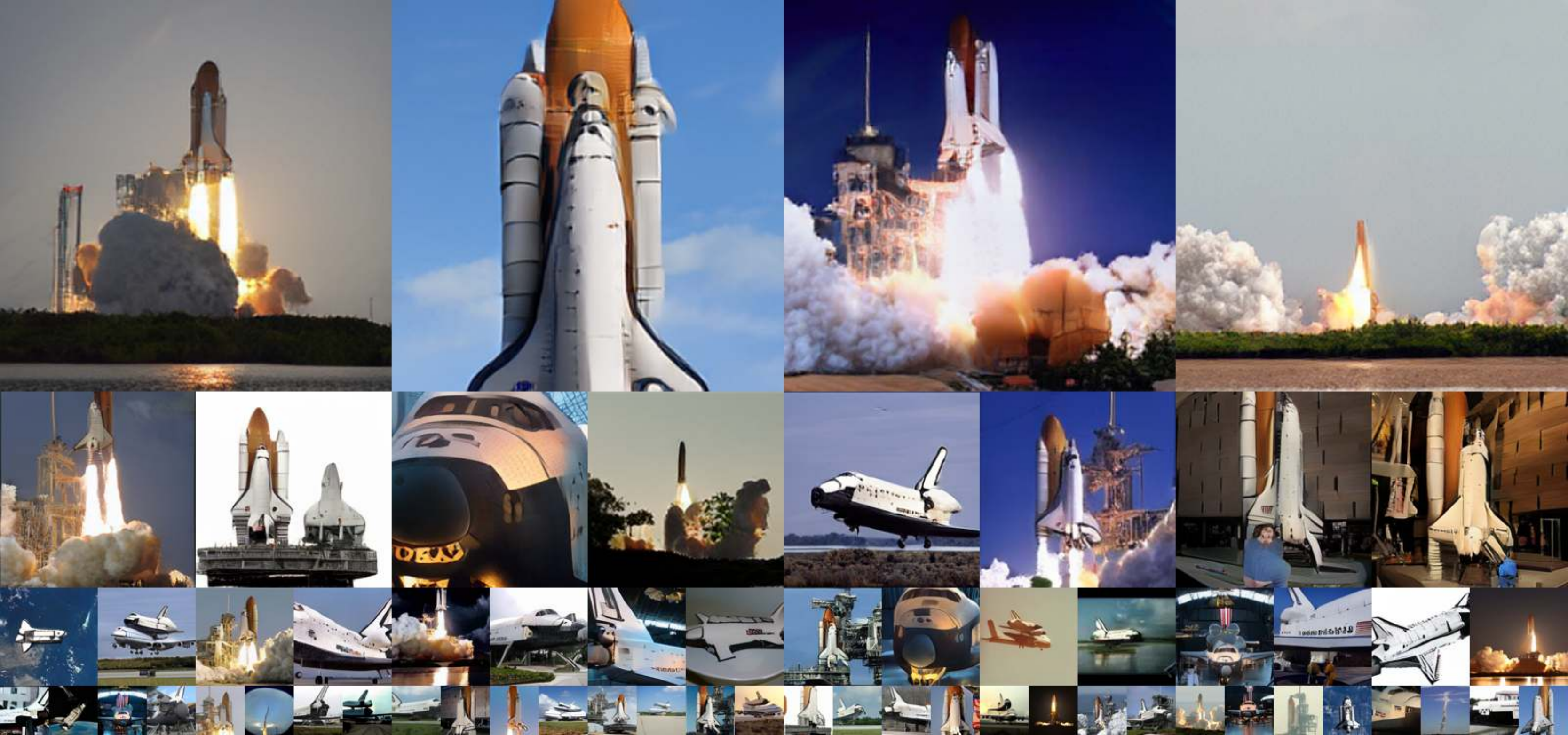}
    \caption{Uncurated 256$\times$256 generation results of SiT-XL with MacTok 64 tokens. We use CFG with 4.0. Class label =``space shuttle'' (812).}
    \label{fig:812}
\end{figure*}

\begin{figure*}[tp]
    \centering
    \includegraphics[width=1\linewidth]{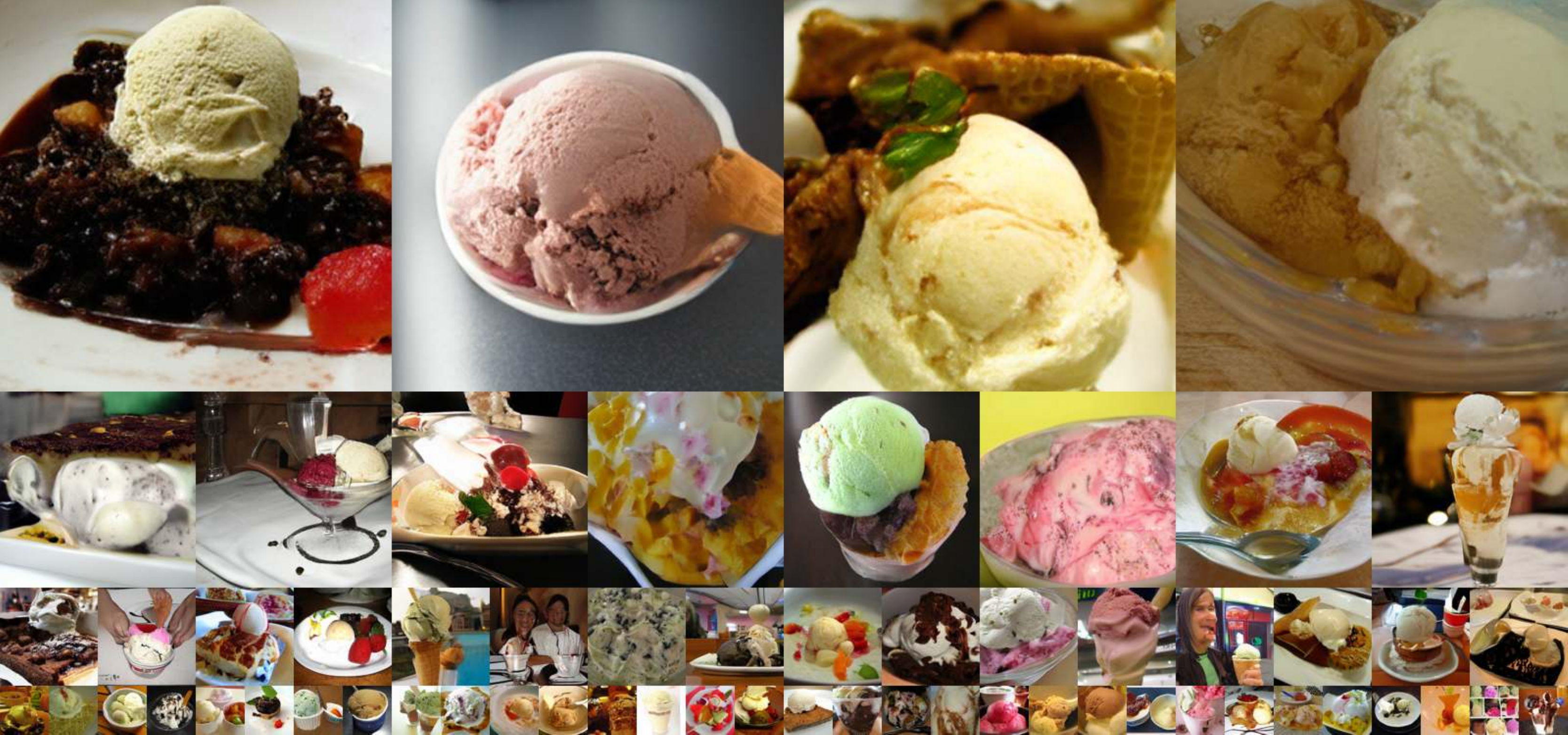}
    \caption{Uncurated 256$\times$256 generation results of SiT-XL with MacTok 64 tokens. We use CFG with 4.0. Class label =``ice cream'' (928).}
    \label{fig:928}
\end{figure*}

\begin{figure*}[tp]
    \centering
    \includegraphics[width=1\linewidth]{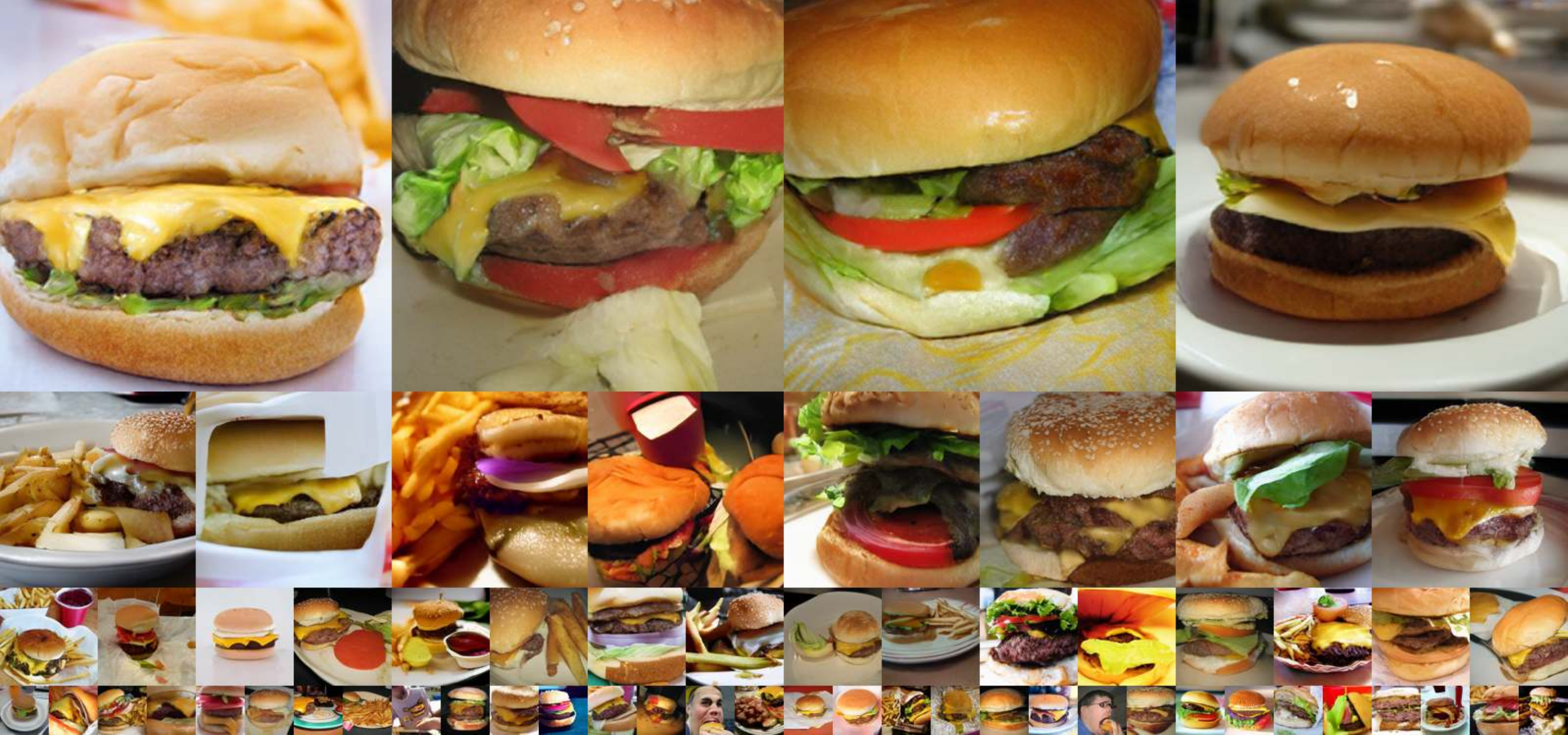}
    \caption{Uncurated 256$\times$256 generation results of SiT-XL with MacTok 64 tokens. We use CFG with 4.0. Class label =``cheeseburger'' (933).}
    \label{fig:933}
\end{figure*}

\begin{figure*}[tp]
    \centering
    \includegraphics[width=1\linewidth]{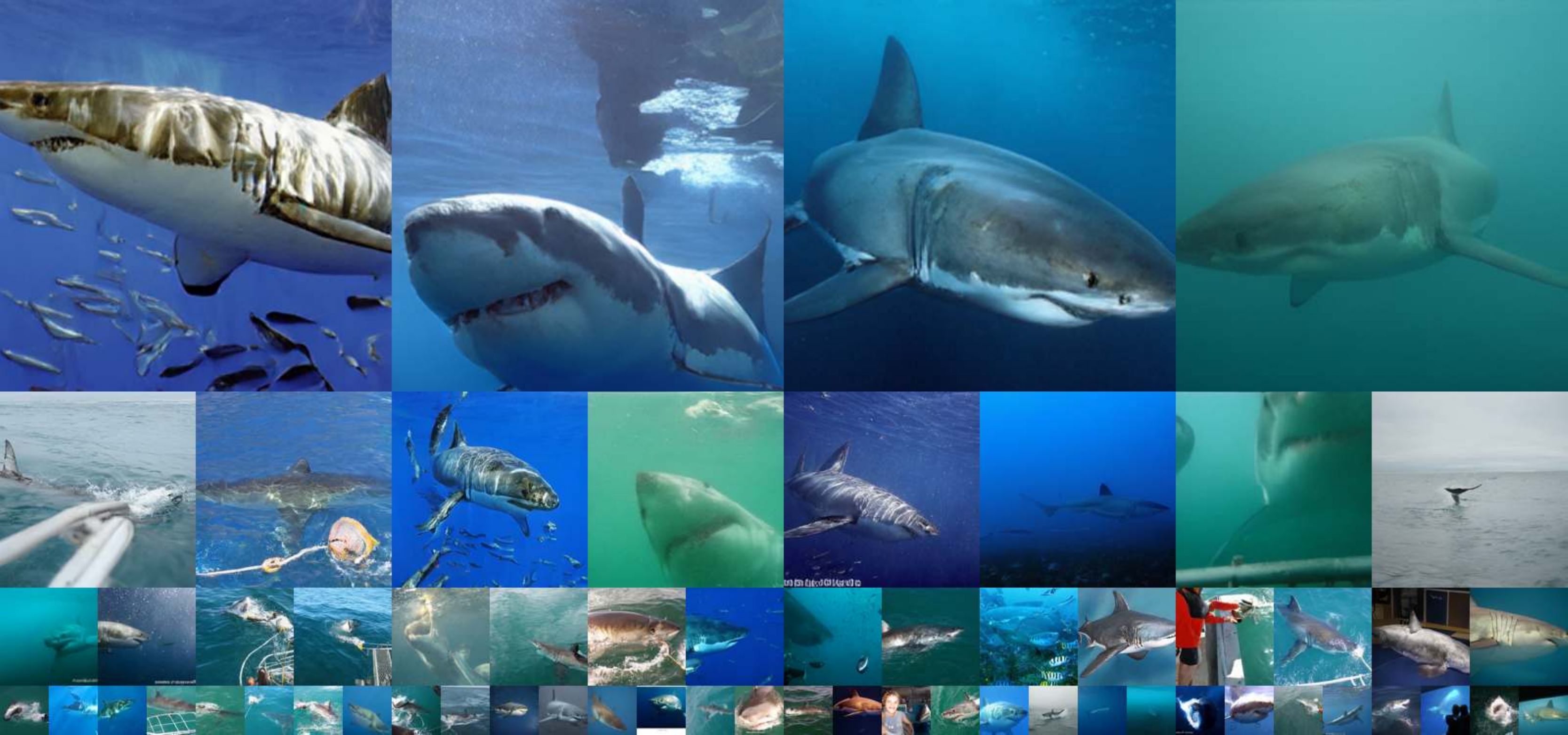}
    \caption{Uncurated 256$\times$256 generation results of LightningDiT-XL with MacTok 128 tokens. We use CFG with 3.0. Class label =``white shark'' (2).}
    \label{fig:002}
\end{figure*}

\begin{figure*}[tp]
    \centering
    \includegraphics[width=1\linewidth]{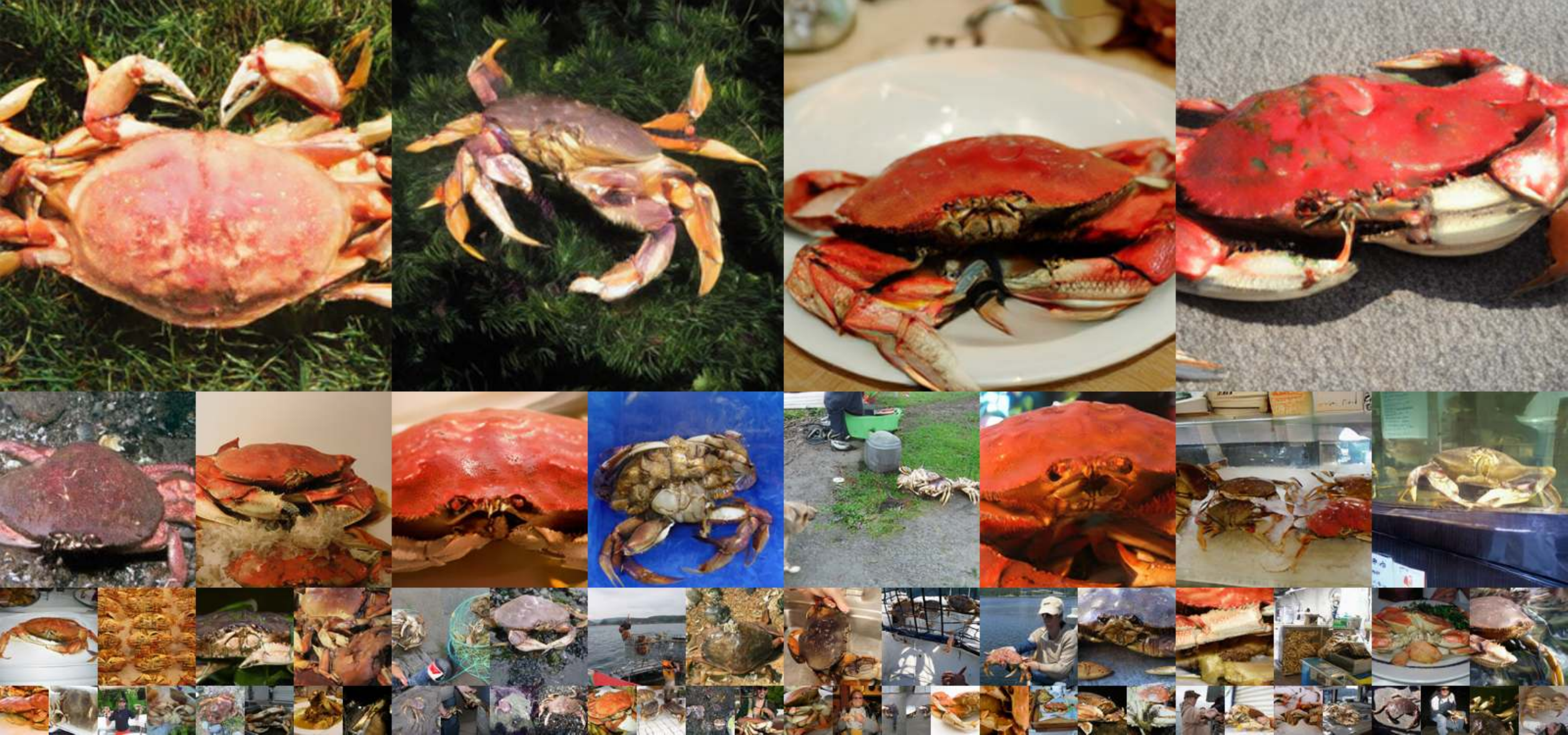}
    \caption{Uncurated 256$\times$256 generation results of LightningDiT-XL with MacTok 128 tokens. We use CFG with 3.0. Class label =``Dungeness crab'' (118).}
    \label{fig:118}
\end{figure*}

\begin{figure*}[tp]
    \centering
    \includegraphics[width=1\linewidth]{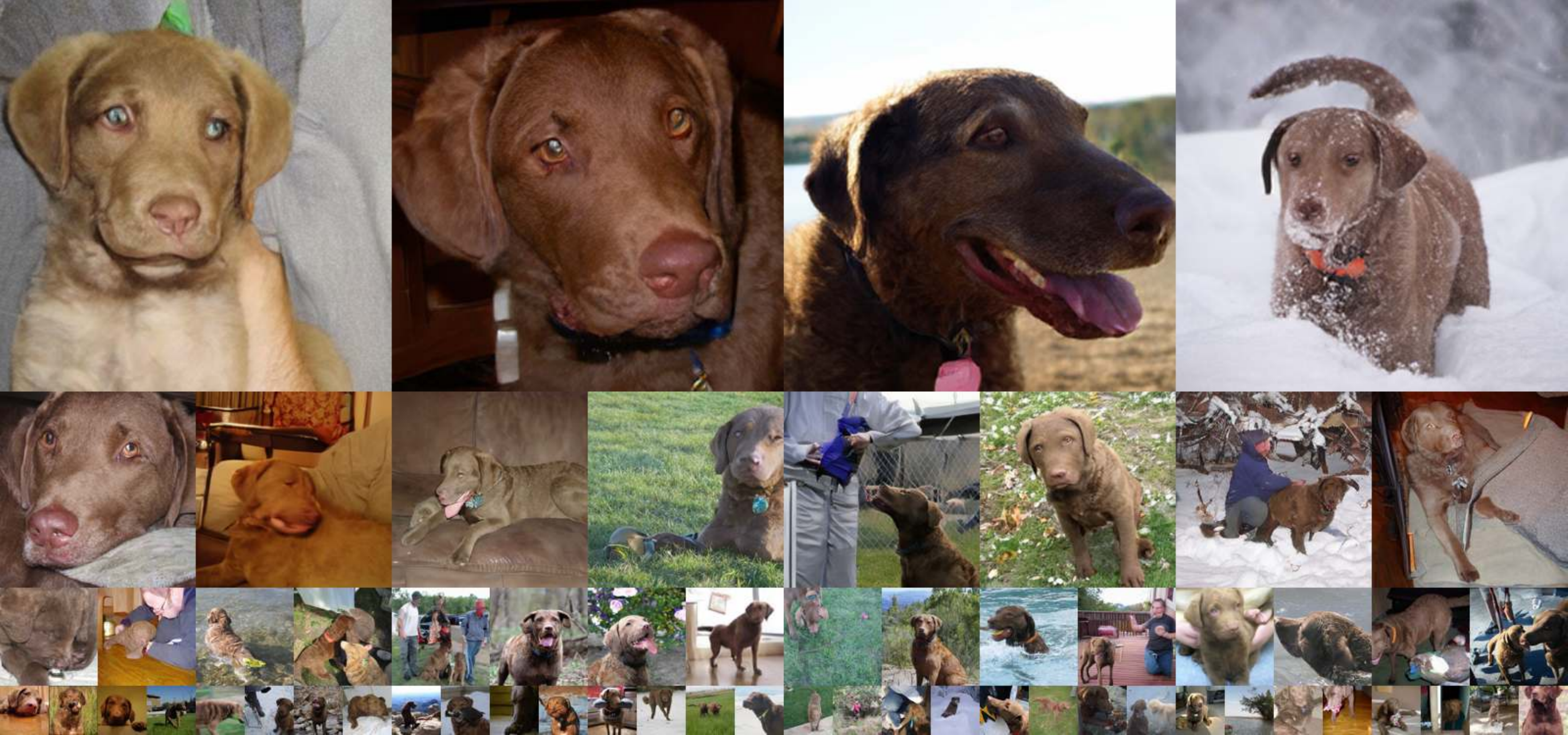}
    \caption{Uncurated 256$\times$256 generation results of LightningDiT-XL with MacTok 128 tokens. We use CFG with 3.0. Class label =``Chesapeake Bay retriever'' (209).}
    \label{fig:209}
\end{figure*}

\begin{figure*}[tp]
    \centering
    \includegraphics[width=1\linewidth]{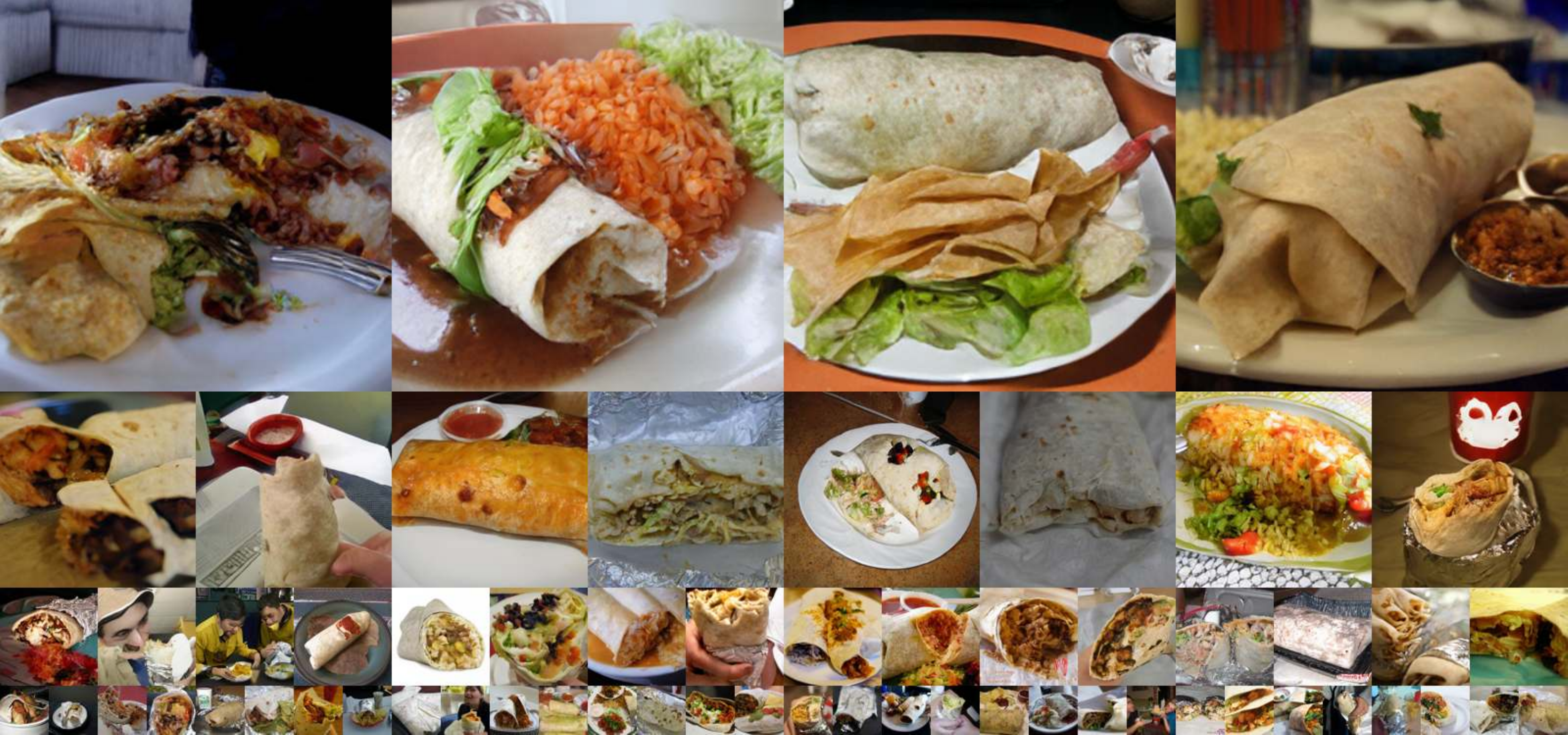}
    \caption{Uncurated 256$\times$256 generation results of LightningDiT-XL with MacTok 128 tokens. We use CFG with 3.0. Class label =``burrito'' (965).}
    \label{fig:965}
\end{figure*}

\begin{figure*}[tp]
    \centering
    \includegraphics[width=1\linewidth]{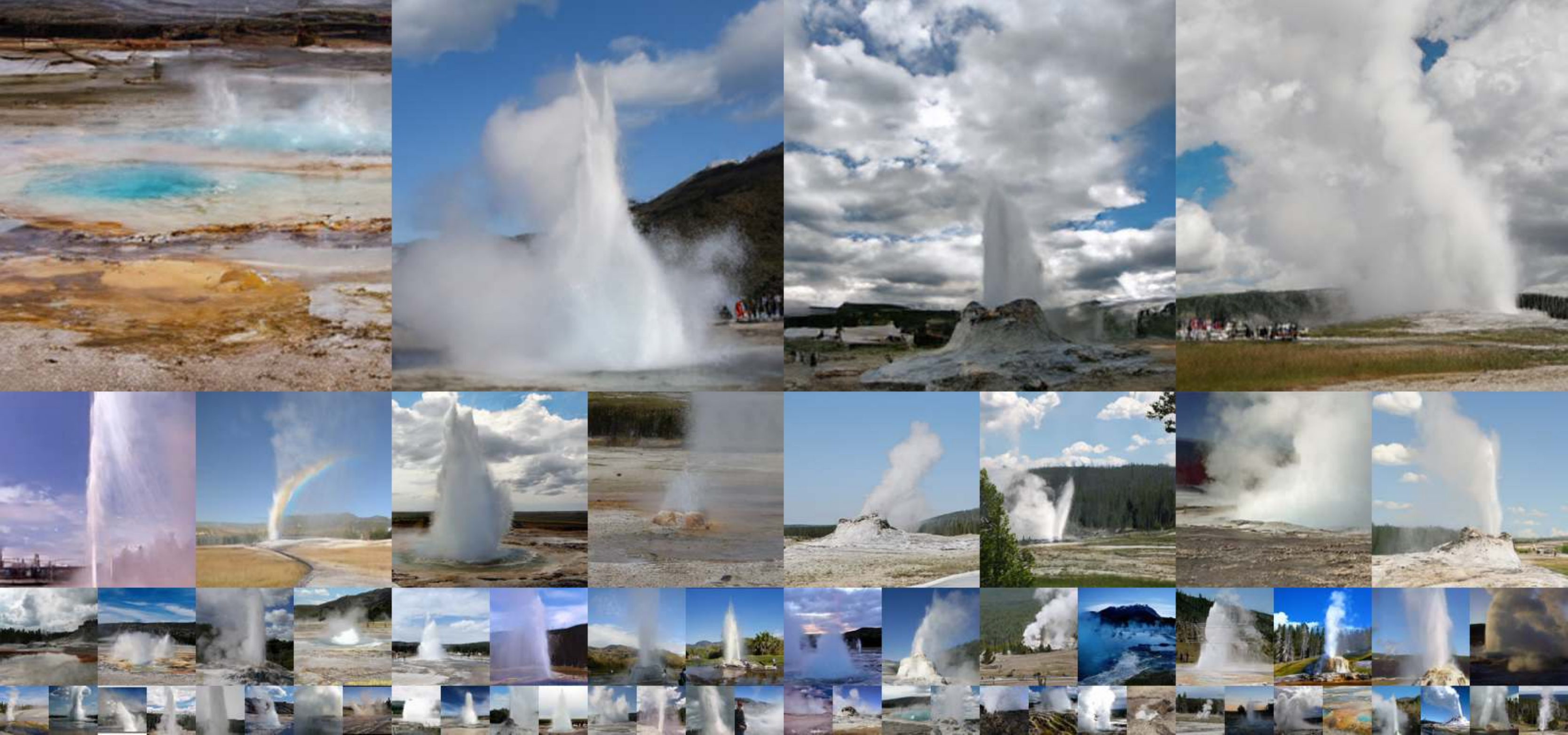}
    \caption{Uncurated 256$\times$256 generation results of LightningDiT-XL with MacTok 64 tokens. We use CFG with 3.0. Class label =``geyser'' (974).}
    \label{fig:974}
\end{figure*}

\begin{figure*}[tp]
    \centering
    \includegraphics[width=1\linewidth]{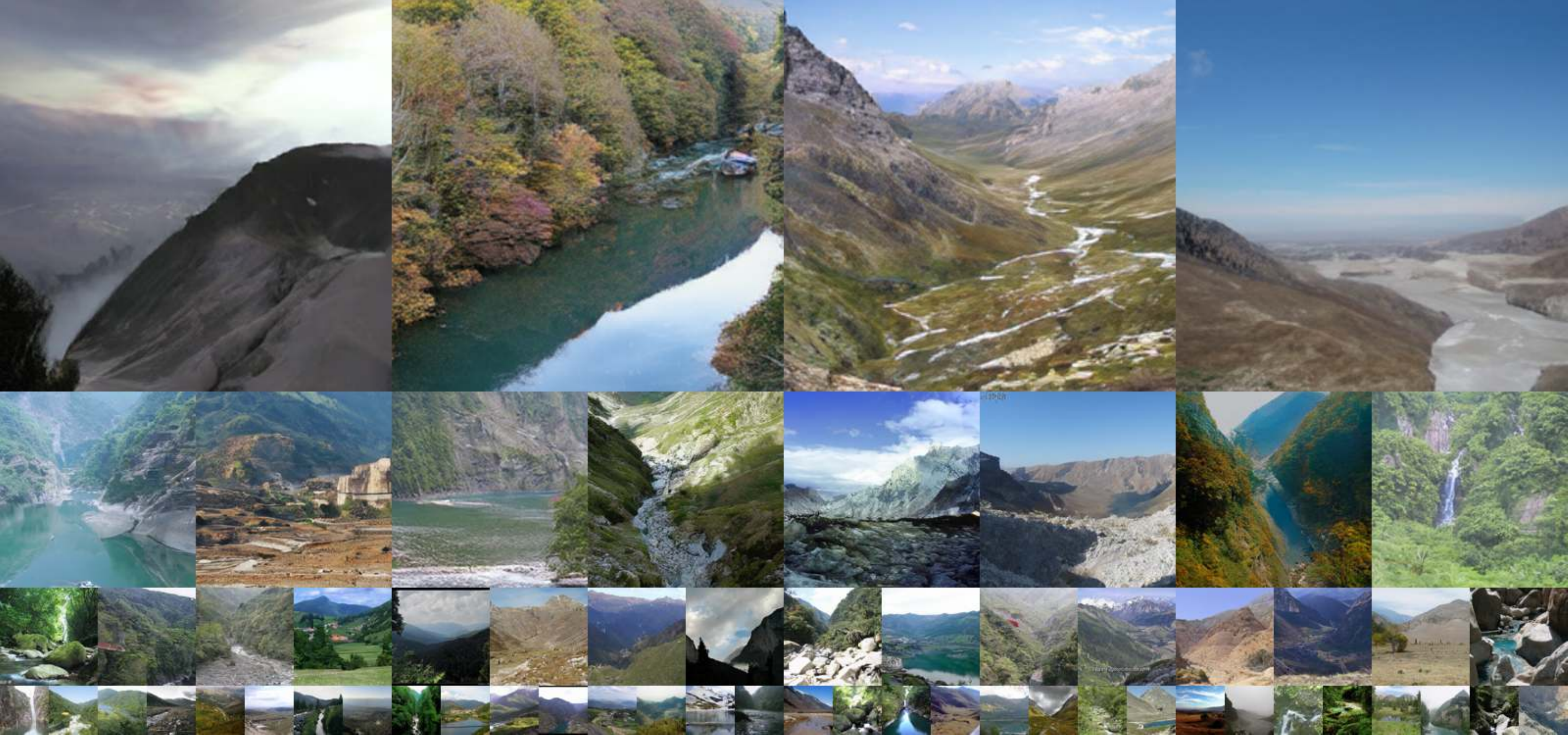}
    \caption{Uncurated 256$\times$256 generation results of LightningDiT-XL with MacTok 64 tokens. We use CFG with 3.0. Class label =``valley'' (979).}
    \label{fig:979}
\end{figure*}

\begin{figure*}[tp]
    \centering
    \includegraphics[width=1\linewidth]{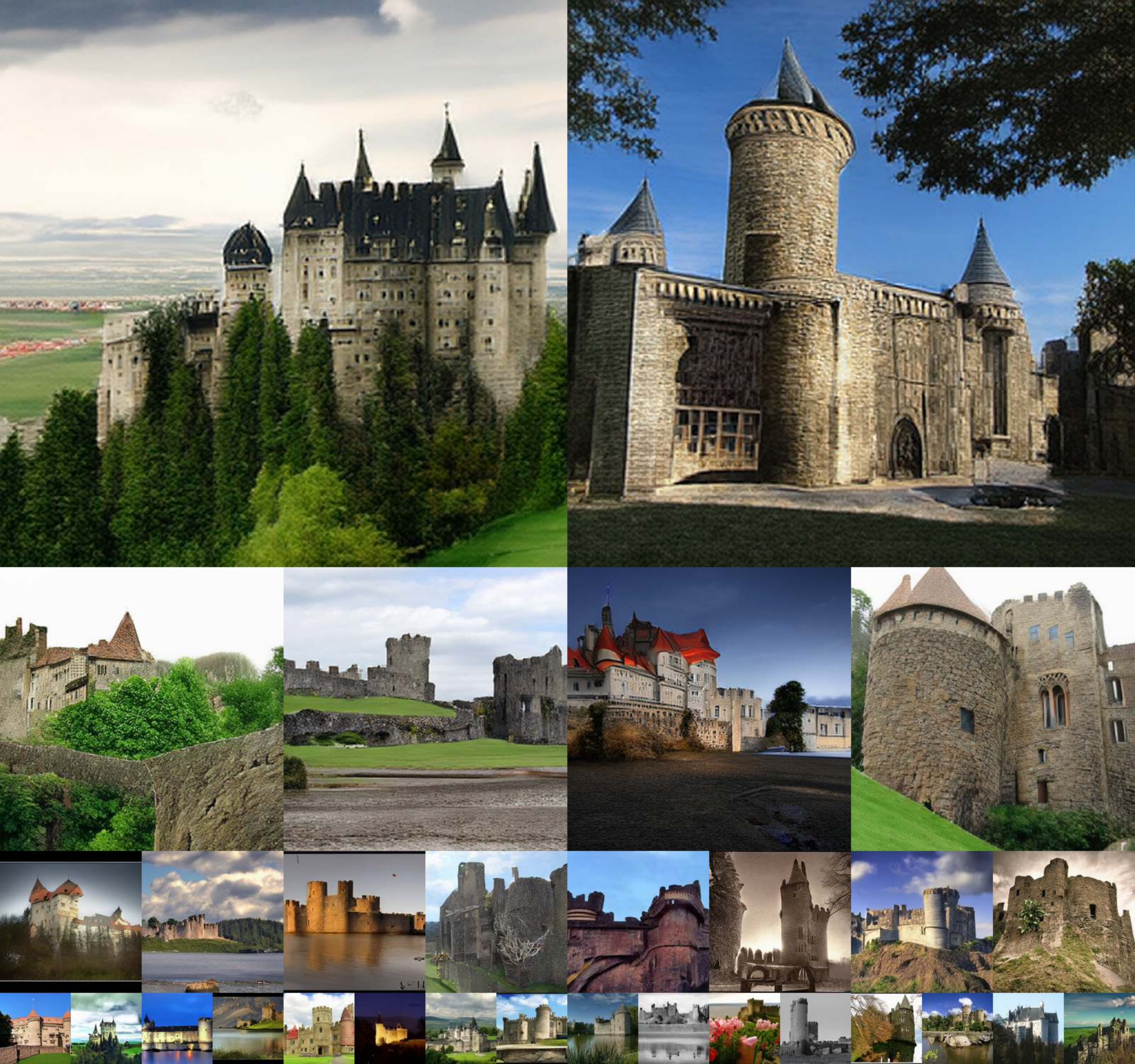}
    \caption{Uncurated 512$\times$512 generation results of SiT-XL with MacTok 128 tokens. We use CFG with 4.0. Class label =``castle'' (483).}
    \label{fig:483}
\end{figure*}

\begin{figure*}[tp]
    \centering
    \includegraphics[width=1\linewidth]{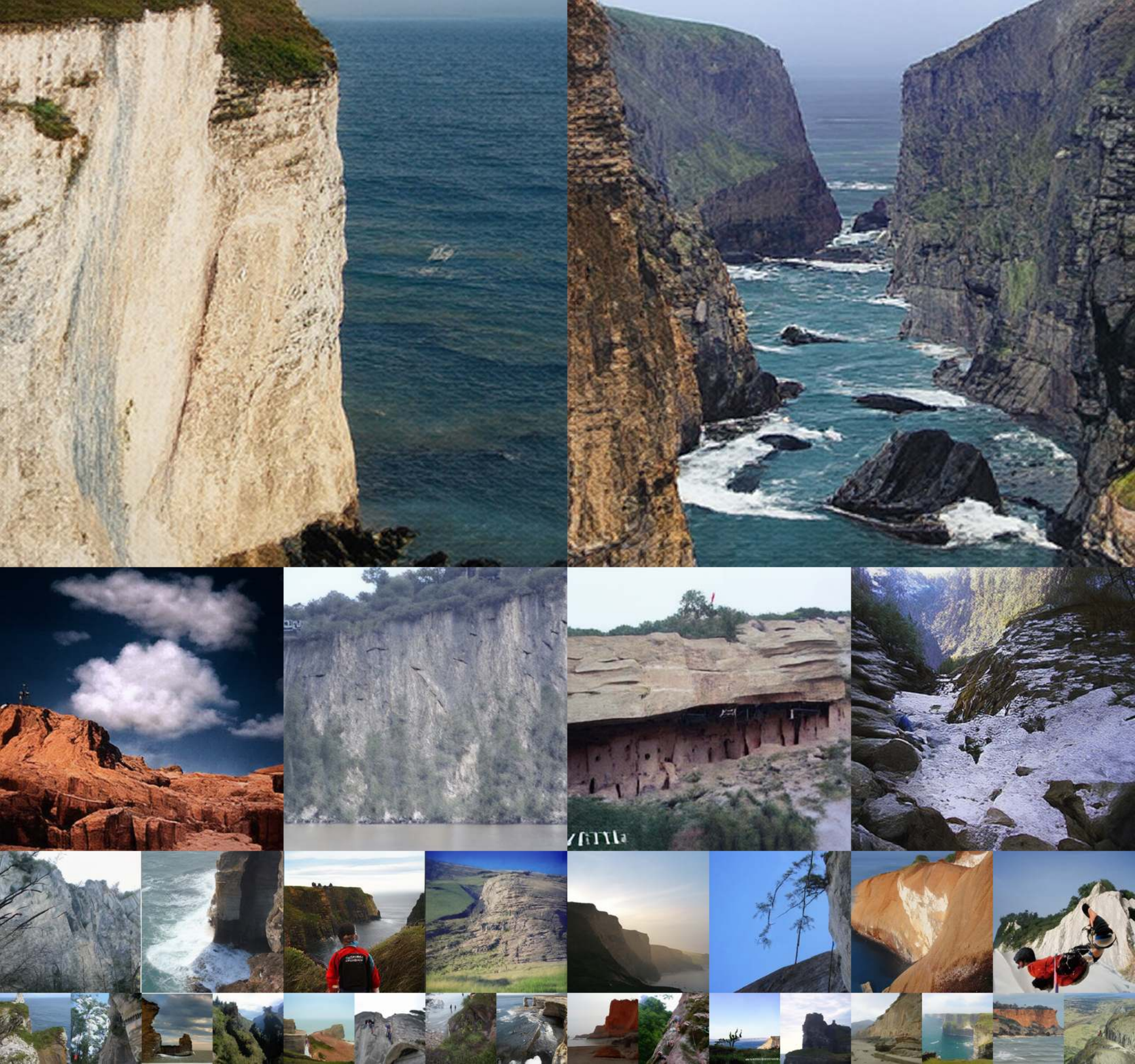}
    \caption{Uncurated 512$\times$512 generation results of SiT-XL with MacTok 128 tokens. We use CFG with 4.0. Class label =``cliff'' (972).}
    \label{fig:972}
\end{figure*}

\begin{figure*}[tp]
    \centering
    \includegraphics[width=1\linewidth]{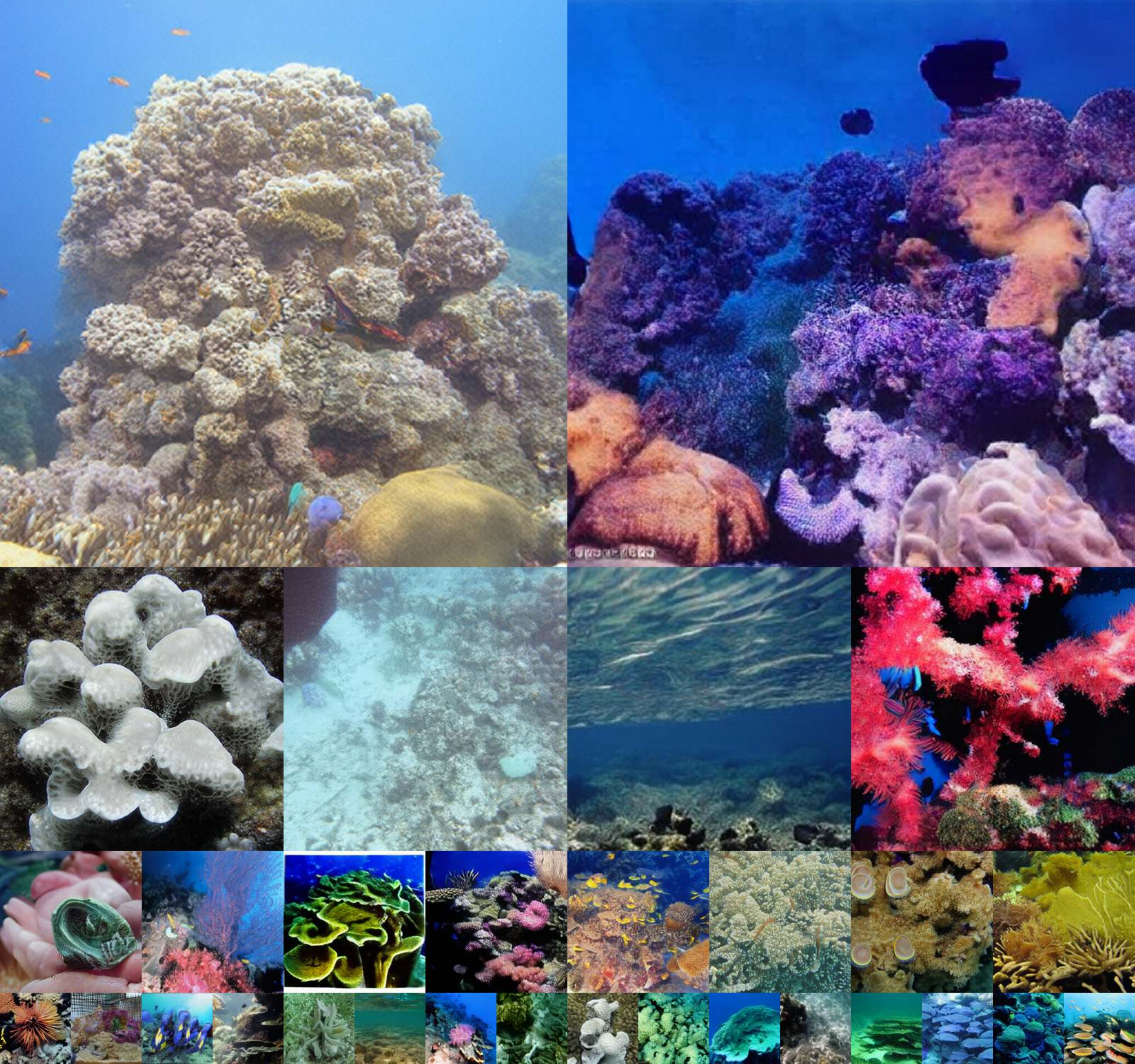}
    \caption{Uncurated 512$\times$512 generation results of SiT-XL with MacTok 64 tokens. We use CFG with 4.0. Class label =``coral reef'' (973).}
    \label{fig:973}
\end{figure*}

\begin{figure*}[tp]
    \centering
    \includegraphics[width=1\linewidth]{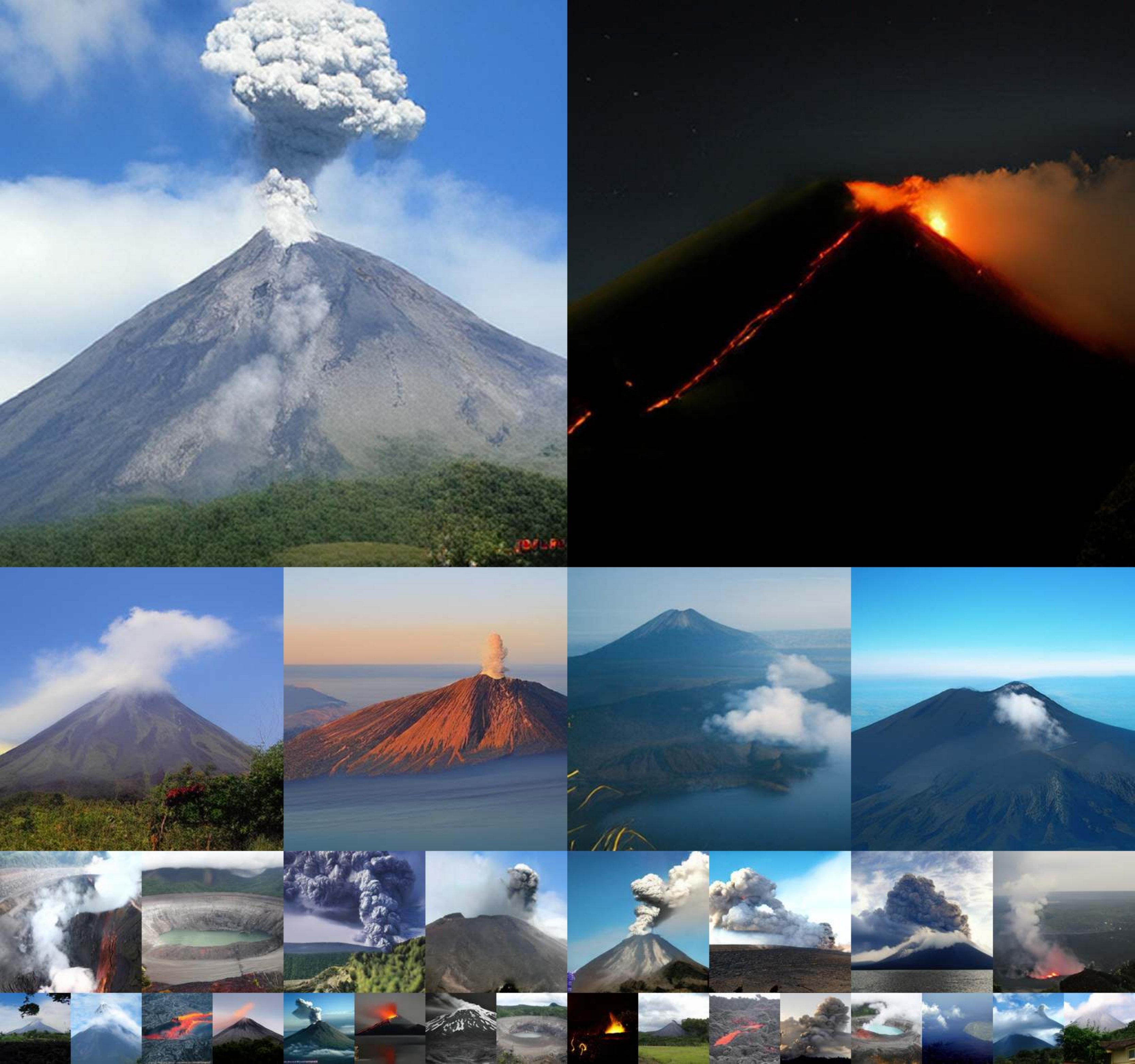}
    \caption{Uncurated 512$\times$512 generation results of SiT-XL with MacTok 64 tokens. We use CFG with 4.0. Class label =``volcano'' (980).}
    \label{fig:980}
\end{figure*}
















\end{document}